\pdfoutput=1

\documentclass{article}

\PassOptionsToPackage{numbers, compress}{natbib}
\usepackage[final]{neurips_2023}

\usepackage[utf8]{inputenc} % allow utf-8 input
\usepackage[T1]{fontenc}    % use 8-bit T1 fonts
\usepackage{graphicx}
\usepackage{xcolor}         % colors
\usepackage{microtype}      % microtypography
\usepackage{url}            % simple URL typesetting
\usepackage{amsmath}        % maths symbols
\usepackage{amsfonts}       % blackboard math symbols
\usepackage{mathtools}
\usepackage{nicefrac}       % compact symbols for 1/2, etc.
\usepackage{subcaption}     % supports subfigures and captions
\usepackage{caption}		% additional caption formatting
\usepackage{booktabs}       % professional-quality tables

\usepackage{hyperref}
\definecolor{mydarkblue}{rgb}{0,0.08,0.45}
\hypersetup{ %
    colorlinks=true,
    linkcolor=mydarkblue,
    citecolor=mydarkblue,
    filecolor=mydarkblue,
    urlcolor=mydarkblue
} %

\usepackage[ruled,longend,linesnumbered]{algorithm2e} % Provides typesetting for algorithms.
\DontPrintSemicolon

\newcommand{\eg}{e.g.,~}

\makeatletter
\newcommand{\settitle}{\@maketitle}
\makeatother

\title{Creating Multi-Level Skill Hierarchies in Reinforcement Learning}

\author{%
  Joshua B.~Evans\\
  Department of Computer Science\\
  University of Bath\\
  Bath, United Kingdom \\
  \texttt{jbe25@bath.ac.uk} \\
  \And
  {\"O}zg{\"u}r {\c{S}}im{\c{s}}ek \\
  Department of Computer Science\\
  University of Bath\\
  Bath, United Kingdom \\
  \texttt{o.simsek@bath.ac.uk} \\
}

\begin{document}

\maketitle

\begin{abstract}
What is a useful skill hierarchy for an autonomous agent? We propose an answer based on a graphical representation of how the interaction between an agent and its environment may unfold. Our approach uses modularity maximisation as a central organising principle to expose the structure of the interaction graph at multiple levels of abstraction. The result is a collection of skills that operate at varying time scales, organised into a hierarchy, where skills that operate over longer time scales are composed of skills that operate over shorter time scales. The entire skill hierarchy is generated automatically, with no human intervention, including the skills themselves (their behaviour, when they can be called, and when they terminate) as well as the hierarchical dependency structure between them. In a wide range of environments, this approach generates skill hierarchies that are intuitively appealing and that considerably improve the learning performance of the agent.
\end{abstract}

\section{Introduction}

One of the most important open problems in artificial intelligence is how to make it possible for autonomous agents to develop useful action hierarchies on their own, without any input from humans, including system designers and domain specialists. Before addressing this algorithmic question, it is useful to first consider a conceptual one: \textit{What constitutes a useful action hierarchy?} Here we focus on this conceptual question with the aim of providing a useful foundation for algorithmic development. 

Our primary contribution is a characterisation of a useful action hierarchy. In defining this action hierarchy, we use no information other than a graphical representation of how the interaction between the agent and its environment may unfold. When this information is not known a priori, it would be discovered naturally by the agent as it operates in its environment. Beyond this interaction graph, no particular domain knowledge is needed. Hence, our approach is applicable broadly. 

Multiple strands of earlier work have used the interaction graph as a basis for defining collections of actions. The main novelty in our approach is our use of \textit{modularity maximisation} as a central organising principle to expose the structure of the interaction graph at multiple levels of abstraction. The outcome is an action hierarchy that enables the agent to explore its environment efficiently at multiple time scales.

Our approach yields an action hierarchy with four desirable properties. First, it contains actions that operate at a wide range of time scales. This is necessary to solve complex problems, which require agents to be able to act, learn, and plan at varying time scales. Secondly, the actions are naturally organised into a hierarchy, with actions that operate over longer time scales being composed of actions that operate over shorter time scales. This hierarchical structure offers substantial benefits over unstructured collections of actions. For example, it allows an agent to learn about not only the action it is currently executing but also all lower-level actions that are called in the process. In addition, a hierarchical structure allows actions to be updated in a modular fashion. For instance, any improvements to an action would be immediately reflected in all higher-level actions that call it. Thirdly, the action hierarchy is fully specified. This includes when each action can be selected for execution, how exactly it behaves, and when it terminates. It also includes the number of levels in the hierarchy and the exact dependency structure between the actions. Fourthly, the action hierarchy is generated automatically, with no human intervention. 

In a diverse set of environments, the proposed approach translates into action hierarchies that are intuitively appealing. When evaluated within the context of reinforcement learning, they substantially improve learning performance compared to alternative approaches, with the largest performance improvement observed in the largest environment tested.

An important question for future research is how such an action hierarchy may be learned when the agent has no prior knowledge of the dynamics of the environment. We present an initial exploration of how this may be achieved, with positive results.

\section{Background}
\label{sect:background}
% MDP defined.
We use the reinforcement learning framework, modelling an agent's interaction with its environment as a finite Markov Decision Process (MDP). An MDP is a six-tuple \((\mathcal{S},\mathcal{A},\mathcal{P},\mathcal{R},\mathcal{D},\gamma)\), where \(\mathcal{S}\) is a finite set of states, \(\mathcal{A}\) is a finite set of actions, \(\mathcal{P}:\mathcal{S}\times\mathcal{A}\times\mathcal{S} \rightarrow [0,1]\) is a transition function, \(\mathcal{R}:\mathcal{S}\times\mathcal{A}\times\mathcal{S} \rightarrow \mathbb{R}\) is a reward function, \(\mathcal{D}:\mathcal{S} \rightarrow [0,1]\) is an initial state distribution, and \(\gamma \in [0,1]\) is a discount factor. Let $\mathcal{A}(s)$ denote the set of actions available in state $s \in \mathcal{S}$. At decision stage $t$, $t\ge 0$, the agent observes state $s_t \in \mathcal{S}$ and executes action $a_t \in \mathcal{A}(s_t)$. Consequently, at decision stage $t+1$, the agent receives reward $r_{t+1} \in \mathbb{R}$ and observes new state $s_{t+1} \in \mathcal{S}$. The \emph{return} at decision stage $t$, denoted by $G_t$, is the discounted sum of future rewards, \(G_{t} = \sum_{k=0}^{\infty} \gamma^{k}r_{t+k+1}\). A policy \(\pi:\mathcal{S}\times\mathcal{A} \rightarrow [0,1]\) specifies the probability of selecting action $a \in \mathcal{A}$ in state $s \in \mathcal{S}$. The objective is to learn a policy that maximises the expected return.

% STG defined.
The \emph{state transition graph} of an MDP is a weighted, directed graph whose nodes correspond to the states of the MDP and whose edges correspond to possible transitions between states. Specifically, an edge \((u,v)\) exists on the graph if it is possible to transition from state \(u \in \mathcal{S}\) to state \(v \in \mathcal{S}\) by taking some action \(a \in \mathcal{A}(u)\). In this paper, we use uniform edge weights of $1$.

% Options defined.
The actions of an MDP take exactly one decision stage to execute. We refer to them as \emph{primitive actions}. Using primitive actions, it is possible to define \emph{abstract actions}, also known as \emph{skills}, whose execution can take a variable number of decision stages. Furthermore, primitive and abstract actions can be combined to form complex action hierarchies. In this work, we represent abstract actions using the options framework~\cite{Sutton1999,Precup2000}. An option \(o\) is a three-tuple \((\mathcal{I}_{o},\pi_{o},\beta_{o})\), where \(\mathcal{I}_{o} \subset \mathcal{S}\) is the initiation set, specifying the set of states in which the option can start execution, \(\pi_{o}:\mathcal{S}\times\mathcal{A} \rightarrow [0,1]\) is the option policy, and \(\beta_{o}:\mathcal{S} \rightarrow [0,1]\) is the termination condition, specifying the probability of option termination in a given state. An option policy is ultimately defined in terms of primitive actions---because primitive actions are the fundamental units of interaction between the agent and its environment---but this can be done indirectly by allowing options to call other options, making it possible for agents to act, learn, and plan with hierarchies of primitive and abstract actions.

\section{Proposed Approach}
\label{sect:approach}

To define a skill hierarchy, we use {modularity maximisation} as a central organising principle, applied at multiple time scales. Specifically, we represent the possibilities of interaction between the agent and its environment as a graph and identify partitions of this graph that maximise modularity~\citep{Newman2004,Leicht2008,Arenas2007}.

A \emph{partition} of a graph is a division of its nodes into mutually exclusive groups, called \emph{clusters}. The \textit{modularity} of a partition composed of a set of clusters \(C = \lbrace c_{1}, c_{2}, \ldots, c_{k}\rbrace\) is
\[ \sum_{i=1}^{k} e_{ii} - \rho a^{2}_{i} \text{ ,} \]
where 
\(e_{ii}\) denotes the proportion of total edge weight in the graph that connects two nodes in cluster \(c_{i}\), and \(a_{i}\) denotes the proportion of total edge weight in the graph with at least one end connected to a node in cluster \(c_{i}\). A resolution parameter \(\rho > 0\) controls the relative importance of \(e_{ii}\) and \(a_{i}\). Intra-cluster edges contribute to both \(e_{ii}\) and \(a_{i}\) while inter-cluster edges contribute only to \(a_{i}\). A partition that maximises modularity will have relatively dense connections within its clusters and relatively sparse connections between its clusters.

Finding a partition that maximises modularity for a given graph is NP-complete~\citep{Brandes2006}. Therefore, when working with large graphs, approximation algorithms are needed. The most widely used approximation algorithm is the \emph{Louvain algorithm}~\citep{Blondel2008}, which is an agglomerative hierarchical graph clustering approach. While no formal analysis exists, the runtime of the Louvain algorithm has been observed empirically to be linear in the number of graph edges~\citep{Lancichinetti2009}. It has been successfully applied to graphs with millions of nodes and billions of edges~\cite{Blondel2008}.

An important feature of the Louvain algorithm is that, as a \textit{hierarchical} graph clustering method, it exposes the structure of a graph at multiple levels of granularity. Specifically, the output of the Louvain algorithm is a sequence of partitions of the input graph. This sequence has a useful structure: multiple clusters found in one partition in the sequence are merged into a single cluster in the next partition in the sequence. In other words, the output is a \emph{hierarchy} of clusters, with earlier partitions containing many smaller clusters that are merged into fewer larger clusters in later partitions. This hierarchical structure forms the basis of our characterisation of a useful multi-level skill hierarchy.

The Louvain algorithm starts by placing each node of the graph in its own cluster. Nodes are then iteratively moved locally, from their current cluster to a neighbouring cluster, until no gain in modularity is possible. This results in a revised partition corresponding to a local maximum of modularity with respect to local node movement. The revised partition is used to define an \emph{aggregate graph} as follows: each cluster in the partition is represented as a single node in the aggregate graph, and a directed edge is added to the aggregate graph if there is at least one edge that connects neighbouring clusters in the corresponding direction. This process is then repeated on the aggregate graph, and then on the next aggregate graph, and so on, until an iteration is reached with no modularity gain. Pseudocode for the Louvain algorithm is presented in Section~\ref{sect:louvain_pseudocode} of the supplementary material.

Let $h$ denote the number of partitions returned by the Louvain algorithm when applied to the state transition graph. We use each of the $h$ partitions to define a single layer of skills, resulting in an action hierarchy with $h$ levels of abstract actions (skills) above primitive actions. Each level of the hierarchy contains one or more skills for efficiently navigating between neighbouring clusters of the state transition graph. Specifically, we define an option for navigating from a cluster \(c_{i}\) to a neighbouring cluster \(c_{j}\) as follows: the initiation set consists of all states in \(c_{i}\); the option policy efficiently takes the agent from a given state in \(c_{i}\) to a state in \(c_{j}\); the option terminates with probability \(1\) in states in \(c_{j}\), with probability \(0\) otherwise.

Taking advantage of the natural hierarchical structure of the partitions produced by the Louvain algorithm, we compose the skills at one level of the hierarchy to define the skills at the next level up. That is, at each level of the hierarchy, option policies call actions (options or primitive actions) from the level below, with primitive actions being called directly by option policies from only the first level of the hierarchy.
We call the resulting set of skills the \emph{Louvain skill hierarchy}.

\section{Related Work}
\label{sect:related_work}

There have been earlier approaches to skill discovery using the state transition graph. The approach we propose here differs from them in two fundamental ways. First, it is novel in its use of modularity maximisation as a central organising principle for skill discovery. Secondly, it produces a multi-level hierarchy, whereas existing graph-based approaches produce hierarchies with only a single level of skills above primitive actions.

% --- Subgoal-Based Methods ---
% Menache2002, Simsek2005, Simsek2009, Kazemitabar2009, Entezari2010, Moradi2010, Rad2010, Imanian2011, Moradi2012, Taghizadeh2013, Kazemitabar2018, Ramesh2019
Many existing approaches to skill discovery use the state transition graph to identify useful subgoal states and define skills that efficiently take the agent to these subgoals. Suggestions for useful subgoals have often been inspired by the concept of a ``bottleneck''. They include states that are on the border of strongly-connected regions of the state space~\citep{Menache2002}, states that allow transitions between different regions of the state space~\citep{Simsek2004}, and states that lie on the shortest path between many pairs of states~\cite{Simsek2009}. To identify such states, several approaches use graph centrality measures~\citep{Simsek2009, Moradi2010, Rad2010, Imanian2011, Moradi2012}. Others use graph partitioning algorithms~\citep{Menache2002, Metzen2013, Simsek2005, Kazemitabar2009, Entezari2010, Bacon2013, Taghizadeh2013, Kazemitabar2018}. Alternatively, it has been proposed that ``landmark'' states found at the centre of strongly-connected regions of the state-space can be used as subgoals~\cite{Ramesh2019}.

% --- Clustering-Based Methods ---
% Clustering-Based: Mannor2004, Davoodabadi2011, Metzen2013, Shoeleh2017, Xu2018, Davoodabadi2019
% Our approach is most directly related to skill discovery methods that use graph partitioning~\citep{Metzen2013, Mannor2004, Davoodabadi2011, Shoeleh2017, Xu2018, Davoodabadi2019}.
The proposed approach is most directly related to skill discovery methods that make use of graph partitioning to identify meaningful regions of the state transition graph and define skills for navigating between them~\citep{Metzen2013, Mannor2004, Davoodabadi2011, Shoeleh2017, Xu2018, Davoodabadi2019}.
Three of these methods use the concept of modularity.
One such approach is to generate a series of possible partitions by successively removing the edge with the highest edge betweenness from a graph, then selecting the partition with the highest modularity~\citep{Davoodabadi2011}.
A second approach is to generate a partition using the label propagation algorithm and then to merge neighbouring clusters until no gain in modularity is possible~\citep{Davoodabadi2019}. In these two approaches, modularity maximisation is applied after first producing candidate partitions using different criteria. Consequently, the final partition does not maximise modularity overall. The time complexity of the label propagation method is near-linear in the number of graph edges, whereas the edge betweenness method has a time complexity of \(O(m^{2}n)\) on a graph with \(m\) edges and \(n\) nodes. A third approach is by \citet{Xu2018}, who use the Louvain algorithm to find a partition that maximises modularity but, unlike our approach, define skills only for moving between clusters in the highest-level partition, discarding all lower-level partitions. All three methods produce a single level of skills above primitive actions.

% --- Laplacian-Based Methods ---
% Machado2017, Jinnai2019
Several approaches have used the graph Laplacian~\citep{Machado2017, Jinnai2019} to identify skills that are specifically useful for efficiently exploring the state space. It is unclear how to arrange such skills to form multi-level skill hierarchies. In contrast, the proposed approach produces a set of skills that are naturally arranged into a multi-level hierarchy.

While existing graph-based methods do not learn multi-level hierarchies, policy-gradient methods have made some progress towards this goal. \citet{Bacon2017} extended policy-gradient theorems~\cite{Sutton1999PolicyGradient} to allow the learning of option policies and termination conditions in a hierarchy with a single level of skills above primitive actions. Riemer~et~al.~\cite{Riemer2018} further generalised these theorems to support multi-level hierarchies. \citet{Fox2017} propose an imitation learning method that finds the multi-level skill hierarchy most likely to generate a given set of example trajectories. \citet{Levy2019} propose a method for learning multi-level hierarchies of goal-directed policies, with each level of the hierarchy producing a subgoal for the lower levels to navigate towards. However, these methods are not without their limitations. Unlike the approach proposed here, they all require the number of hierarchy levels to be pre-defined instead of finding a suitable number automatically. They do not judiciously define initiation sets, instead making all skills available in all states. They also target different types of problems than we do, such as imitation-learning or goal-directed problems.

\section{Empirical Analysis}
\label{sect:analysis}

\begin{figure}[b]
     \centering
     \begin{subfigure}{1.0\linewidth}
	    \centering
     	\begin{subfigure}[b]{0.14\linewidth}
        	\centering
         	\includegraphics[width=\linewidth]{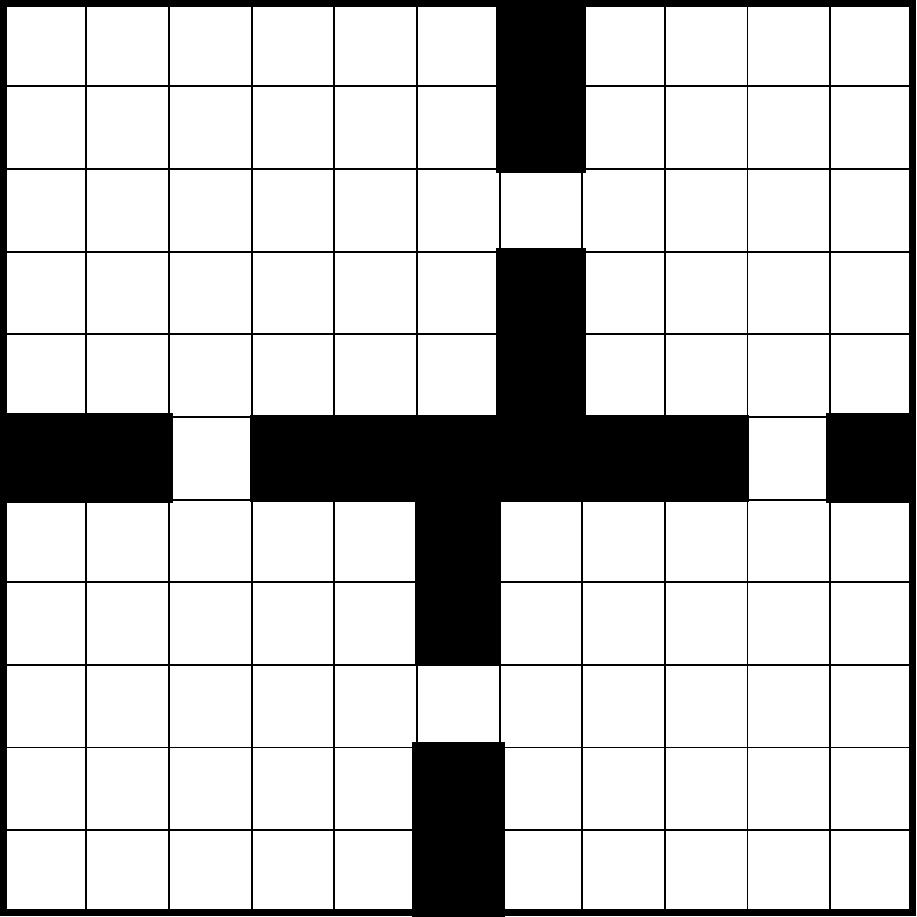}
		 	\caption*{Rooms}
     	\end{subfigure}%
     	\hfill
     	\begin{subfigure}[b]{0.14\linewidth}
        	\centering
         	\includegraphics[width=\linewidth]{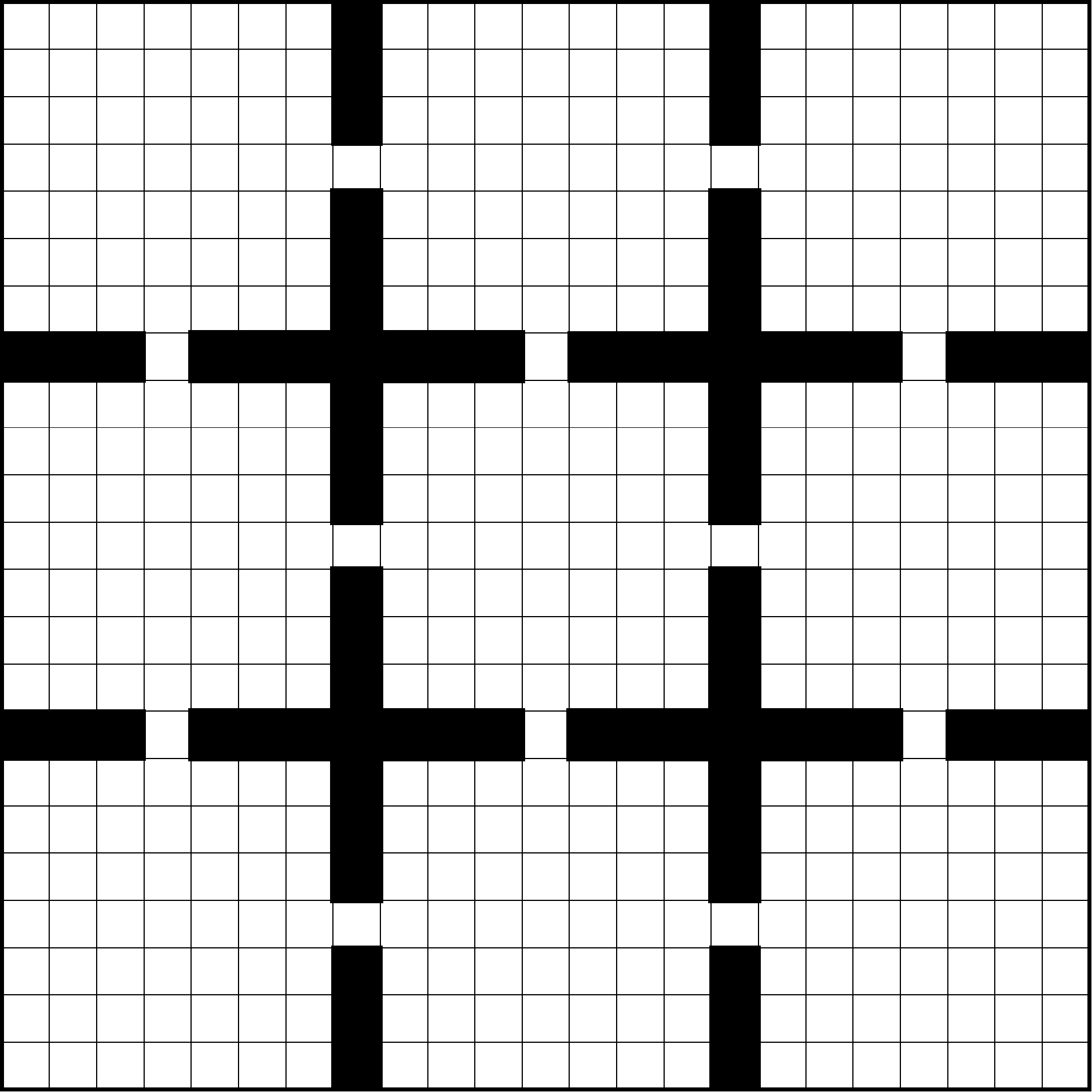}
		 	\caption*{Grid}
     	\end{subfigure}%
     	\hfill
     	\begin{subfigure}[b]{0.1558\linewidth}
        	\centering
        	\includegraphics[width=\linewidth]{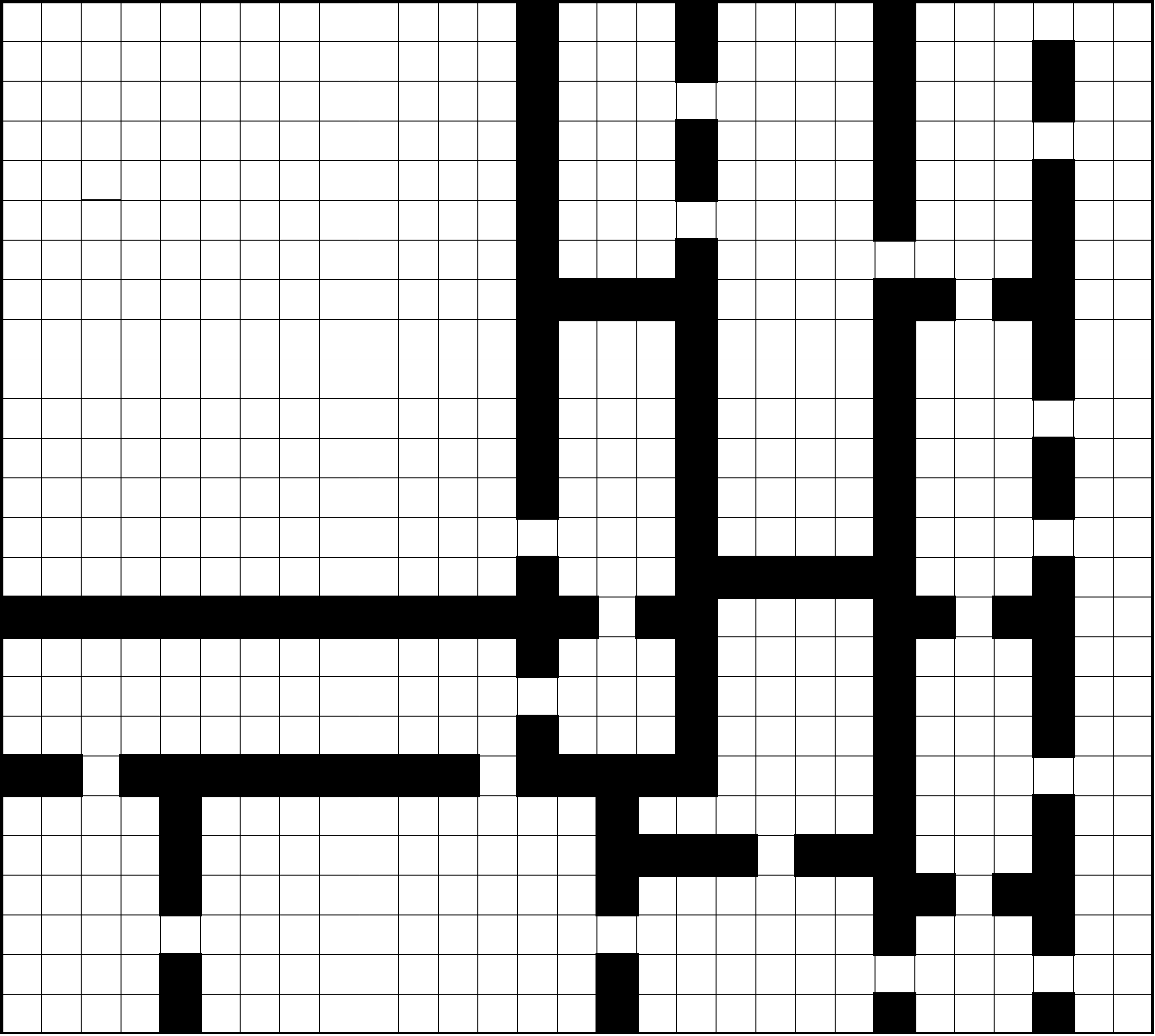}
	        \caption*{Maze}
     	\end{subfigure}%
     	\hfill
     	\begin{subfigure}[b]{0.175\linewidth}
        	\centering
			\includegraphics[width=\linewidth]{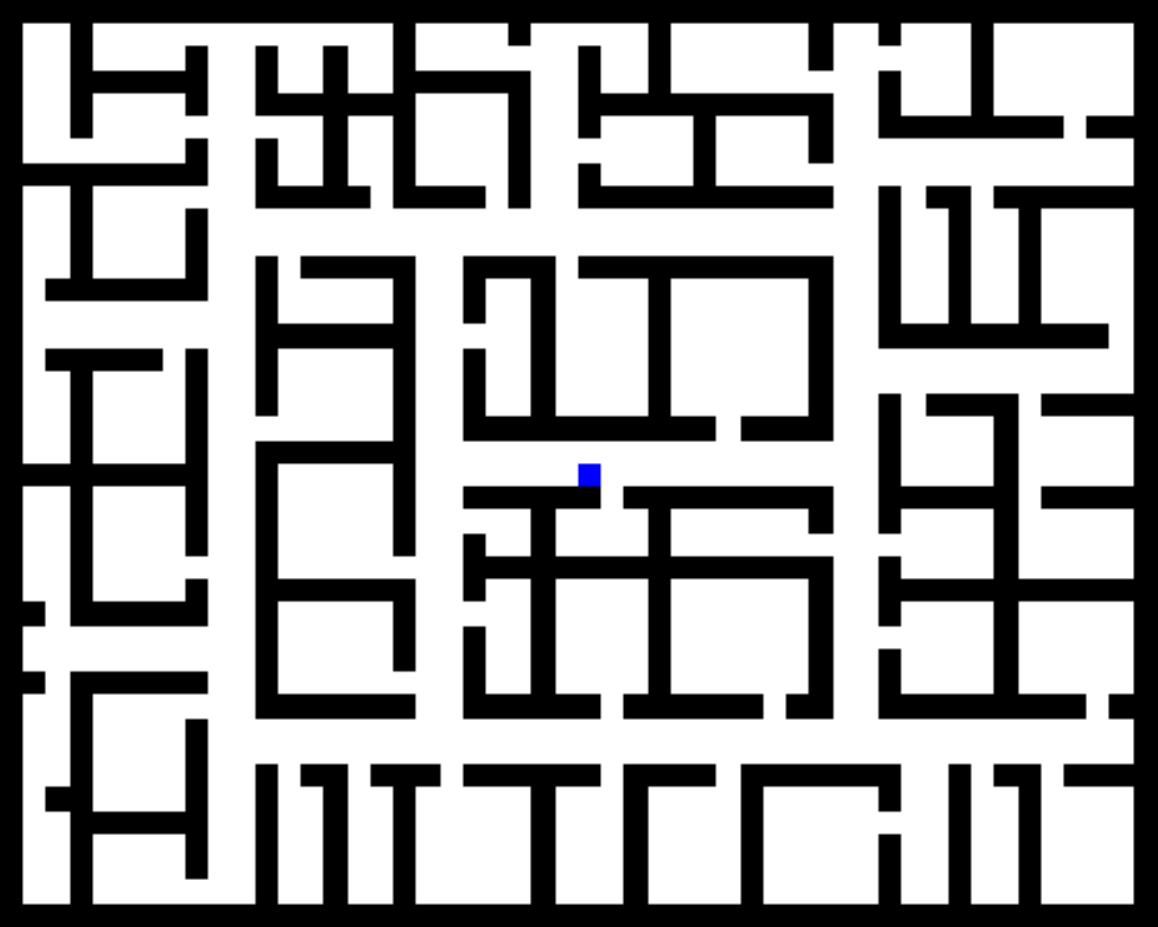}
			\caption*{Office}
     	\end{subfigure}%
     	\hfill
	    \begin{subfigure}[b]{0.14\linewidth}
			\centering    	     
			\includegraphics[width=\linewidth]{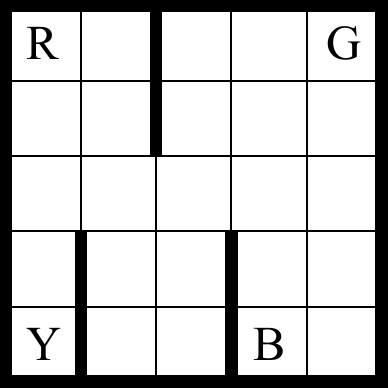}
         	\caption*{Taxi}
    	 \end{subfigure}%
	     \hfill
    	 \begin{subfigure}[b]{0.18\linewidth}
    	 	\centering
    	 	\includegraphics[width=\linewidth]{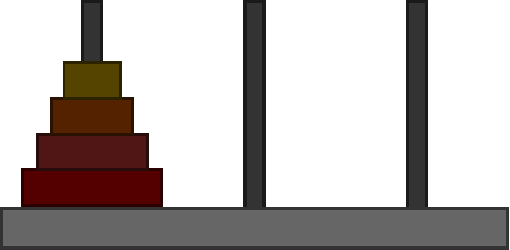}
         	\caption*{Towers of Hanoi}
	    \end{subfigure}%
	\end{subfigure}%
	\caption{The environments.}
	\label{fig:Domains}
\end{figure}

We analyse the Louvain skill hierarchy in the six environments depicted in Figure~\ref{fig:Domains}: Rooms, Grid, Maze~\citep{Ramesh2019}, Office, Taxi~\cite{Dietterich2000}, and Towers of Hanoi. In all environments, the reward is \(-0.001\) for each action and an additional \(+1.0\) for reaching a goal state. In  Rooms, Grid, Maze, and Office, there are four primitive actions: north, south, east, and west. In Taxi, there are two additional primitive actions: pick-up-passenger and put-down-passenger. Some decisions in Taxi are irreversible. For instance, after picking up the passenger, the agent cannot return to a state where the passenger is still waiting to be picked up. We describe each of these environments fully in Section~\ref{sect:sample_domains} of the supplementary material.

When generating partitions of the state transition graph using the Louvain algorithm, we used a resolution parameter of \(\rho = 0.05\), unless stated otherwise. When converting the output of the Louvain algorithm into a concrete skill hierarchy, we discarded all levels of the cluster hierarchy where the mean number of nodes per cluster was less than \(4\).
Our reasoning is that skills that navigate between such small clusters execute for only a very small number of decision stages (often only one or two) and are not meaningfully more abstract than primitive actions.
For all methods used in our comparisons, we generated options using the complete state transition graph and learned their policies offline using macro-Q learning~\cite{McGovern1997}.
We trained all hierarchical agents using macro-Q learning and intra-option learning~\cite{Precup1998}. Although these algorithms have not previously been applied to multi-level hierarchies, they both extend naturally to this case.
The primitive agent was trained using Q-Learning \cite{Watkins1989}.
The shaded regions on the learning curves represent the standard error in \(40\) independent runs.
All experiments are fully described in Section~\ref{sect:methodology} of the supplementary material.

Our analysis is directed by the following questions:
What is the Louvain skill hierarchy generated in each environment?
How does this skill hierarchy impact the learning performance of the agent? 
How do the results change as the number of states increases? 
Does arranging skills into a multi-level hierarchy provide benefits over a flat arrangement of the same skills?
What is the impact of varying the value of the resolution parameter \(\rho\)?

\textbf{Louvain Skill Hierarchy.} We first examine the cluster hierarchies generated by applying the Louvain algorithm to the state transition graphs of various  environments. Figure~\ref{fig:skills} shows the results in Rooms, Office, Taxi, and Towers of Hanoi. Section~\ref{sect:full_sample_hierarchies} of the supplementary material shows additional results in Grid and Maze.

\newsavebox{\imagebox}
\begin{figure}[t]
    \centering
	\begin{subfigure}{\linewidth}    
        \savebox{\imagebox}{\begin{subfigure}{0.24\linewidth}\includegraphics[width=0.8\linewidth]{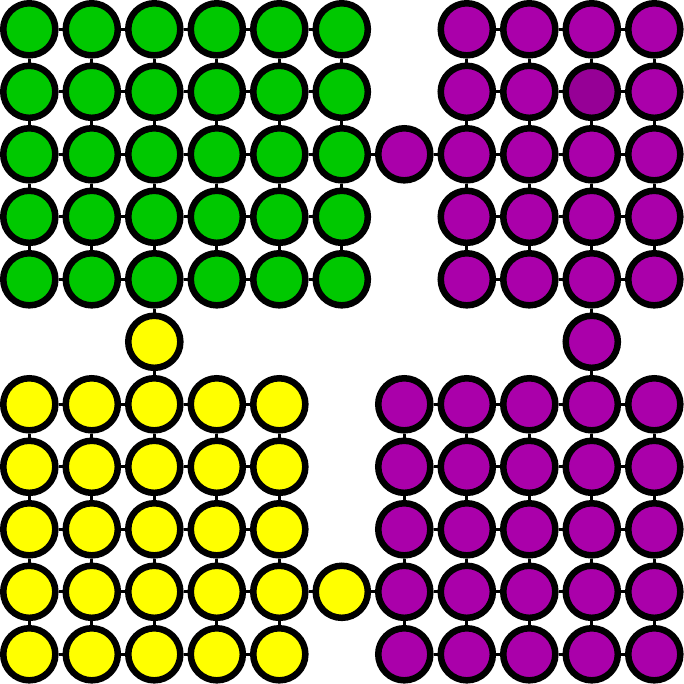}\end{subfigure}}% 
    	\begin{subfigure}{0.0175 \linewidth}
			\centering\raisebox{\dimexpr.5\ht\imagebox-.5\height+\baselineskip}{% Raise smaller image into place
            \includegraphics[width=\linewidth]{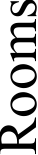}}%
		\end{subfigure}%
        \hspace{0.01\linewidth}%
        \begin{subfigure}{0.24 \linewidth}
        	\setlength{\abovecaptionskip}{2.0pt}
            \centering
            \includegraphics[width=0.8\linewidth]{"figures/hierarchy_examples/rooms/rooms_level_4".pdf}
            \caption*{Level 4}
        \end{subfigure}%
        \hfill%
        \begin{subfigure}{0.24 \linewidth} 
            \centering
        	\setlength{\abovecaptionskip}{2.0pt}
            \includegraphics[width=0.8\linewidth]{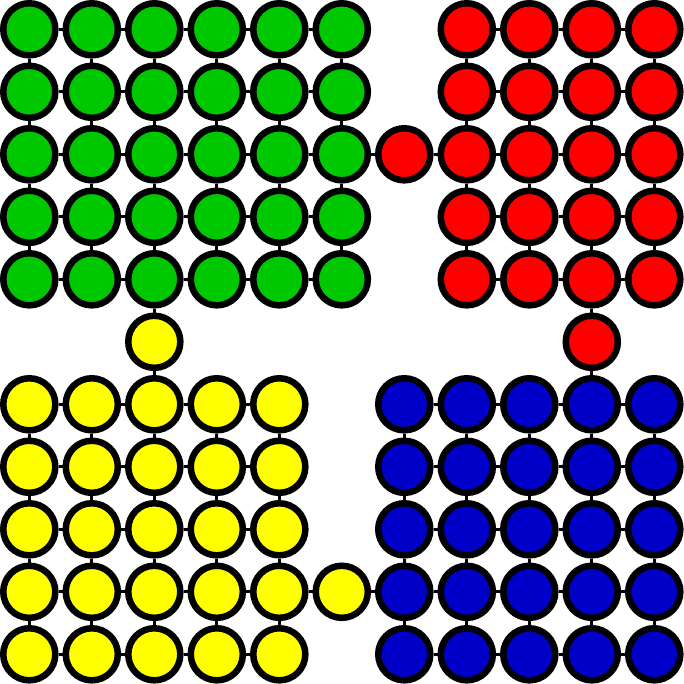}
            \caption*{Level 3}
        \end{subfigure}%
        \hfill     
        \begin{subfigure}{0.24 \linewidth}     
            \centering
           	\setlength{\abovecaptionskip}{2.0pt}
            \includegraphics[width=0.8\linewidth]{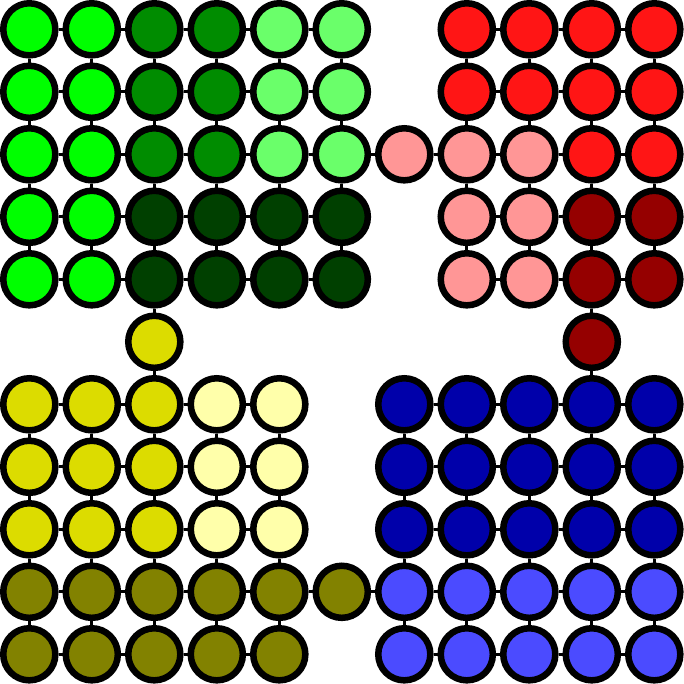}
            \caption*{Level 2}
        \end{subfigure}%
        \hfill 
        \begin{subfigure}{0.24 \linewidth}     
            \centering
            \setlength{\abovecaptionskip}{2.0pt}
            \includegraphics[width=0.8\linewidth]{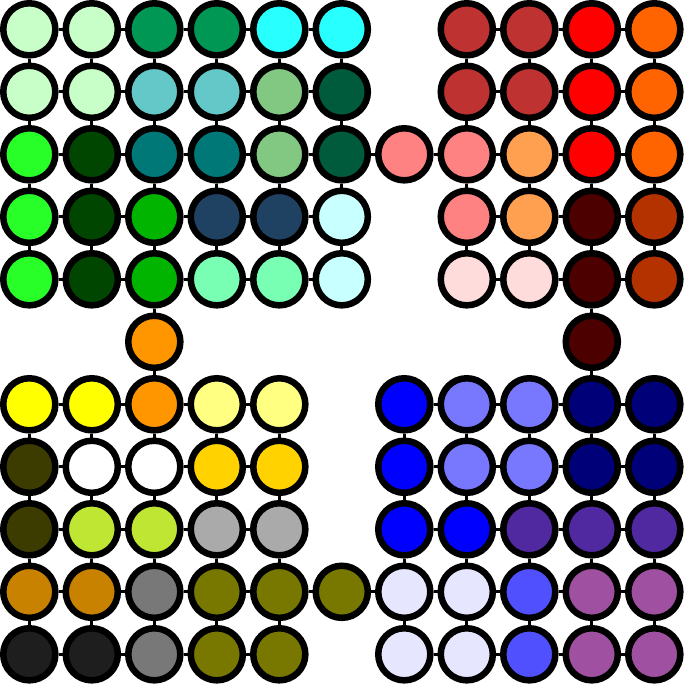}
			\caption*{Level 1}
        \end{subfigure}%
    \end{subfigure}%

	\vspace{0.4cm}
	\begin{subfigure}{\linewidth}
        \savebox{\imagebox}{\begin{subfigure}{0.24\linewidth}\includegraphics[width=0.9\linewidth]{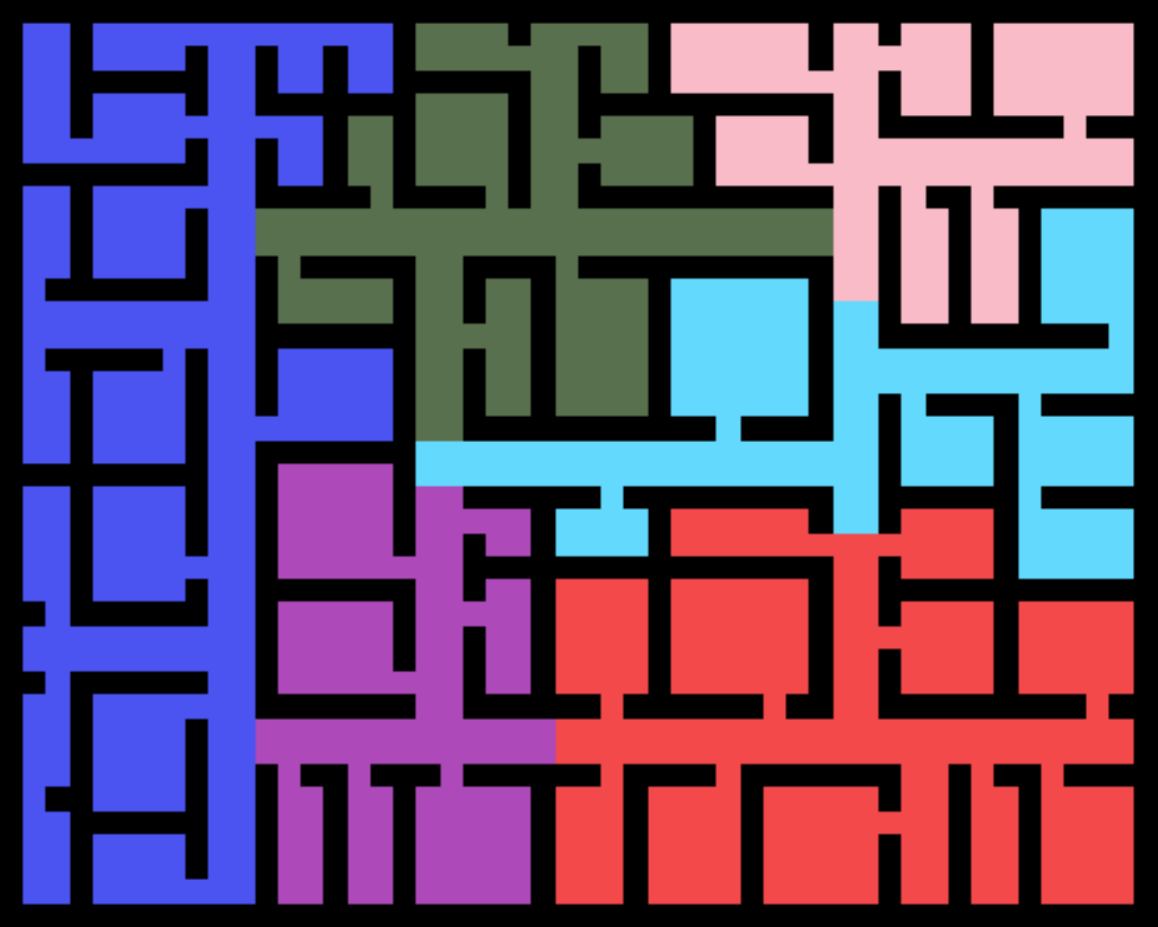}\end{subfigure}}% 
    	\begin{subfigure}{0.0175 \linewidth}
			\centering\raisebox{\dimexpr.5\ht\imagebox-.5\height+\baselineskip}{% Raise smaller image into place
            \includegraphics[width=\linewidth]{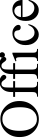}}%
		\end{subfigure}%
        \hspace{0.01\linewidth}%
        \begin{subfigure}{0.24 \linewidth}
            \centering
            \setlength{\abovecaptionskip}{2.0pt}
            \includegraphics[width=0.9\linewidth]{"figures/hierarchy_examples/office/office1k_level_4".pdf}
			\caption*{Level 5}
        \end{subfigure}%
        \hfill
        \begin{subfigure}{0.24 \linewidth}
            \centering
           	\setlength{\abovecaptionskip}{2.0pt}
            \includegraphics[width=0.9\linewidth]{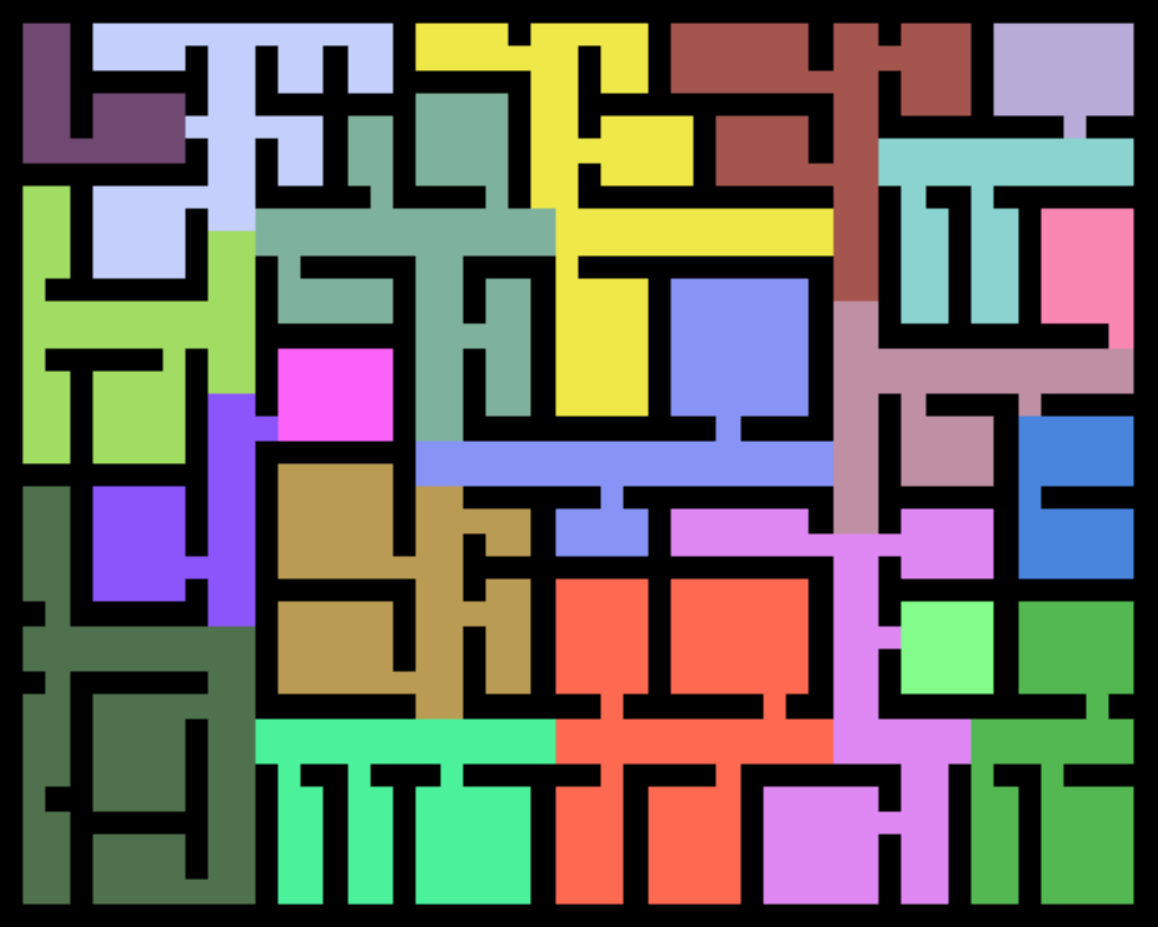}
			\caption*{Level 4}
        \end{subfigure}%
        \hfill
        \begin{subfigure}{0.24 \linewidth}
            \centering
           	\setlength{\abovecaptionskip}{2.0pt}
            \includegraphics[width=0.9\linewidth]{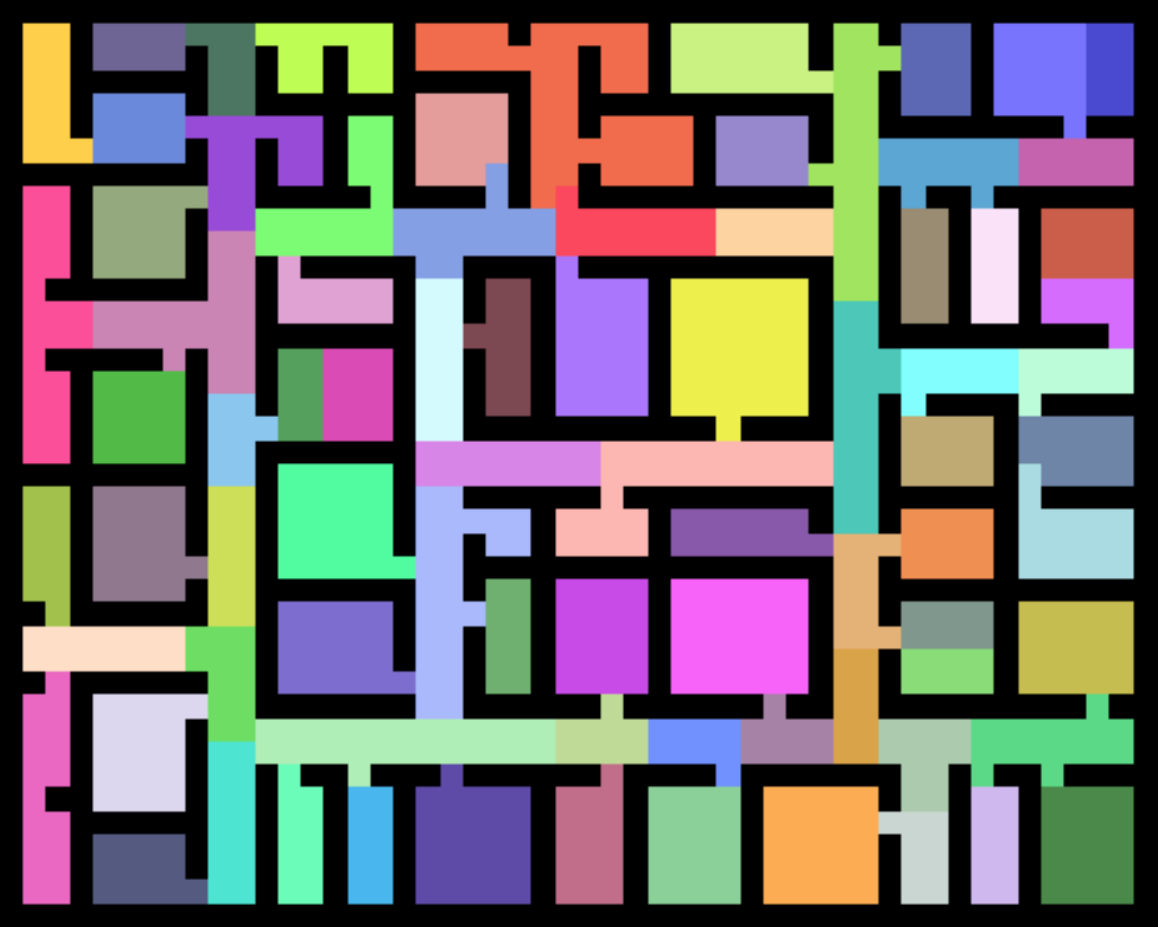}
            \caption*{Level 3}
        \end{subfigure}%
        \hfill
        \begin{subfigure}{0.24 \linewidth}
            \centering
           	\setlength{\abovecaptionskip}{2.0pt}
            \includegraphics[width=0.9\linewidth]{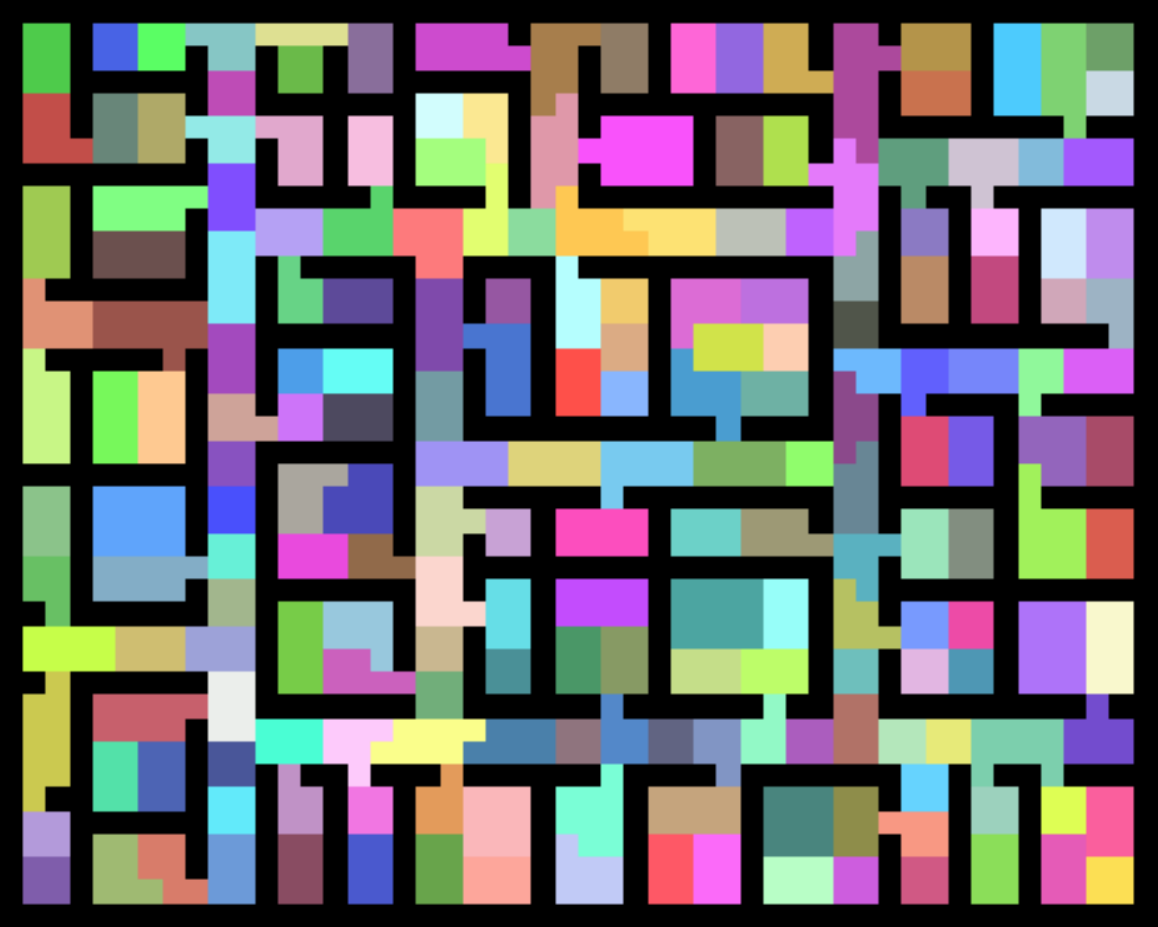}
            \caption*{Level 2}
        \end{subfigure}%
    \end{subfigure}% 

	\vspace{0.4cm}
    \begin{subfigure}{\linewidth}
        \savebox{\imagebox}{\begin{subfigure}{0.24\linewidth}\includegraphics[width=0.95\linewidth]{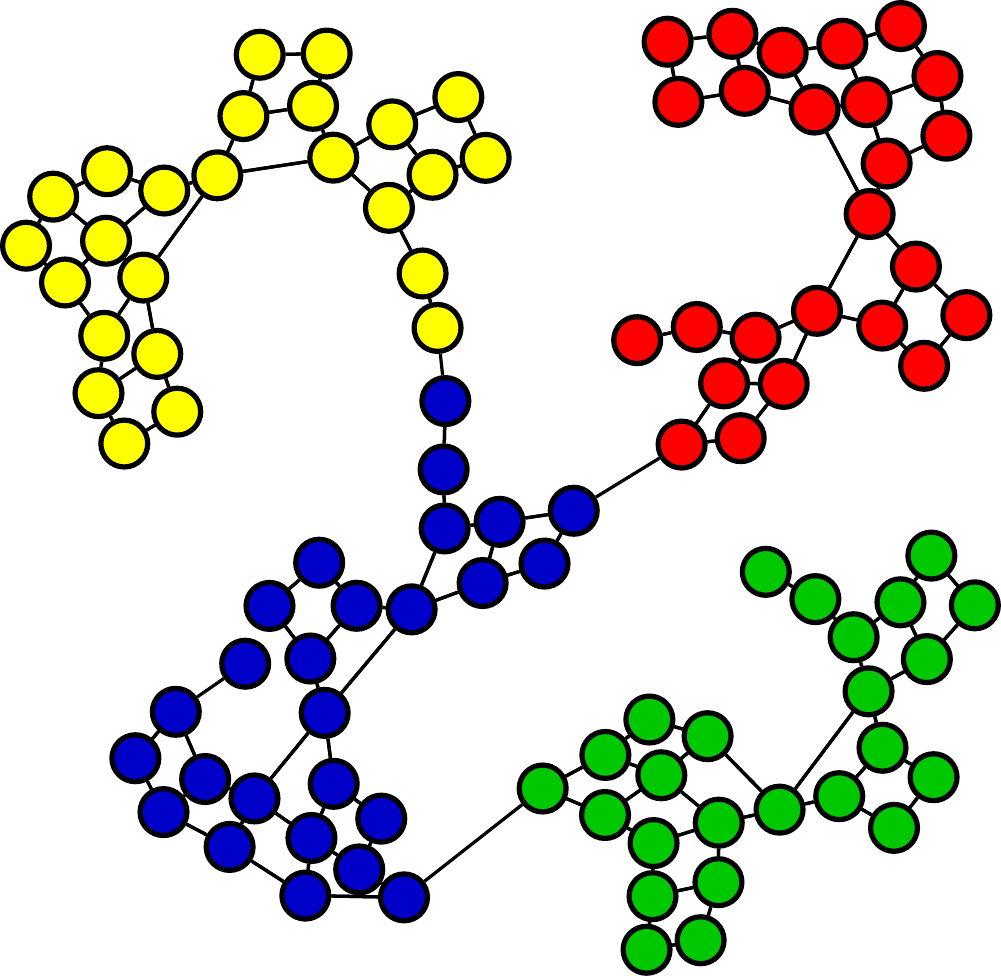}\end{subfigure}}% 
    	\begin{subfigure}{0.0175 \linewidth}
			\centering\raisebox{\dimexpr.5\ht\imagebox-.5\height+\baselineskip}{% Raise smaller image into place
            \includegraphics[width=\linewidth]{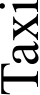}}%
		\end{subfigure}%
        \hspace{0.01\linewidth}%
        \begin{subfigure}{0.24\linewidth}
            \centering
           	\setlength{\abovecaptionskip}{2.0pt}
            \includegraphics[width=0.95\linewidth]{"figures/hierarchy_examples/taxi/taxi_level_4".pdf}
            \caption*{Level 4}
        \end{subfigure}%
        \hfill
        \begin{subfigure}{0.24\linewidth}
            \centering
            \setlength{\abovecaptionskip}{2.0pt}
            \includegraphics[width=0.95\linewidth]{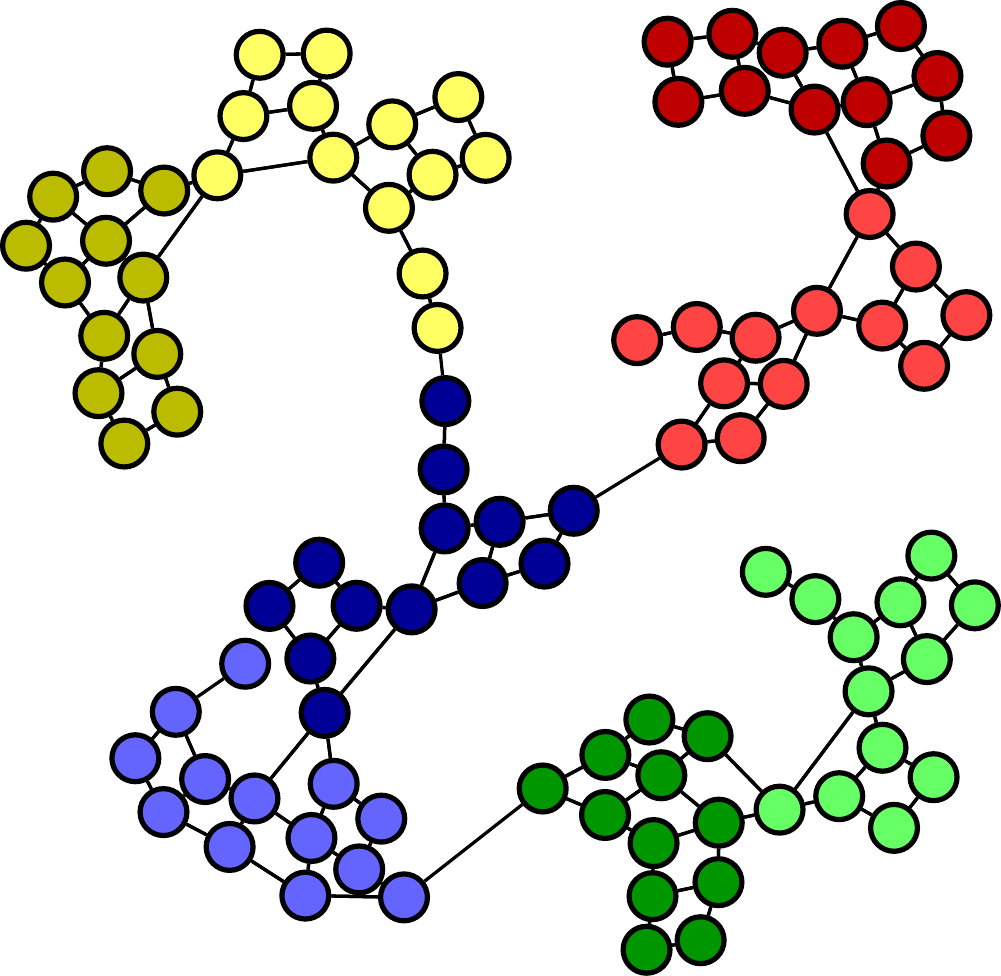}
			\caption*{Level 3}
        \end{subfigure}%
        \hfill
        \begin{subfigure}{0.24\linewidth}
            \centering
            \setlength{\abovecaptionskip}{2.0pt}
            \includegraphics[width=0.95\linewidth]{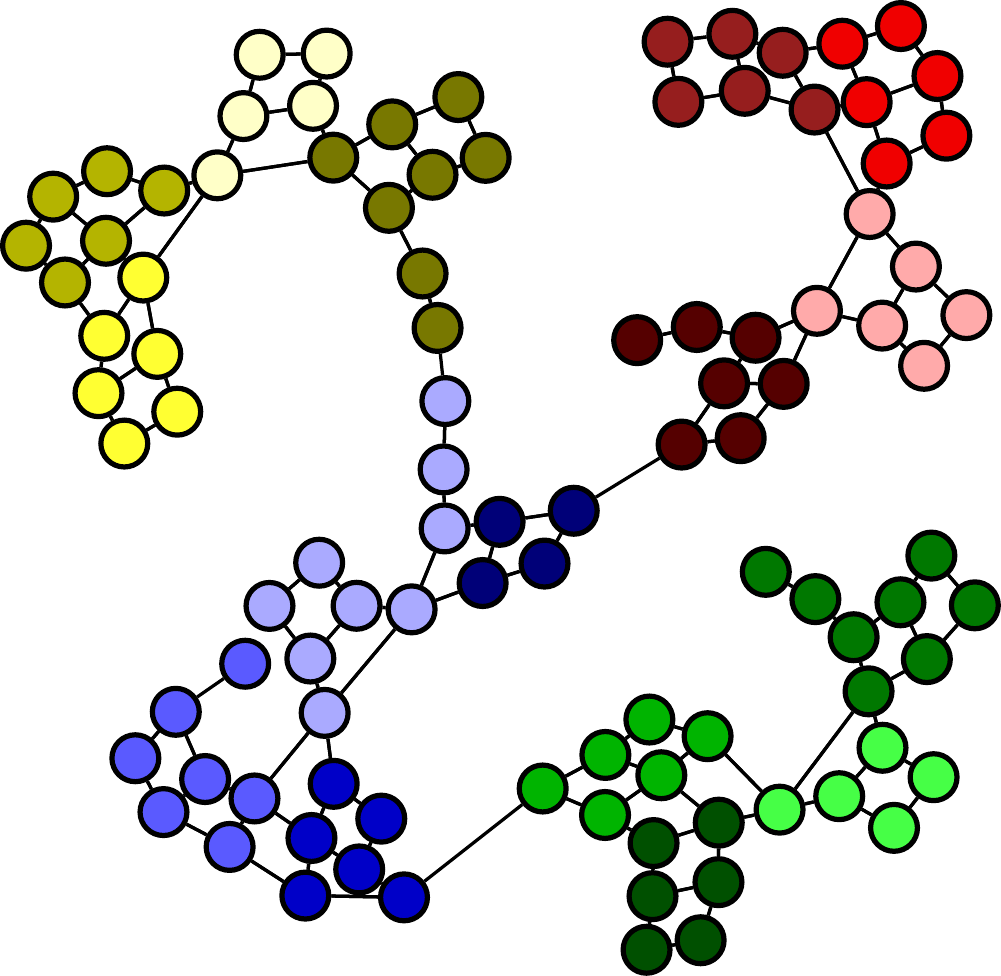}
            \caption*{Level 2}
        \end{subfigure}%
        \hfill
        \begin{subfigure}{0.24\linewidth}
            \centering
           	\setlength{\abovecaptionskip}{2.0pt}
            \includegraphics[width=0.95\linewidth]{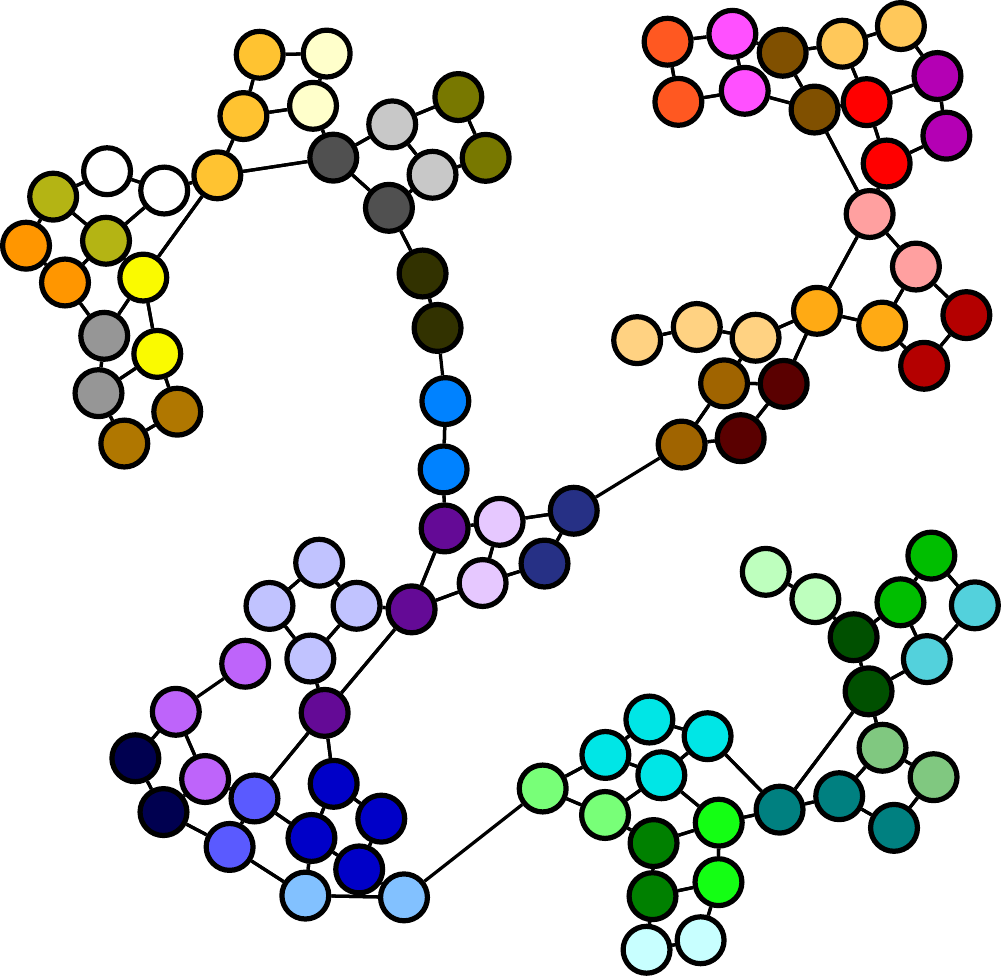}
			\caption*{Level 1}
        \end{subfigure}%
    \end{subfigure}%  
    
    \vspace{0.4cm}
    \begin{subfigure}{\linewidth}
        \centering
        \savebox{\imagebox}{\begin{subfigure}{0.27\linewidth}\includegraphics[width=\linewidth]{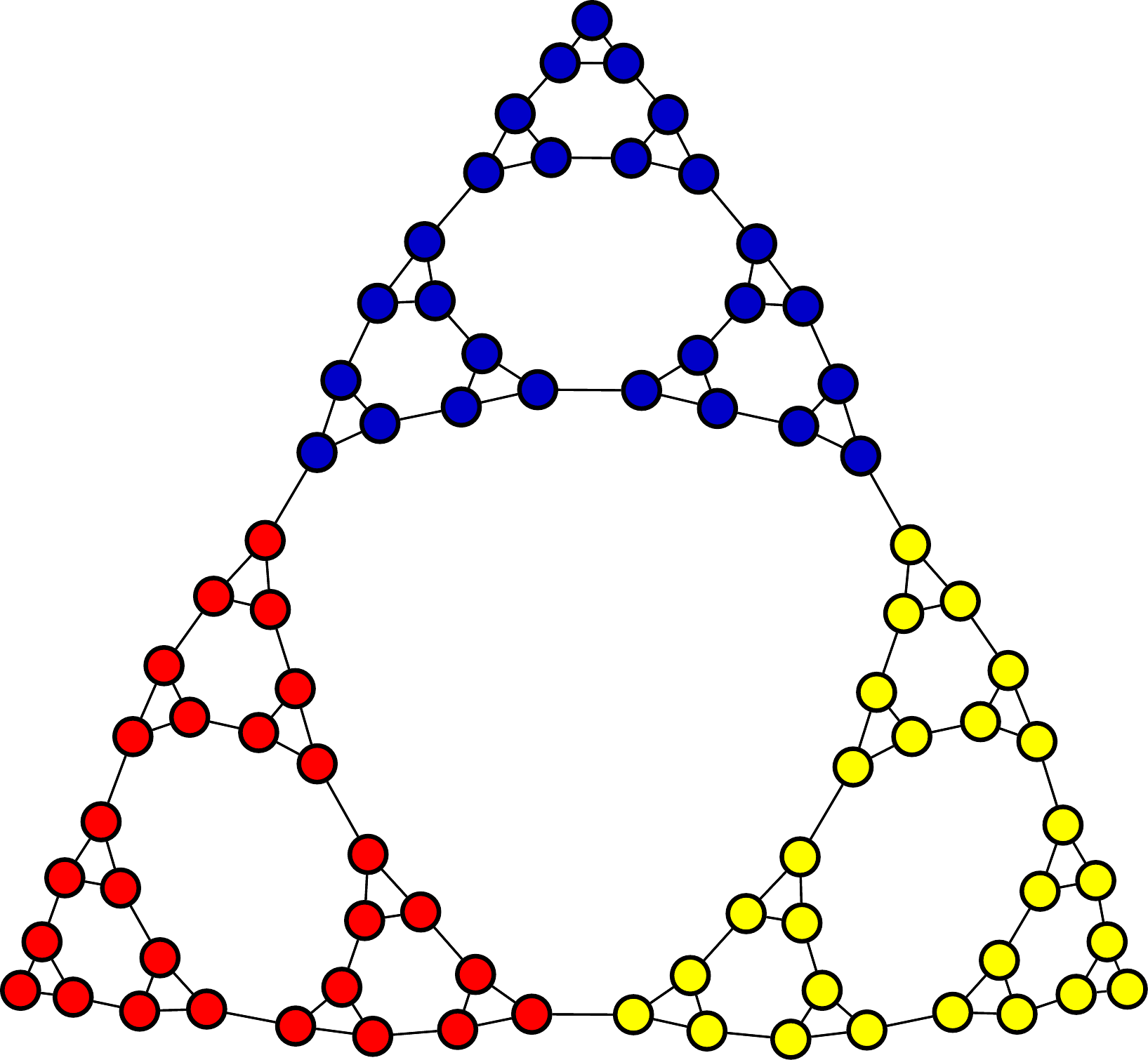}\end{subfigure}}% 
    	\begin{subfigure}{0.0175 \linewidth}
			\centering\raisebox{\dimexpr.5\ht\imagebox-.5\height+\baselineskip}{% Raise smaller image into place
            \includegraphics[width=\linewidth]{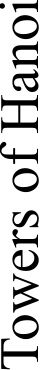}}%
		\end{subfigure}%
        \hspace{0.01\linewidth}%
        \begin{subfigure}{0.27\linewidth}
            \centering
           	\setlength{\abovecaptionskip}{2.0pt}
            \includegraphics[width=\linewidth]{"figures/hierarchy_examples/hanoi/hanoi_level_3".pdf}
			\caption*{Level 3}
        \end{subfigure}%
        \hfill
        \begin{subfigure}{0.27\linewidth}
            \centering
            \setlength{\abovecaptionskip}{2.0pt}
            \includegraphics[width=\linewidth]{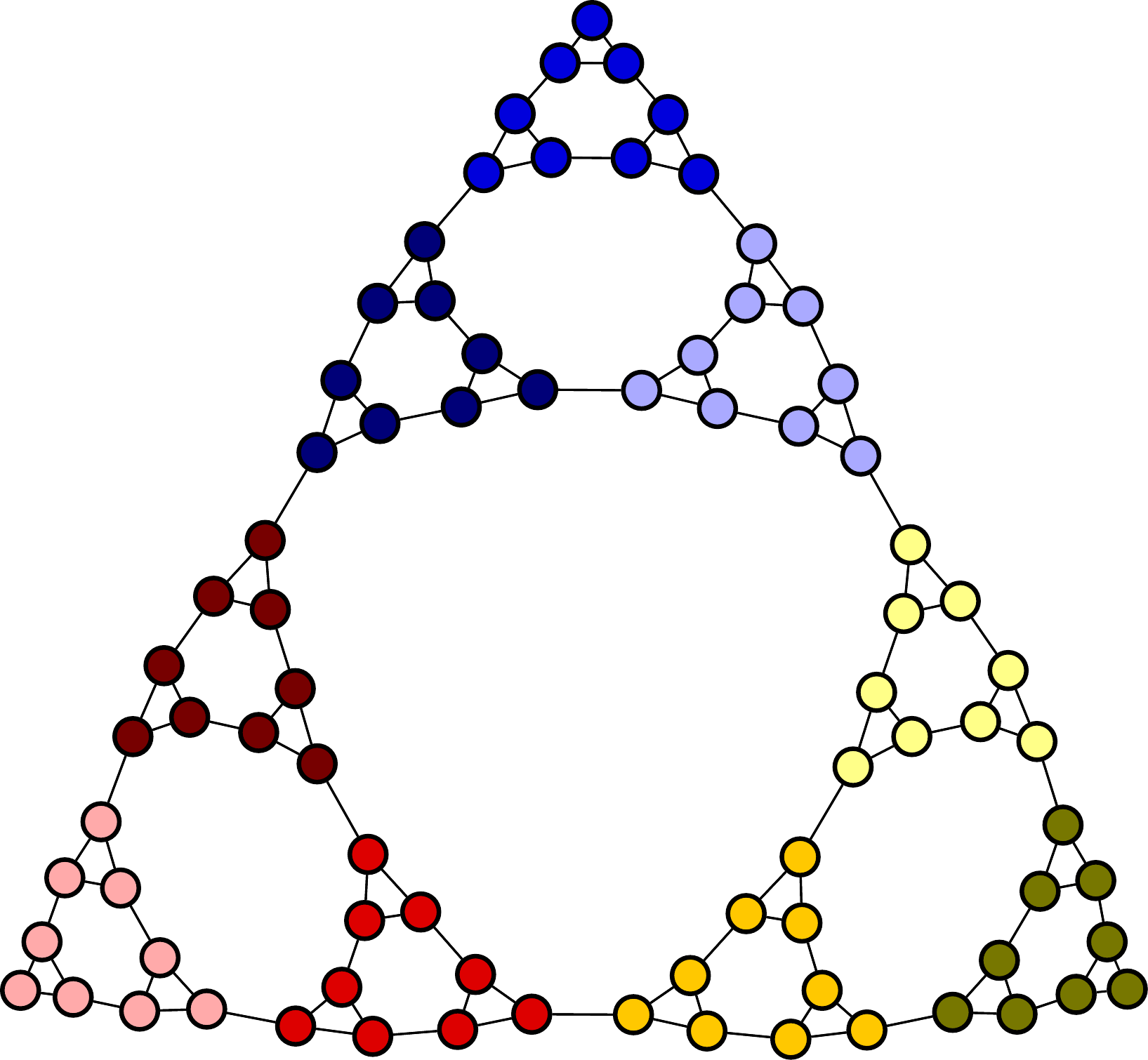}
            \caption*{Level 2}
        \end{subfigure}%
        \hfill
        \begin{subfigure}{0.27\linewidth}
            \centering
            \setlength{\abovecaptionskip}{2.0pt}
            \includegraphics[width=\textwidth]{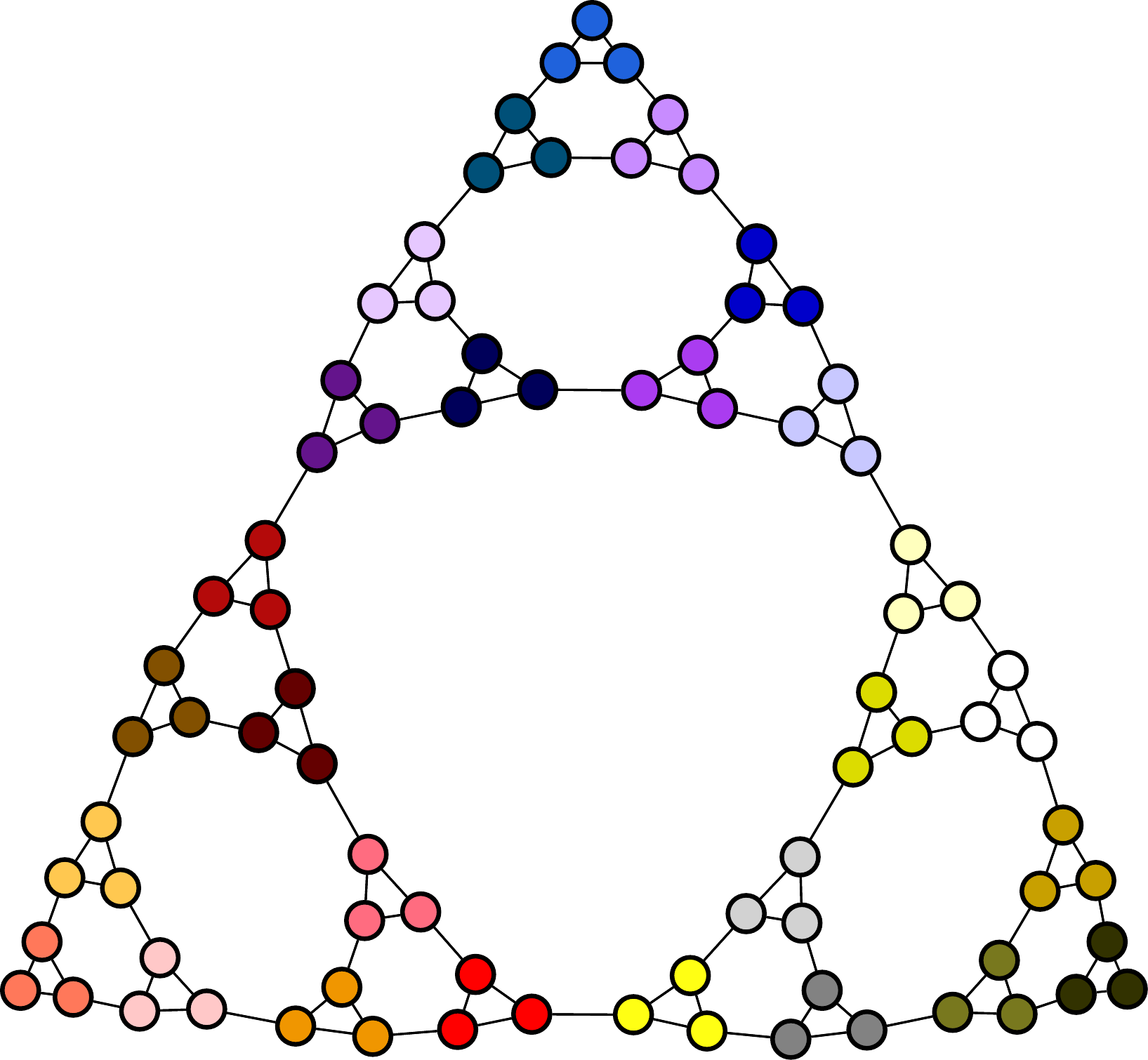}
			\caption*{Level 1}
        \end{subfigure}%
    \end{subfigure}%
    \caption{The cluster hierarchies produced by the Louvain algorithm when applied to the state transition graphs representing Rooms, Office, Taxi, and Towers of Hanoi. For Taxi and Towers of Hanoi, the graph layout was determined by using a force-directed algorithm that models nodes as charged particles that repel each other and edges as springs that attract connected nodes.}
\label{fig:skills}
%\vspace{-1cm}
\end{figure}

In Rooms, the hierarchy has four levels. At level three, we see that each room has been placed in its own cluster. Moving up the hierarchy, at level four, two of these rooms have been joined together into a single cluster. Moving down the hierarchy, each room has been divided into smaller clusters at level two, and then into even smaller clusters at level one. The figure illustrates how the corresponding skill hierarchy would enable efficient navigation between and within rooms.

In Office, at the top level, we see six large clusters connected by corridors. As we move lower down the hierarchy, we see that these large clusters have been divided into increasingly smaller regions. At level three, there are many rooms that form their own cluster. At level two, most rooms have been divided into multiple clusters. The figure reveals how the corresponding skill hierarchy would enable efficient navigation of the environment at multiple time scales.

In Taxi, the state transition graph has four disconnected components, each corresponding to one particular passenger destination: R, G, B, or Y. In Figure~\ref{fig:skills}, we show only one of these components, the one where the passenger destination is B. The results for the other components are similar. The Louvain cluster hierarchy has four levels. At the top level, we see four clusters. In three of these clusters, the passenger is waiting at their starting location (R, G, or Y). In the fourth cluster, the passenger is either in-taxi or has been delivered to their destination (B). Navigation between these clusters is unidirectional, with only three possibilities. The three corresponding skills navigate the taxi to the passenger location \emph{and} pick up the passenger. Moving one level down the hierarchy, the clusters produce skills that move the taxi between the left and the right-hand side of the grid, which are connected by the bottleneck state at the centre of the taxi navigation grid.

In Towers of Hanoi, the hierarchy has three levels. At level three, moving between the three clusters corresponds to moving the largest disk to a different pole. Each of these clusters has been divided into three smaller clusters at level two and then into three even smaller clusters at level one. A similar structure exists at levels two and one. The smaller clusters at level two correspond to moving the second-largest disc between different poles; the smallest clusters at level one correspond to moving the third-largest disc between different poles.

In all of these domains, the Louvain skill hierarchy closely matches human intuition. In addition, it is clear how skills at one level of the hierarchy can be composed to produce the skills at the next level.

\begin{figure}[t]
	\centering
	\begin{subfigure}{0.33 \linewidth}
		\centering
        \includegraphics[width=\linewidth]{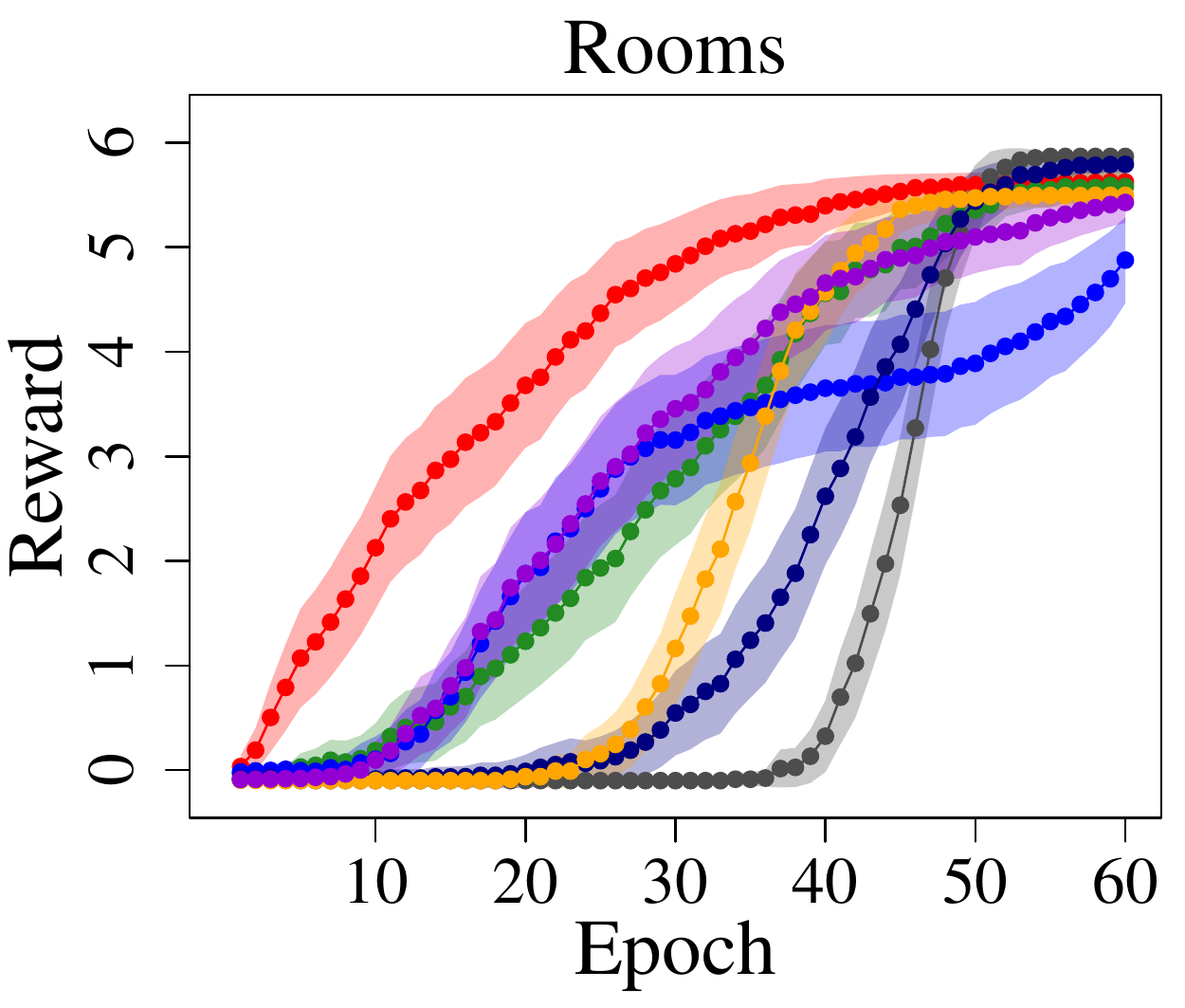}
	\end{subfigure}%
	\hfill
	\begin{subfigure}{0.33 \linewidth}
		\centering
		\includegraphics[width=\linewidth]{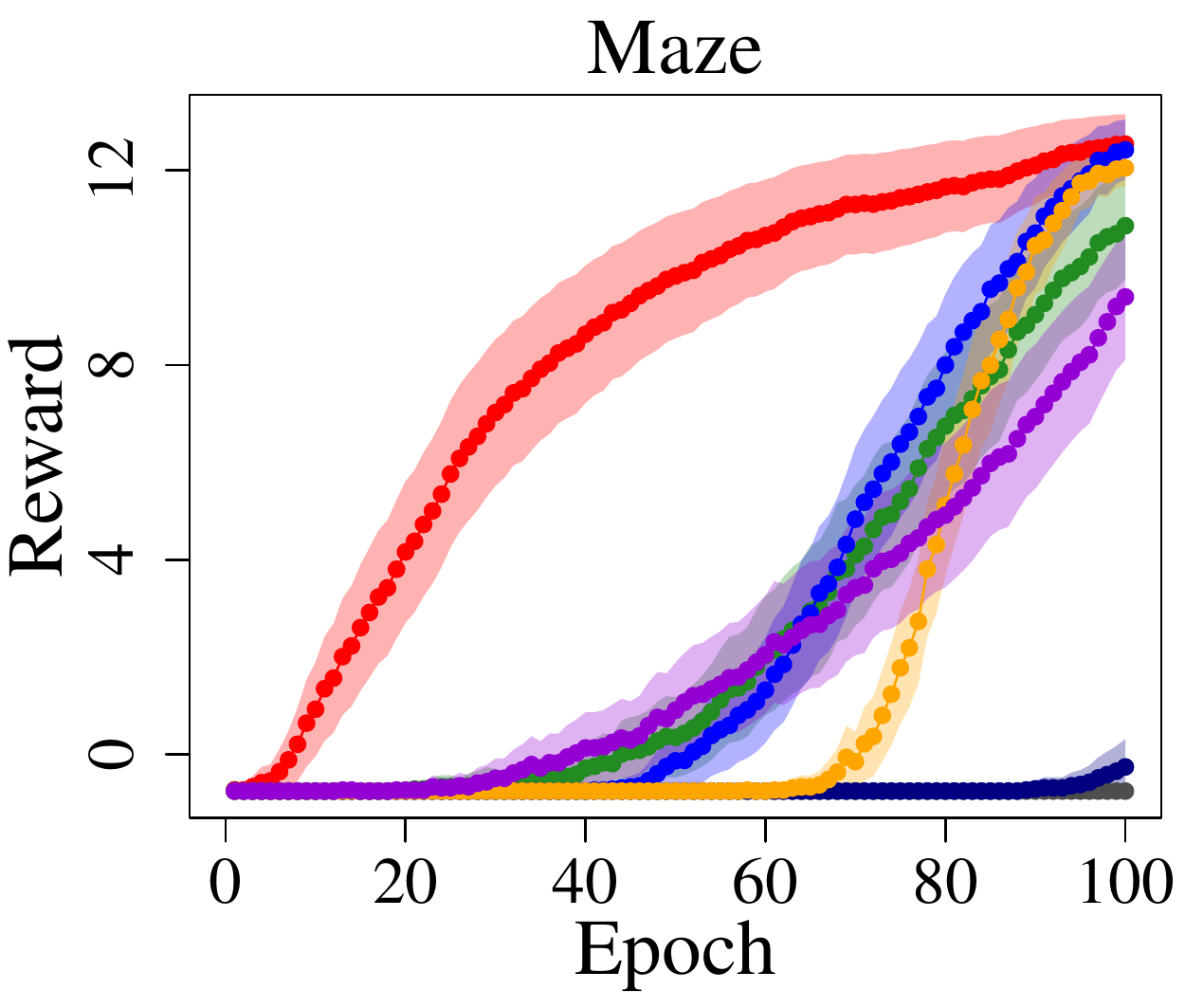}
	\end{subfigure}%	
	\hfill
	\begin{subfigure}{0.33 \linewidth}
		\centering
		\includegraphics[width=\linewidth]{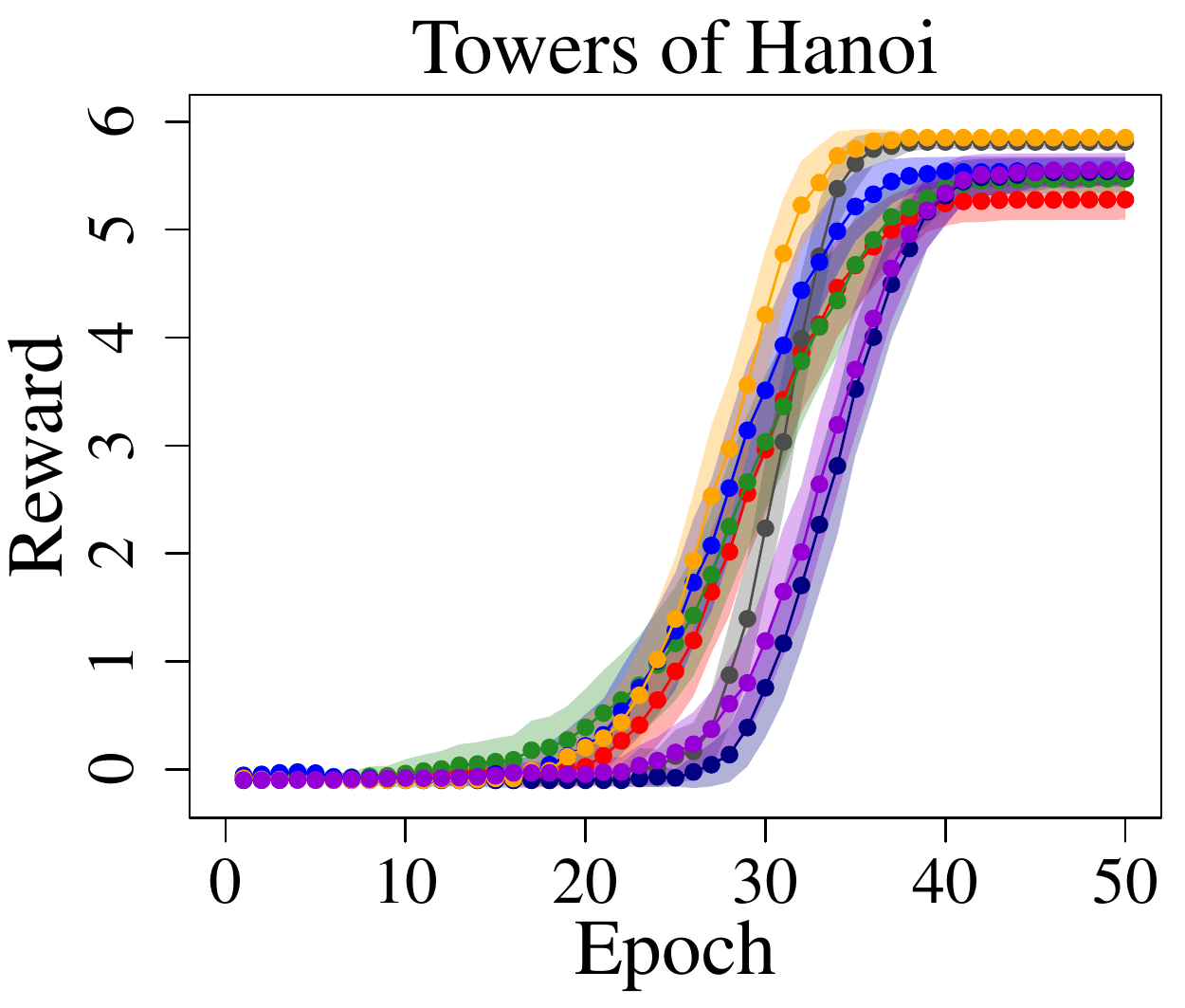}
	\end{subfigure}%

	\vspace{0.4cm}
	\begin{subfigure}{0.33 \linewidth}
		\centering
		\includegraphics[width=\linewidth]{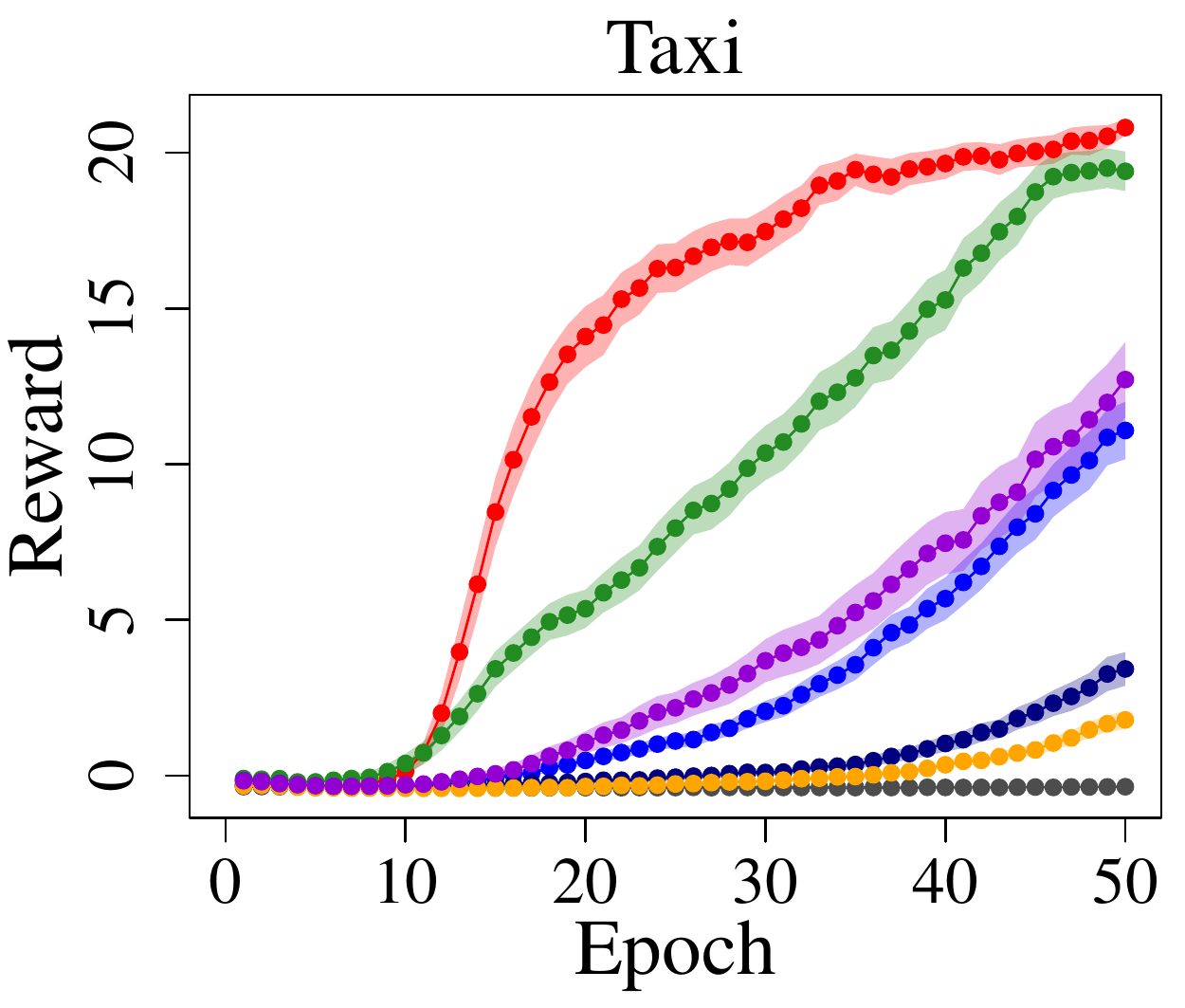}
	\end{subfigure}%	
	\hfill	
	\begin{subfigure}{0.33 \linewidth}
		\centering
		\includegraphics[width=\linewidth]{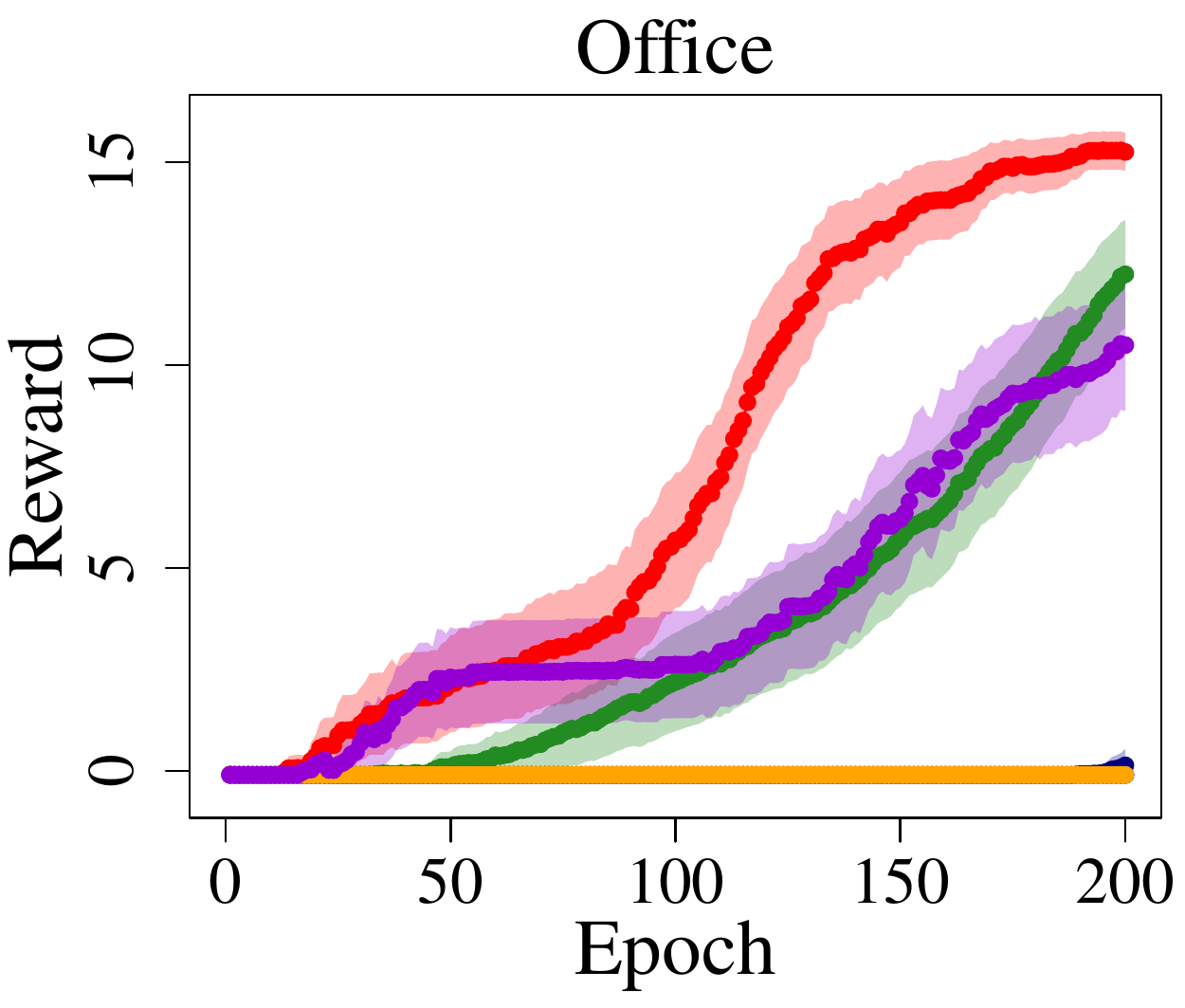}
	\end{subfigure}%	
	\hfill	
	\begin{subfigure}{0.33 \linewidth}
		\centering
		\includegraphics[width=\linewidth]{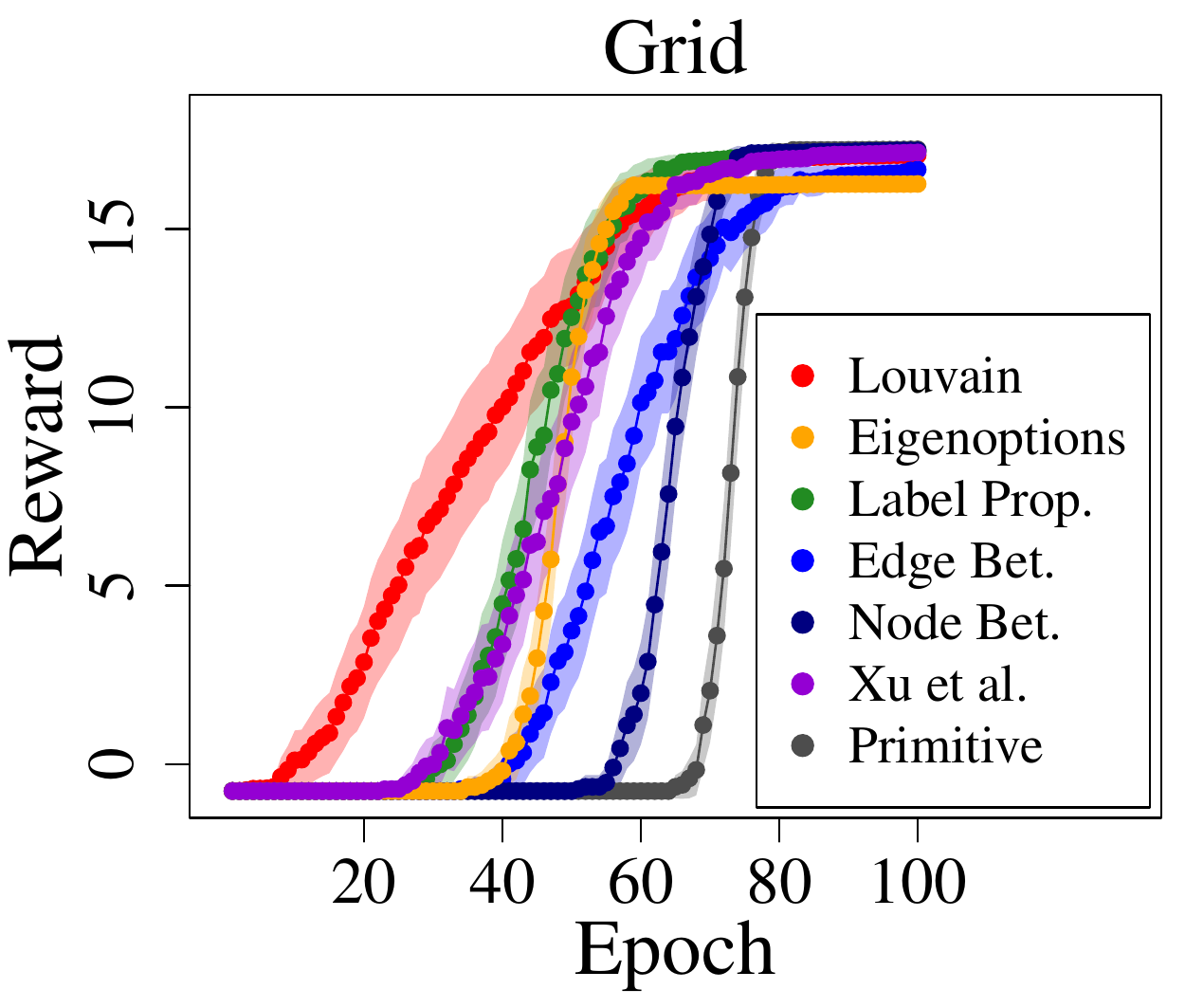}
	\end{subfigure}%	
	\caption{Learning performance. An epoch corresponds to 100 decision stages in Rooms and Towers of Hanoi, 300 in Taxi, 750 in Maze and Grid, and 1000 in Office.}
	\label{fig:Performance}
	%\vspace{-0.5cm}
\end{figure}

\textbf{Learning Performance.} We compare learning performance with the Louvain skill hierarchy to approaches based on \texttt{Edge Betweenness}~\citep{Davoodabadi2011} and \texttt{Label Propagation}~\citep{Davoodabadi2019}, as well as to the method proposed by \texttt{Xu~et~al.}~\citep{Xu2018}. These methods are the most directly related to the proposed approach, as discussed in Section~\ref{sect:related_work}. In addition, we compare to options that navigate to local maxima of \texttt{Node Betweenness}~\cite{Simsek2009}, a subgoal-based approach that captures the many different notions of the bottleneck concept in the literature. Similarly to the proposed approach, it is also one of the few approaches in the literature to explicitly characterise a target skill hierarchy for an agent to learn. We also compare to \texttt{Eigenoptions}~\cite{Machado2017}, which are derived from the Laplacian of the state transition graph and encourage efficient exploration. Finally, we include a \texttt{Primitive} agent that uses only primitive actions. Primitive actions are available to all agents.

We present learning curves in Figure \ref{fig:Performance}. The Louvain agent has a clear and substantial advantage over other approaches in all domains except for Towers of Hanoi, where its performance is much closer to that of the other hierarchical agents, with none of them performing much better than the primitive agent. This is consistent with existing results reported in this domain~(\eg by \citet{Jinnai2019}).

Section~\ref{sect:node_betweenness} of the supplementary material contains further analysis comparing Louvain skills and skills based on node betweenness.

\textbf{Scaling to larger domains.} We experimented with a multi-floor version of the Office environment, with floors connected by a central elevator, where two primitive actions move the agent up and down between two adjacent floors. The number of states in this domain can be varied by adjusting parameters such as the number of office floors. We use this environment to explore how the performance of the agent and the Louvain skill hierarchy change as the number of states in the environment increases.

\begin{figure}[b]
	\centering
	\begin{subfigure}{0.485 \linewidth}
		\setlength{\abovecaptionskip}{-1.5pt}
		\centering
		\includegraphics[width=\linewidth]{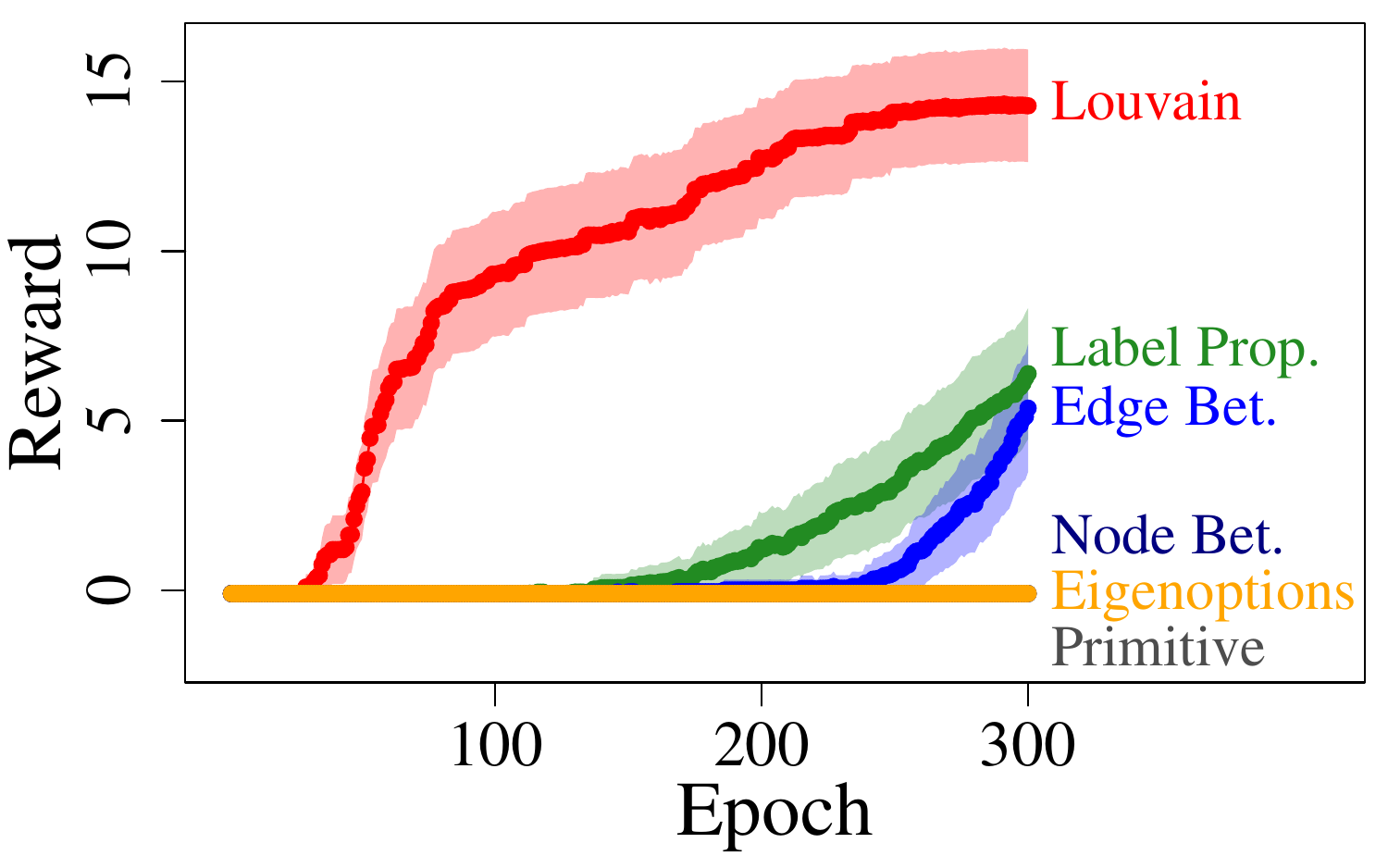}%
		\caption{}
		\label{fig:ScalingPerf}
	\end{subfigure}%
	\hfill
	\begin{subfigure}{0.485 \linewidth}
		\setlength{\abovecaptionskip}{-1.5pt}
		\centering
		\includegraphics[width=\linewidth]{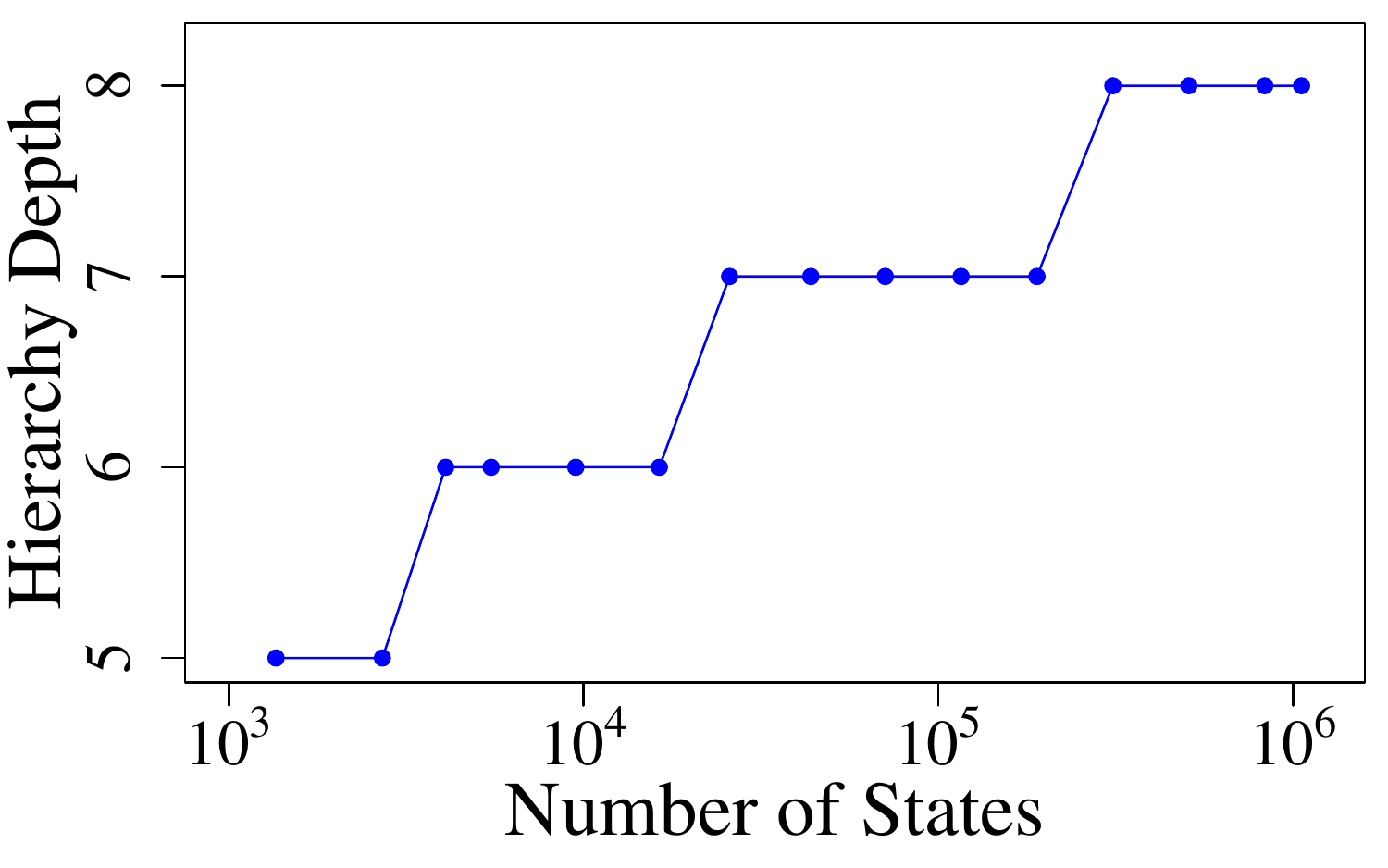}%
		\caption{}
		\label{fig:ScalingDepth}
	\end{subfigure}%
	\caption{(a) Learning performance in a two-floor Office containing \(2537\) states. (b) How the depth of the skill hierarchy scales with the size of the state space in multi-floor Office.} 
	\label{fig:Scaling}
	%\vspace{-0.3cm}
\end{figure}

Figure~\ref{fig:ScalingPerf} presents results in an Office environment with $2537$ states. It shows that the Louvain agent learns much more quickly than all other agents. Some alternatives, including the Eigenoptions agent, fail to achieve any learning.

Figure~\ref{fig:ScalingDepth} shows how the Louvain skill hierarchy changes with the size of the state space in the Office environment.  The figure shows the depth of the skill hierarchy in a series of fifteen Office environments of increasing size, ranging from a single floor (\({\sim}10^3\)~states) to one thousand floors (\({\sim}10^6\)~states). The depth of the hierarchy increased very gradually with the number of states in the environment. The maximum hierarchy depth reached was eight in the largest environment tested.

\begin{figure*}[t]
	\centering
	\begin{subfigure}{0.33 \linewidth}
		\centering
        \includegraphics[width=\linewidth]{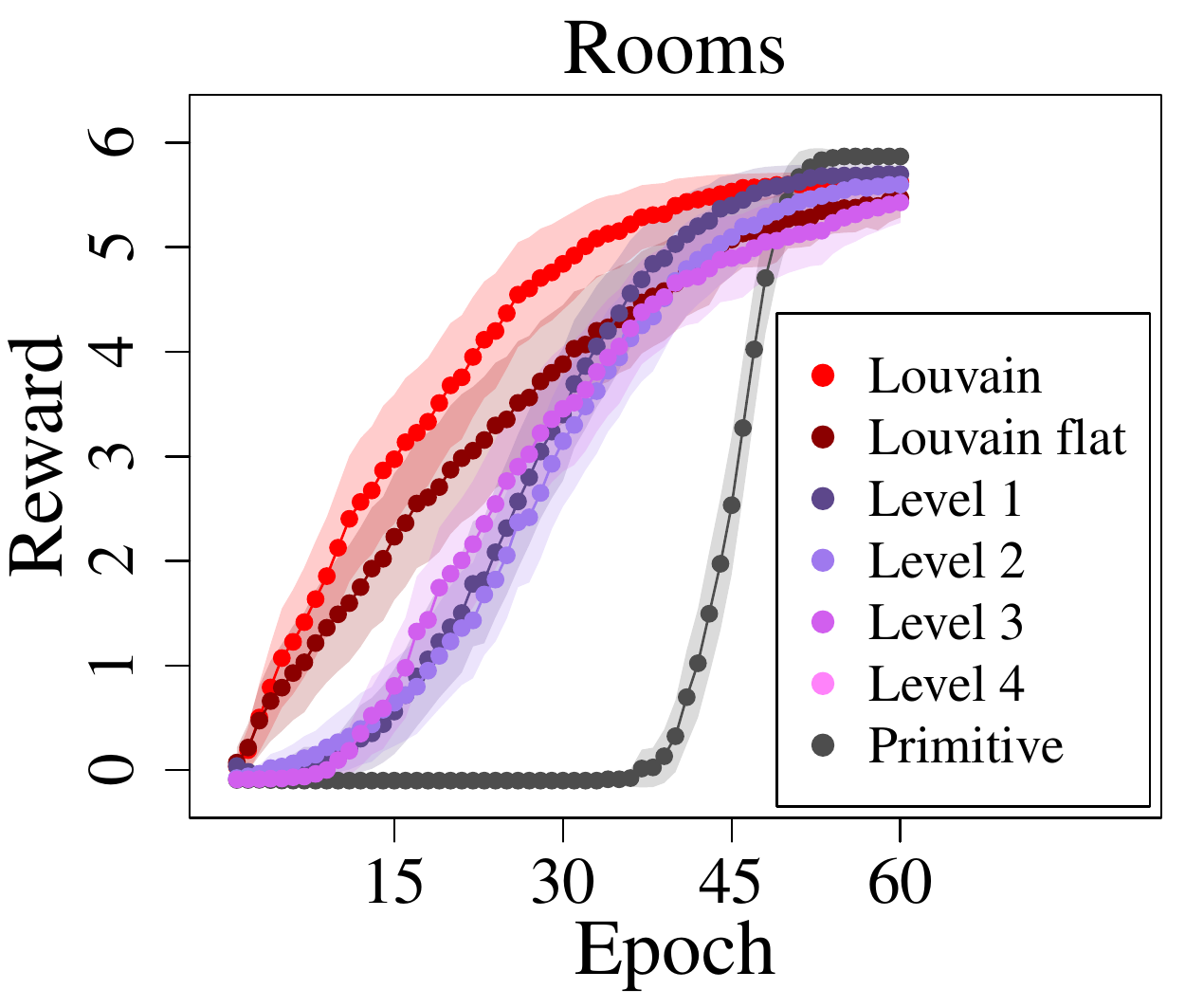}
	\end{subfigure}%
	\hfill
	\begin{subfigure}{0.33 \linewidth}
		\centering
		\includegraphics[width=\linewidth]{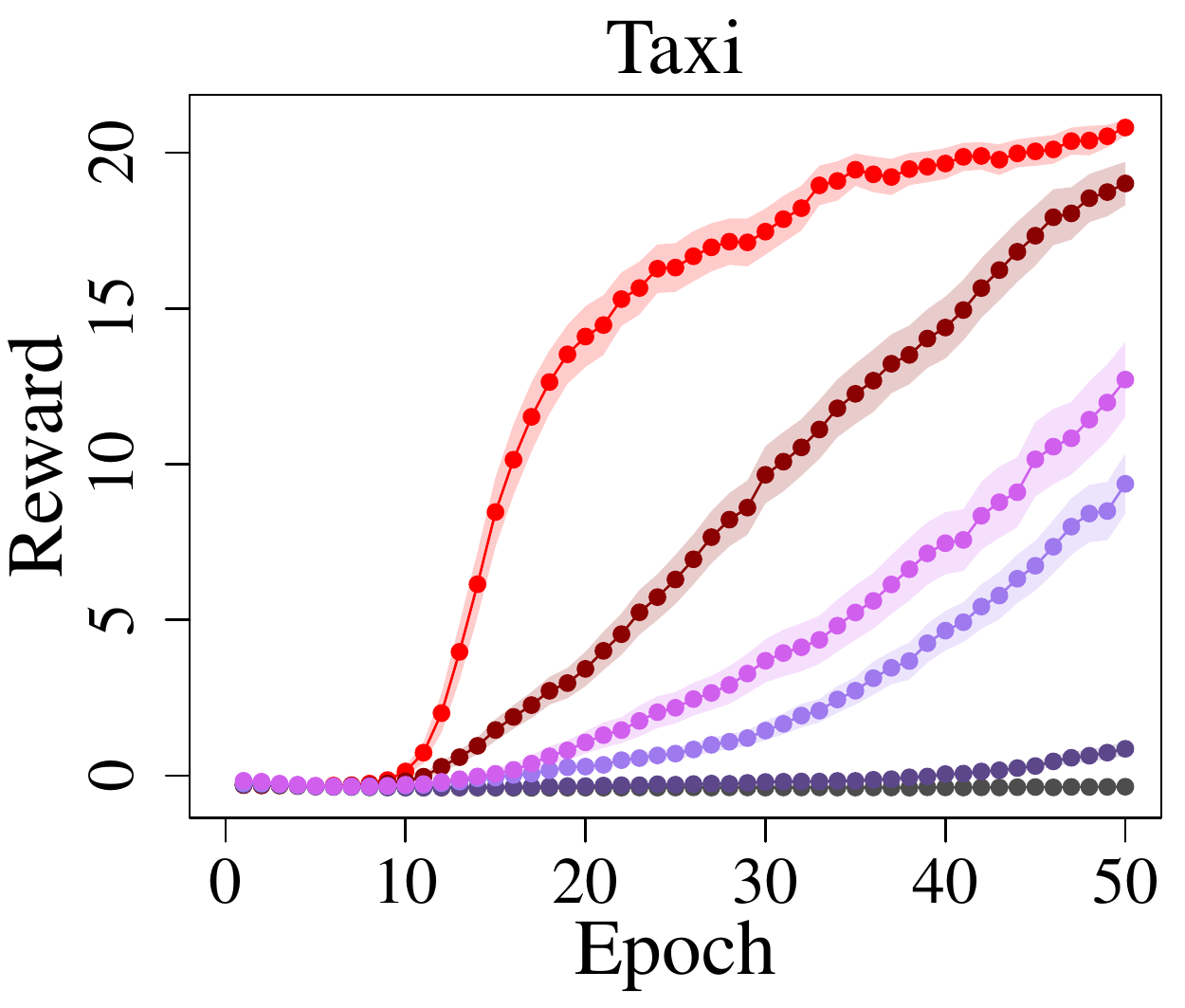}
	\end{subfigure}%
	\hfill
	\begin{subfigure}{0.33 \linewidth}
		\centering
		\includegraphics[width=\linewidth]{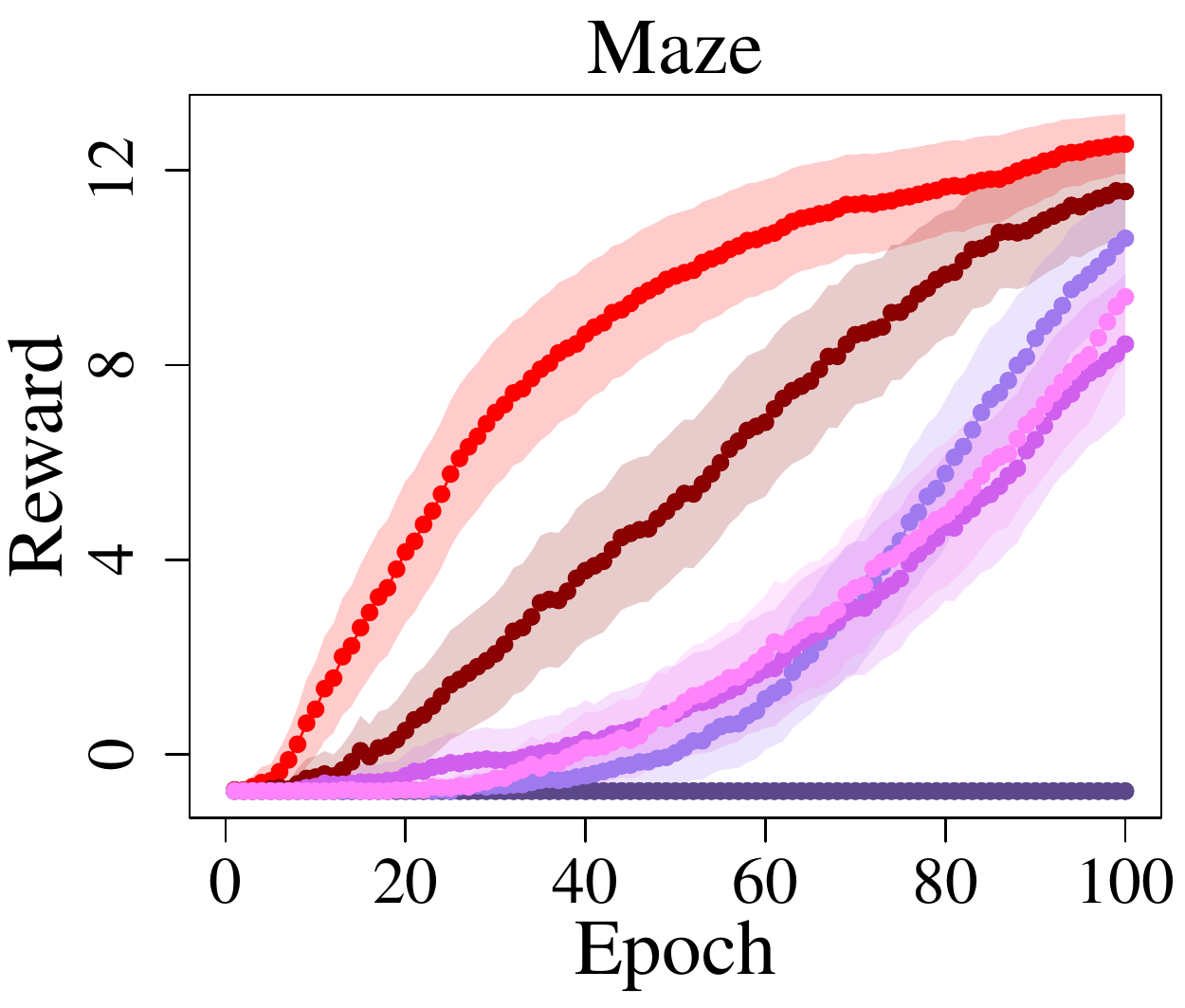}
	\end{subfigure}%
	\caption{Learning curves comparing various different Louvain agents. An epoch corresponds to 100 decision stages in Rooms, 300 in Taxi, and 750 in Maze. The skill hierarchy contained three levels in Rooms and Taxi, and four levels in Maze.}
	\label{fig:PerformanceAlt}
\end{figure*}

\textbf{Hierarchical versus flat arrangement of skills.} An alternative to the Louvain skill hierarchy is a flat arrangement of the skills, where each skill calls primitive actions directly rather than indirectly through other (lower-level) skills. We expected the hierarchical arrangement to lead to faster learning than the flat arrangement due to the additional learning updates enabled by the hierarchical relationship between the skills. Figure \ref{fig:PerformanceAlt} shows that this is indeed the case. In the figure, \texttt{Louvain} shows an agent that uses the Louvain skill hierarchy while \texttt{Louvain flat} shows an agent that uses the Louvain skills but where the skill policies call primitive actions directly rather than through other skills. In addition, the figure shows a number of agents that use only a single level of the Louvain hierarchy (\texttt{Level 1}, \texttt{Level 2}, \texttt{Level 3}, or \texttt{Level 4}), with option policies that call primitive actions directly. Primitive actions were available to all agents. The figure shows that the hierarchical agent learns more quickly than the flat agent. Furthermore, the agents using individual levels of the Louvain hierarchy learn more quickly than the primitive agent but not as quickly as the agent using the full Louvain hierarchy.

\begin{figure}[b]
	\hfill
	\begin{subfigure}{0.5\linewidth}
		\centering
		\includegraphics[width=0.75\linewidth]{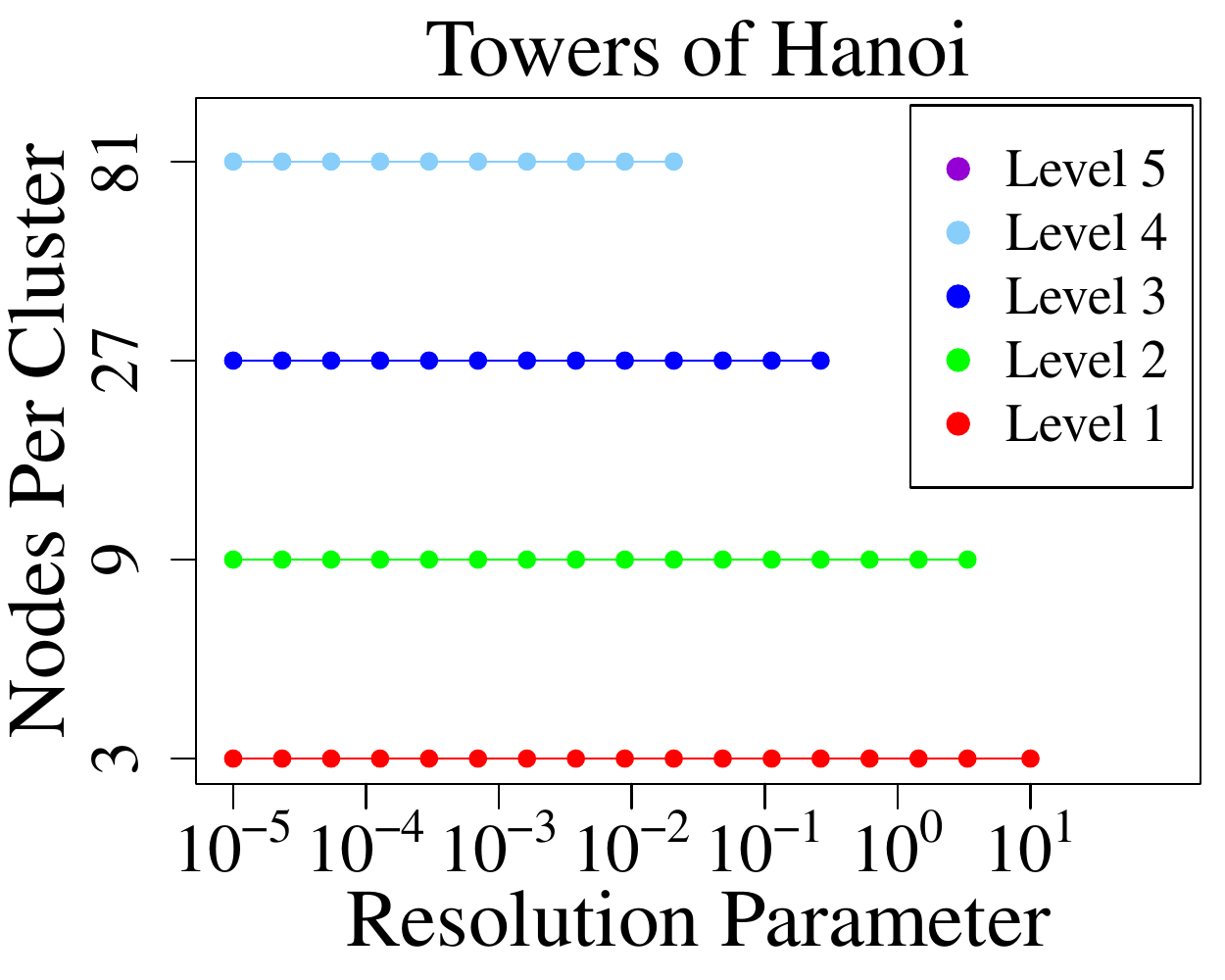}
	\end{subfigure}%
	\hfill
	\begin{subfigure}{0.5\linewidth}
		\centering
		\includegraphics[width=0.75\linewidth]{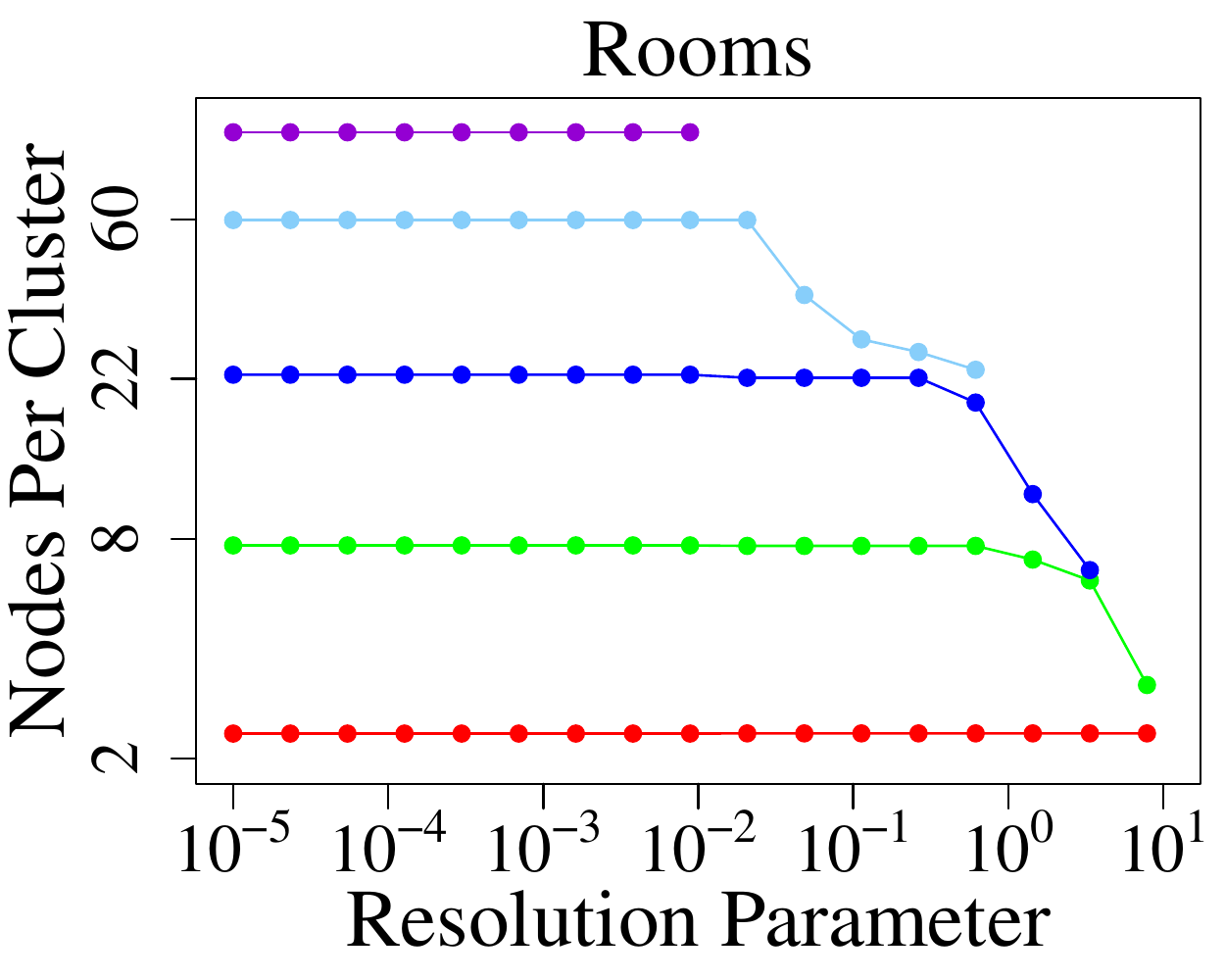}
	\end{subfigure}%
	\hfill
	\caption{How the Louvain algorithm's output when applied to Towers of Hanoi and Rooms changes with the resolution parameter. The cluster hierarchy had a maximum of four levels in Towers of Hanoi and five levels in Rooms.}%
	\label{fig:ResolutionNumClusters}
\end{figure}

\textbf{Impact of the resolution parameter.} When using very high values of the resolution parameter \(\rho\), the Louvain algorithm terminates without producing any partitions. On the other hand, when using very low values of \(\rho\), it runs until a partition is produced where all nodes are merged into a single cluster.

Figure~\ref{fig:ResolutionNumClusters} shows how changing the resolution parameter \(\rho\) impacts the Louvain cluster hierarchy in Towers of Hanoi and Rooms.  
In Towers of Hanoi, at \(\rho=10\), the output is a single level containing many small clusters comprised of three nodes. As \(\rho\) gets smaller, the cluster hierarchy remains the same until \(\rho=3.3\), at which point a second level is added. The first level remains identical to the single level produced at \(\rho=10\). The second level contains larger clusters, each formed by merging three of the clusters from the first level. As \(\rho\) is reduced further, additional levels are added to the hierarchy. When a new level is added to the hierarchy, the existing levels generally remain the same. 

A sensitivity analysis on the value of \(\rho\) showed that a wide range of \(\rho\) values led to useful skills, and that performance gradually decreased to no worse than that of a primitive agent at higher values of \(\rho\). Section~\ref{sect:resolution_sensitivity} of the supplementary material contains full details of this sensitivity analysis and a more detailed discussion on the choice of \(\rho\).

\section{Discussion and Future Work}
\label{sect:discussion}

\begin{figure}[t]
	\centering
	\begin{subfigure}{0.35 \linewidth}
		\setlength{\abovecaptionskip}{-1.5pt}
		\centering
		\includegraphics[width=\linewidth]{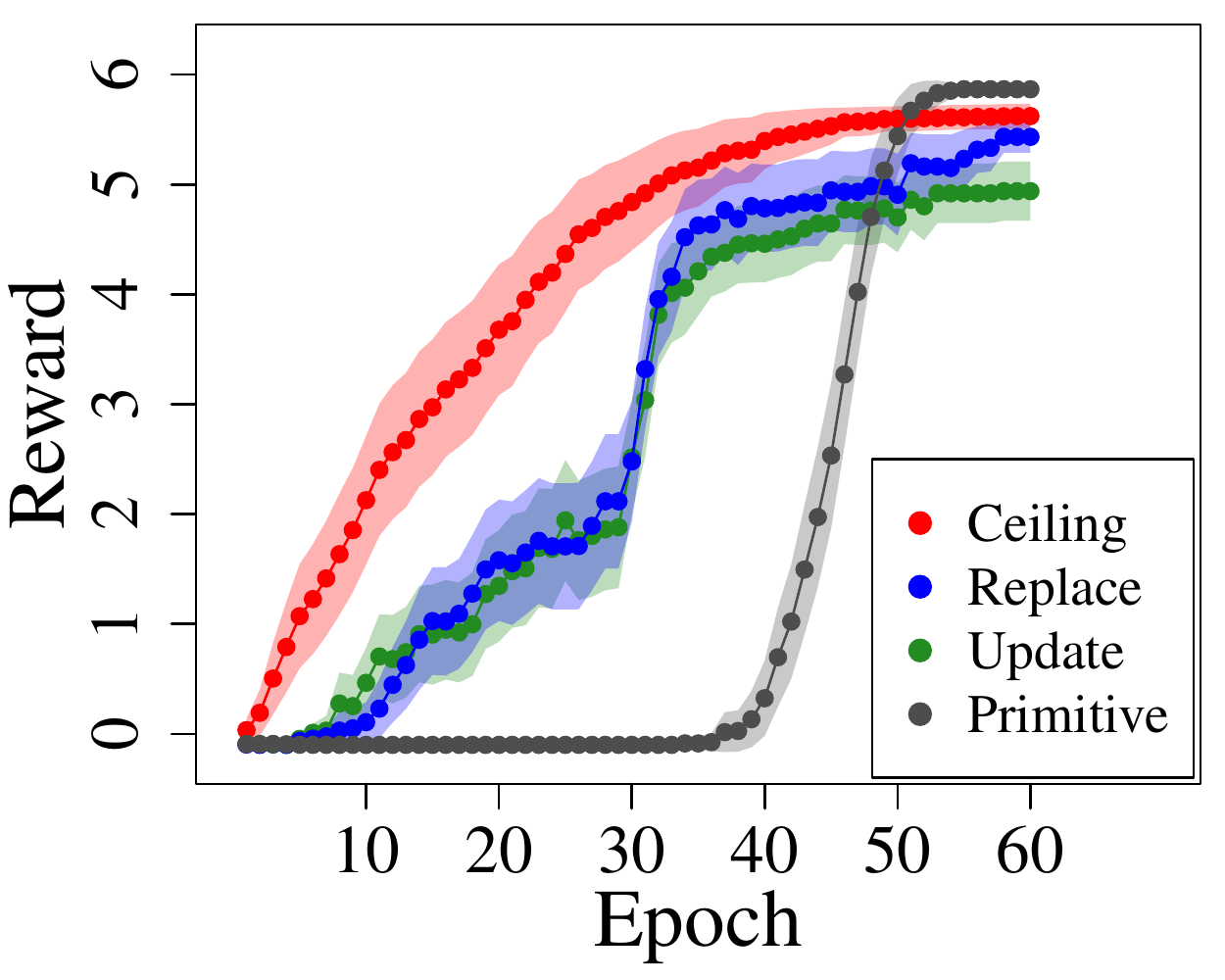}
		\caption{Incremental~Performance}
		\label{fig:FourRoomsIncrementalPerformance}
	\end{subfigure}%
	\hfill
	\begin{subfigure}{0.19 \linewidth}% 0.173
		\centering
		\includegraphics[width=\linewidth]{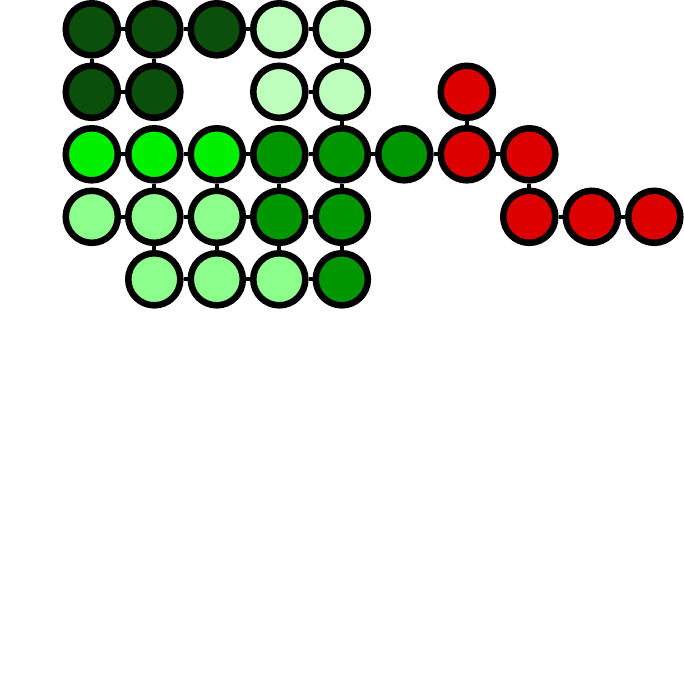}
		\caption{30~States}
		\label{fig:FourRoomsIncrementalSTG1}
	\end{subfigure}%
	\hfill
	\begin{subfigure}{0.19 \linewidth}
		\centering
		\includegraphics[width=\linewidth]{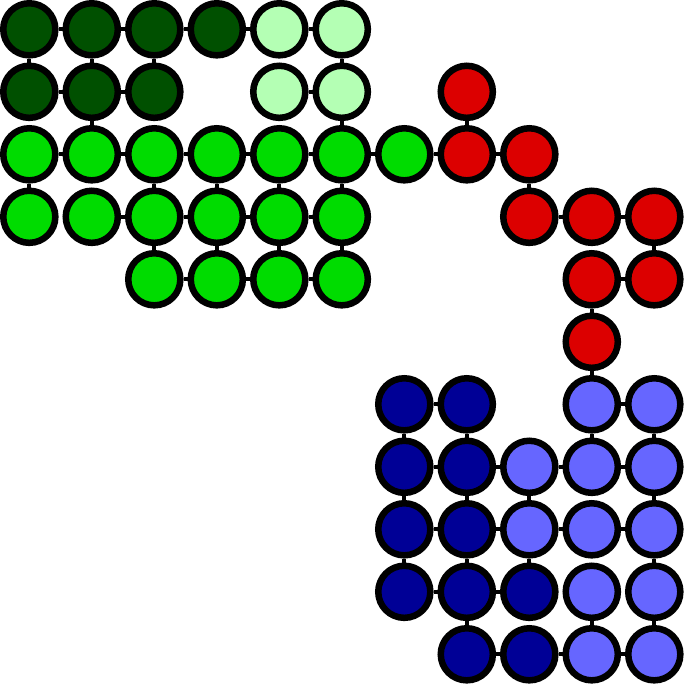}
		\caption{60~States}
		\label{fig:FourRoomsIncrementalSTG2}
	\end{subfigure}%
	\hfill
	\begin{subfigure}{0.19 \linewidth}
		\centering
		\includegraphics[width=\linewidth]{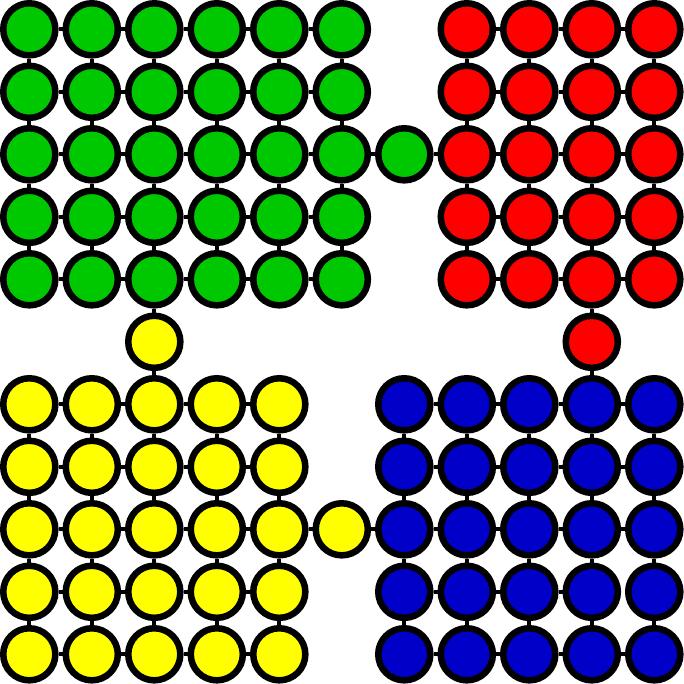}
		\caption{All~States}
		\label{fig:FourRoomsIncrementalSTG3}
	\end{subfigure}%
	\caption{(a) Performance when incrementally discovering Louvain options in Rooms. An epoch corresponds to \(100\) decision stages. (b-d) How the state transition graph and top-level partitions produced by the \texttt{Update} approach evolved as an agent explored Rooms. The hierarchy contained 2 levels after visiting 30 states, 3 levels after visiting 60 states, and 5 levels after observing all possible transitions.}
	\label{fig:incremental}
	\vspace{-0.4cm}
\end{figure}

An important research direction for future work is incremental learning of Louvain skill hierachies as the agent interacts with its environment---because the state transition graph will not always be available in advance. We explored the feasibility of incremental learning in the Rooms environment. The results are shown in Figure~\ref{fig:incremental}. The agent started with an empty state transition graph and no skills. Every \(m\) decision stages, it updated its state transition graph with new nodes and edges, and it revised its skill hierarchy using one of two approaches: \texttt{Replace} or \texttt{Update}.

In the first approach (\texttt{Replace}), the agent applied the Louvain algorithm to create a new skill hierarchy from scratch. In the second approach (\texttt{Update}), the agent incorporated the new information into its existing skill hierarchy, using an approach similar to the Louvain algorithm. This second approach starts by assigning each new node to its own cluster; the new nodes are then iteratively moved locally, between neighbouring clusters (both new and existing) until no modularity gain is possible. This revised partition is used to define an aggregate graph and the entire process is repeated on the aggregate graph, and the next aggregate graph, and so on, until an iteration is reached with no modularity gain. The cluster membership of existing nodes stays fixed; only new nodes have their cluster membership updated. The result of both approaches is a revised set of partitions, from which a revised hierarchy of Louvain skills is derived. Section~\ref{sect:incremental_psuedocode} of the supplementary material presents pseudocode for the two incremental approaches.

Figure \ref{fig:FourRoomsIncrementalPerformance} shows the performance of the incremental agents. The agents started with only primitive actions; after decision stages \(100\), \(500\), \(1000\), \(3000\), and \(5000\), the state transition graph was updated and the skill hierarchy was revised. The figure compares performance to a \texttt{Primitive} agent and an agent using the full Louvain skill hierarchy, whose performance acts as a \texttt{Ceiling} for the incremental agents. Both incremental agents learned much faster than the primitive agent and only marginally slower than the fully-informed Louvain agent. The two incremental agents had similar performance throughout training but \texttt{Replace} reached a higher level of asymptotic performance than \texttt{Update}, as expected.
The reason is that partitions produced early in training are based on incomplete information; \texttt{Replace} discards these early imperfections while \texttt{Update} carries them forward. But there is a trade-off: building a new skill hierarchy from scratch has a higher computational cost than updating an existing one.

Figures \ref{fig:FourRoomsIncrementalSTG1}--\ref{fig:FourRoomsIncrementalSTG3} show the evolution of the partitions as the \texttt{Update} agent performed a random walk in Rooms. The partitions were updated after observing \(30\) states, \(60\) states, and all possible transitions. The figure shows that, as more nodes were added, increasingly higher-level skills were produced.

These results demonstrate the feasibility of learning Louvain skills incrementally. A full incremental method for learning Louvain skills may take many forms, and different approaches may be useful under different circumstances, with each having its own strengths and weaknesses. We leave the full development of such algorithms to future work.

\begin{figure}[t]
	\centering
	\begin{subfigure}[b]{0.23 \linewidth}
		\caption*{Pinball Layout}
		\centering
		\includegraphics[width=\linewidth]{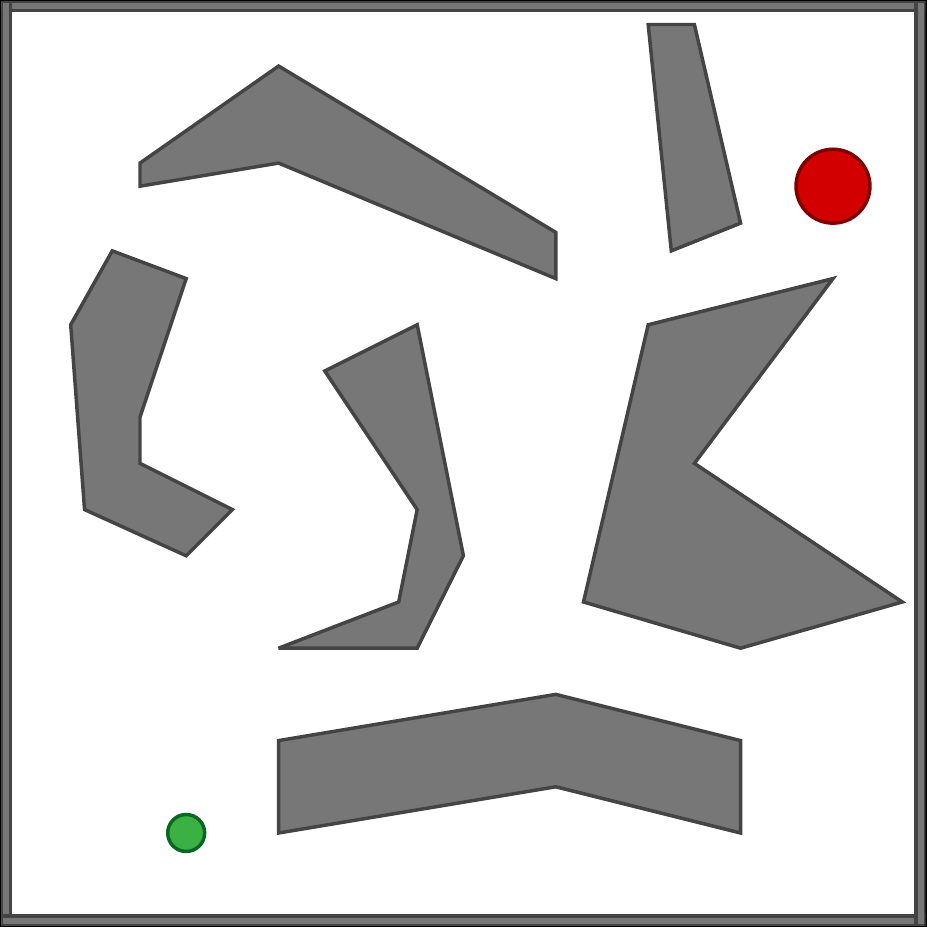}
	\end{subfigure}%
	\hfill
	\begin{subfigure}[b]{0.23 \linewidth}
		\centering
		\caption*{Level 3}
		\includegraphics[width=\linewidth]{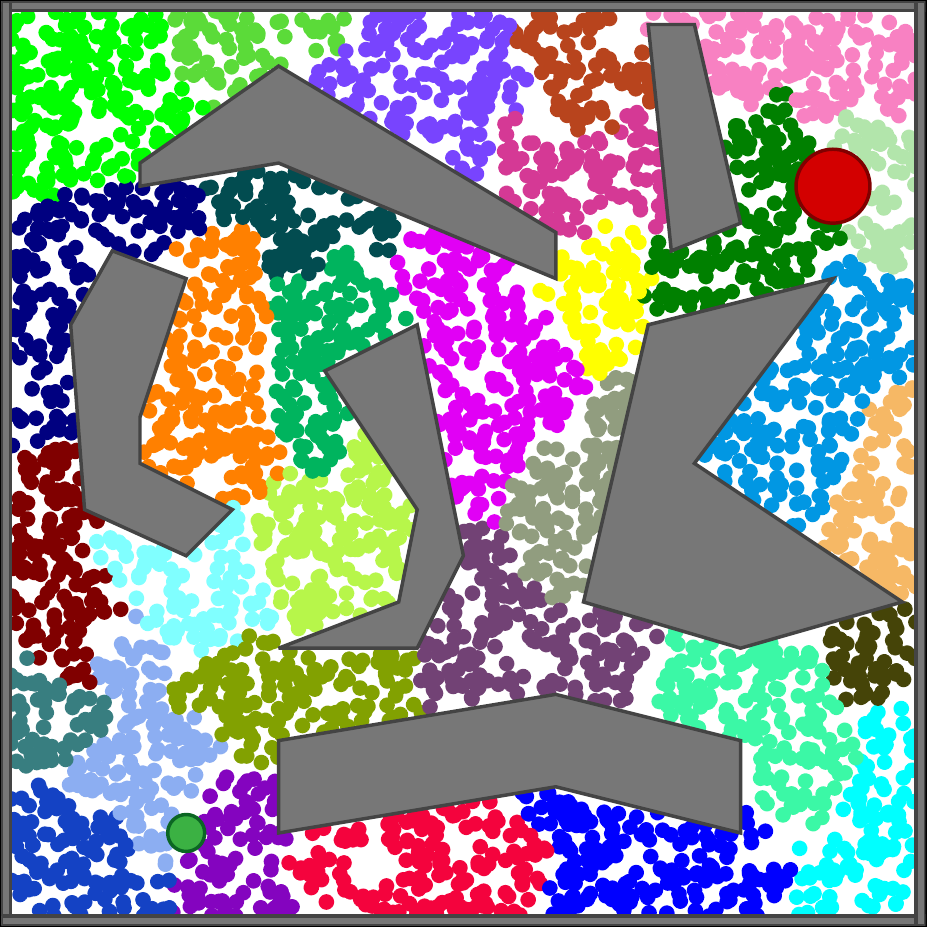}
	\end{subfigure}%
	\hfill
	\begin{subfigure}[b]{0.23 \linewidth}
		\centering
		\caption*{Level 2}
		\includegraphics[width=\linewidth]{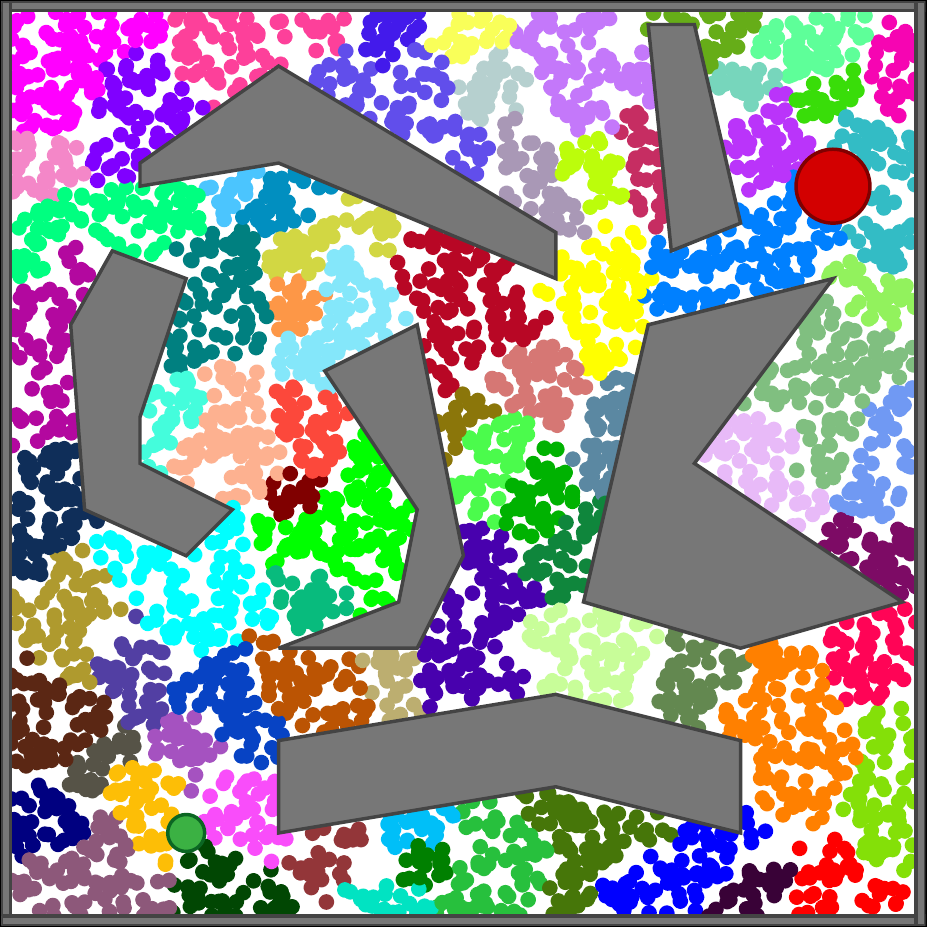}
	\end{subfigure}%
	\hfill
	\begin{subfigure}[b]{0.23 \linewidth}
		\centering
		\caption*{Level 1}
		\includegraphics[width=\linewidth]{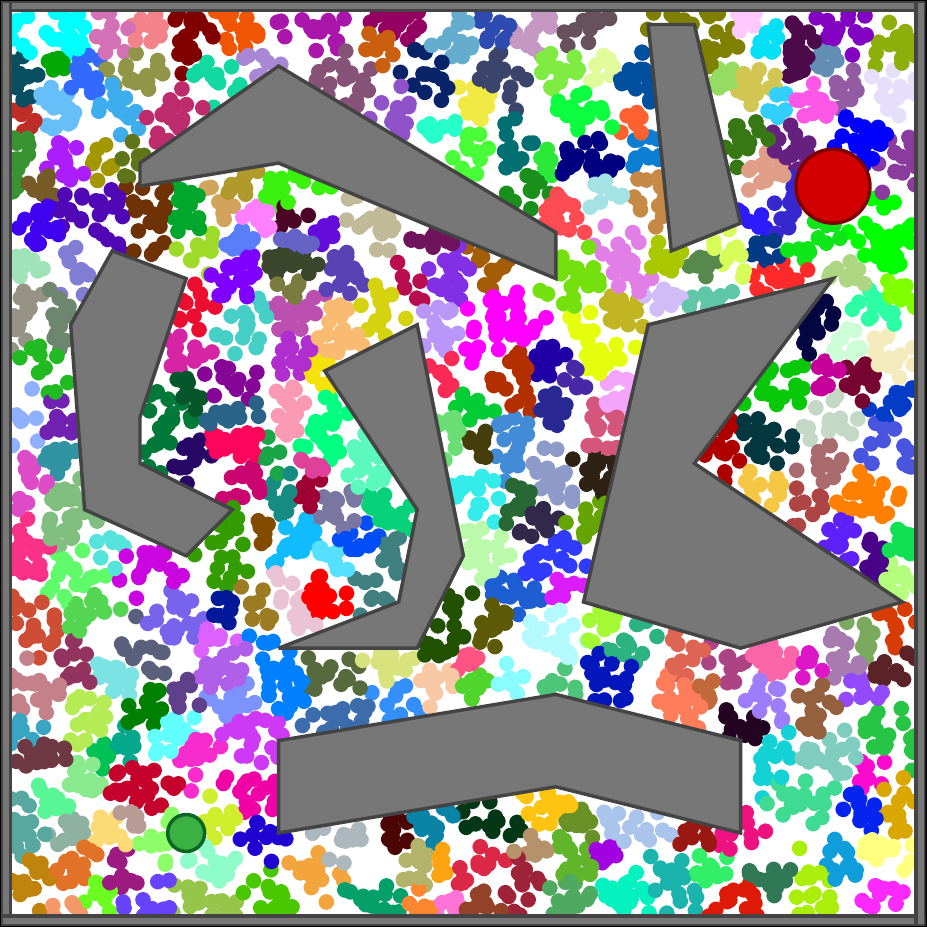}
	\end{subfigure}%
	\caption{The layout of the Pinball environment---with the green ball in its initial position, the red goal, and several obstacles---and the Louvain cluster hierarchy produced by the Louvain algorithm.}
	\label{fig:Pinball}
	%\vspace{-0.5cm}
\end{figure}

Another direction for future work is extending Louvain skills to environments with continuous state spaces such as robotic control tasks. Such domains present a difficulty to all skill discovery methods that use the state transition graph due to the inherently discrete nature of the graph. If the critical step of constructing an appropriate graphical representation of a continuous state space can be achieved, all graph-based methods would benefit. Some approaches have already been proposed in the literature. We use one such approach~\cite{Mahadevan2007,Bacon2013,Shoeleh2017} to examine the Louvain hierarchy in a variant of the Pinball domain~\cite{Konidaris2009}, which involves manoeuvring a ball around a series of obstacles to reach a goal, as shown in the leftmost panel of Figure~\ref{fig:Pinball}. The state is represented by two continuous values: the ball's position in the horizontal and vertical directions. At each decision stage, the agent can choose to apply a small force to the ball in each direction. The amount of force applied is stochastic and causes the ball to roll until friction causes it to come to a rest. Collisions between the ball and the obstacles are elastic.

We sampled \(4000\) states, added them to the state transition graph, then added an edge between each node and its \(k\)-nearest neighbours according to Euclidean distance, assigning an edge weight of \(e^{-4d^{2}}\) to an edge that connects two locations \(u\) and \(v\) with Euclidean distance \(d\) between them. We then applied the Louvain algorithm to the resulting graph. The result was the cluster hierarchy shown in Figure~\ref{fig:Pinball}. It had three levels. 
Moving between clusters in the top level corresponds to skills that enable high-level navigation of the state space, and take into account features such as the natural bottlenecks caused by the obstacles, allowing the agent to efficiently change its position.
Moving between clusters in the lower-level partitions enable more local navigation.

% Limitations
Currently, Louvain skills at one level of the hierarchy are composed of skills from only the previous level. Such skills may not provide the most efficient navigation between clusters. Future work could consider composing skills from all lower levels, including primitive actions.

Because Louvain skills are derived from the connectivity of the state transition graph, we expect them to be suitable for solving a range of problems in a given environment. Examining their use for transfer learning is a useful direction for future work.

A possible difficulty with building multi-level skill hierarchies is that having a large number of skills can end up hurting performance by increasing the branching factor. One solution that has been explored in the context of two-level hierarchies is pruning less useful skills from the hierarchy \cite{Taghizadeh2013, Davoodabadi2020}. Further research is needed to address this potential difficulty.

A key open problem for graph-based skill discovery methods generally is how to best make use of state abstraction.
We suggest that the most natural place to introduce state abstraction is when constructing the state transition graph itself. Instead of a concrete state, each node could represent an abstract state based on some learned representation of the environment.
The proposed method---and any other existing graph-based method---could then directly use this abstract state transition graph to define a skill hierarchy.

Lastly, we note that the various characterisations of useful skills proposed in the literature, including the one proposed here, are not necessarily competitors. To solve complex tasks, it is likely that an agent will need to use many different types of skills. An important avenue for future work is exploring how different ideas on skills discovery can be used together to enable agents that can autonomously develop complex and varied skill hierarchies.

% Acknowlegdements.
\begin{ack}
We would like to thank the members of the Bath Reinforcement Learning Laboratory for their constructive feedback.
This research was supported by the Engineering and Physical Sciences Research Council [EP/R513155/1] and the University of Bath.
This research made use of Hex, the GPU Cloud in the Department of Computer Science at the University of Bath.
\end{ack}

\bibliography{bibliography}

\begin{thebibliography}{42}
\providecommand{\natexlab}[1]{#1}
\providecommand{\url}[1]{\texttt{#1}}
\expandafter\ifx\csname urlstyle\endcsname\relax
  \providecommand{\doi}[1]{doi: #1}\else
  \providecommand{\doi}{doi: \begingroup \urlstyle{rm}\Url}\fi

\bibitem[Sutton et~al.(1999{\natexlab{a}})Sutton, Precup, and
  Singh]{Sutton1999}
R.~S. Sutton, D.~Precup, and S.~Singh.
\newblock Between {MDPs} and {semi-MDPs}: A framework for temporal abstraction
  in reinforcement learning.
\newblock \emph{Artificial Intelligence}, 112\penalty0 (1-2):\penalty0
  181--211, 1999{\natexlab{a}}.

\bibitem[Precup(2000)]{Precup2000}
D.~Precup.
\newblock \emph{Temporal abstraction in reinforcement learning}.
\newblock PhD thesis, University of Massachusetts Amherst, 2000.

\bibitem[Newman and Girvan(2004)]{Newman2004}
M.~E. Newman and M.~Girvan.
\newblock Finding and evaluating community structure in networks.
\newblock \emph{Physical Review E}, 69\penalty0 (2):\penalty0 026113, 2004.

\bibitem[Leicht and Newman(2008)]{Leicht2008}
E.~A. Leicht and M.~E. Newman.
\newblock Community structure in directed networks.
\newblock \emph{Physical Review Letters}, 100\penalty0 (11):\penalty0 118703,
  2008.

\bibitem[Arenas et~al.(2007)Arenas, Duch, Fern{\'a}ndez, and
  G{\'o}mez]{Arenas2007}
A.~Arenas, J.~Duch, A.~Fern{\'a}ndez, and S.~G{\'o}mez.
\newblock Size reduction of complex networks preserving modularity.
\newblock \emph{New Journal of Physics}, 9\penalty0 (6):\penalty0 176, 2007.

\bibitem[Brandes et~al.(2006)Brandes, Delling, Gaertler, G{\"o}rke, Hoefer,
  Nikoloski, and Wagner]{Brandes2006}
U.~Brandes, D.~Delling, M.~Gaertler, R.~G{\"o}rke, M.~Hoefer, Z.~Nikoloski, and
  D.~Wagner.
\newblock Maximizing modularity is hard.
\newblock \emph{arXiv preprint:physics/0608255}, 2006.

\bibitem[Blondel et~al.(2008)Blondel, Guillaume, Lambiotte, and
  Lefebvre]{Blondel2008}
V.~D. Blondel, J.-L. Guillaume, R.~Lambiotte, and E.~Lefebvre.
\newblock Fast unfolding of communities in large networks.
\newblock \emph{Journal of Statistical Mechanics: Theory and Experiment},
  2008\penalty0 (10):\penalty0 10008, 2008.

\bibitem[Lancichinetti and Fortunato(2009)]{Lancichinetti2009}
A.~Lancichinetti and S.~Fortunato.
\newblock Community detection algorithms: a comparative analysis.
\newblock \emph{Physical Review E}, 80\penalty0 (5):\penalty0 056117, 2009.

\bibitem[Menache et~al.(2002)Menache, Mannor, and Shimkin]{Menache2002}
I.~Menache, S.~Mannor, and N.~Shimkin.
\newblock Q-{C}ut --- {D}ynamic discovery of sub-goals in reinforcement
  learning.
\newblock In \emph{Proceedings of the 13th European Conference on Machine
  Learning}, ECML '02, pages 295--306. Springer-Verlag, 2002.

\bibitem[{\c{S}}im{\c{s}}ek and Barto(2004)]{Simsek2004}
{\"O}.~{\c{S}}im{\c{s}}ek and A.~G. Barto.
\newblock Using relative novelty to identify useful temporal abstractions in
  reinforcement learning.
\newblock In \emph{Proceedings of the 21st International Conference on Machine
  Learning}, ICML '04, pages 95--102. ACM, 2004.

\bibitem[{\c{S}}im{\c{s}}ek and Barto(2009)]{Simsek2009}
{\"O}.~{\c{S}}im{\c{s}}ek and A.~G. Barto.
\newblock Skill characterization based on betweenness.
\newblock In \emph{Advances in Neural Information Processing Systems 21},
  NeurIPS '09, pages 1497--1504. Curran Associates, Inc., 2009.

\bibitem[Moradi et~al.(2010)Moradi, Shiri, and Entezari]{Moradi2010}
P.~Moradi, M.~E. Shiri, and N.~Entezari.
\newblock Automatic skill acquisition in reinforcement learning agents using
  connection bridge centrality.
\newblock In \emph{International Conference on Future Generation Communication
  and Networking}, pages 51--62. Springer, 2010.

\bibitem[Rad et~al.(2010)Rad, Hasler, and Moradi]{Rad2010}
A.~A. Rad, M.~Hasler, and P.~Moradi.
\newblock Automatic skill acquisition in reinforcement learning using
  connection graph stability centrality.
\newblock In \emph{Proceedings of 2010 IEEE International Symposium on Circuits
  and Systems}, pages 697--700. IEEE, 2010.

\bibitem[Imanian and Moradi(2011)]{Imanian2011}
M.~A. Imanian and P.~Moradi.
\newblock Autonomous subgoal discovery in reinforcement learning agents using
  bridgeness centrality measure.
\newblock \emph{International Journal of Electrical and Computer Sciences},
  11\penalty0 (5):\penalty0 54--62, 2011.

\bibitem[Moradi et~al.(2012)Moradi, Shiri, Rad, Khadivi, and
  Hasler]{Moradi2012}
P.~Moradi, M.~E. Shiri, A.~A. Rad, A.~Khadivi, and M.~Hasler.
\newblock Automatic skill acquisition in reinforcement learning using graph
  centrality measures.
\newblock \emph{Intelligent Data Analysis}, 16\penalty0 (1):\penalty0 113--135,
  2012.

\bibitem[Metzen(2013)]{Metzen2013}
J.~H. Metzen.
\newblock Online skill discovery using graph-based clustering.
\newblock In \emph{Proceedings of the Tenth European Workshop on Reinforcement
  Learning}, EWRL '13, pages 77--88. PMLR, 2013.

\bibitem[{\c{S}}im{\c{s}}ek et~al.(2005){\c{S}}im{\c{s}}ek, Wolfe, and
  Barto]{Simsek2005}
{\"O}.~{\c{S}}im{\c{s}}ek, A.~P. Wolfe, and A.~G. Barto.
\newblock Identifying useful subgoals in reinforcement learning by local graph
  partitioning.
\newblock In \emph{Proceedings of the 22nd International Conference on Machine
  learning}, ICML '05, pages 816--823. ACM, 2005.

\bibitem[Kazemitabar and Beigy(2009)]{Kazemitabar2009}
S.~J. Kazemitabar and H.~Beigy.
\newblock Using strongly connected components as a basis for autonomous skill
  acquisition in reinforcement learning.
\newblock In \emph{6th International Symposium on Neural Networks}, ISSN '09,
  pages 794--803. Springer, 2009.

\bibitem[Entezari et~al.(2010)Entezari, Shiri, and Moradi]{Entezari2010}
N.~Entezari, M.~E. Shiri, and P.~Moradi.
\newblock A local graph clustering algorithm for discovering subgoals in
  reinforcement learning.
\newblock In \emph{International Conference on Future Generation Communication
  and Networking}, pages 41--50. Springer, 2010.

\bibitem[Bacon and Precup(2013)]{Bacon2013}
P.-L. Bacon and D.~Precup.
\newblock Using label propagation for learning temporally abstract actions in
  reinforcement learning.
\newblock In \emph{Proceedings of the Workshop on Multiagent Interaction
  Networks}, MAIN '13, 2013.

\bibitem[Taghizadeh and Beigy(2013)]{Taghizadeh2013}
N.~Taghizadeh and H.~Beigy.
\newblock A novel graphical approach to automatic abstraction in reinforcement
  learning.
\newblock \emph{Robotics and Autonomous Systems}, 61\penalty0 (8):\penalty0
  821--835, 2013.

\bibitem[Kazemitabar et~al.(2018)Kazemitabar, Taghizadeh, and
  Beigy]{Kazemitabar2018}
S.~J. Kazemitabar, N.~Taghizadeh, and H.~Beigy.
\newblock A graph-theoretic approach toward autonomous skill acquisition in
  reinforcement learning.
\newblock \emph{Evolving Systems}, 9\penalty0 (3):\penalty0 227--244, 2018.

\bibitem[Ramesh et~al.(2019)Ramesh, Tomar, and Ravindran]{Ramesh2019}
R.~Ramesh, M.~Tomar, and B.~Ravindran.
\newblock Successor options: an option discovery framework for reinforcement
  learning.
\newblock In \emph{Proceedings of the 28th International Joint Conference on
  Artificial Intelligence}, IJCAI '19, pages 3304--3310. AAAI Press, 2019.

\bibitem[Mannor et~al.(2004)Mannor, Menache, Hoze, and Klein]{Mannor2004}
S.~Mannor, I.~Menache, A.~Hoze, and U.~Klein.
\newblock Dynamic abstraction in reinforcement learning via clustering.
\newblock In \emph{Proceedings of the 21st International Conference on Machine
  Learning}, ICML '04, pages 71--78. ACM, 2004.

\bibitem[Davoodabadi and Beigy(2011)]{Davoodabadi2011}
M.~Davoodabadi and H.~Beigy.
\newblock A new method for discovering subgoals and constructing options in
  reinforcement learning.
\newblock In \emph{Proceedings of the 5th Indian International Conference on
  Artificial Intelligence}, pages 441--450, 2011.

\bibitem[Shoeleh and Asadpour(2017)]{Shoeleh2017}
F.~Shoeleh and M.~Asadpour.
\newblock Graph based skill acquisition and transfer learning for continuous
  reinforcement learning domains.
\newblock \emph{Pattern Recognition Letters}, 87:\penalty0 104--116, 2017.

\bibitem[Xu et~al.(2018)Xu, Yang, and Li]{Xu2018}
X.~Xu, M.~Yang, and G.~Li.
\newblock Constructing temporally extended actions through incremental
  community detection.
\newblock \emph{Computational Intelligence and Neuroscience}, 2018.

\bibitem[Farahani and Mozayani(2019)]{Davoodabadi2019}
M.~D. Farahani and N.~Mozayani.
\newblock Automatic construction and evaluation of macro-actions in
  reinforcement learning.
\newblock \emph{Applied Soft Computing}, 82:\penalty0 105574, 2019.

\bibitem[Machado et~al.(2017)Machado, Bellemare, and Bowling]{Machado2017}
M.~C. Machado, M.~G. Bellemare, and M.~Bowling.
\newblock A {Laplacian} framework for option discovery in reinforcement
  learning.
\newblock In \emph{Proceedings of the 34th International Conference on Machine
  Learning}, ICML '17, pages 2295--2304. PMLR, 2017.

\bibitem[Jinnai et~al.(2019)Jinnai, Park, Abel, and Konidaris]{Jinnai2019}
Y.~Jinnai, J.~W. Park, D.~Abel, and G.~Konidaris.
\newblock Discovering options for exploration by minimizing cover time.
\newblock In \emph{Proceedings of the 36th International Conference on Machine
  Learning}, ICML '19, pages 3130--3139. PMLR, 2019.

\bibitem[Bacon et~al.(2017)Bacon, Harb, and Precup]{Bacon2017}
P.-L. Bacon, J.~Harb, and D.~Precup.
\newblock The option-critic architecture.
\newblock In \emph{Proceedings of the AAAI Conference on Artificial
  Intelligence}, volume~31, 2017.

\bibitem[Sutton et~al.(1999{\natexlab{b}})Sutton, McAllester, Singh, and
  Mansour]{Sutton1999PolicyGradient}
R.~S. Sutton, D.~McAllester, S.~Singh, and Y.~Mansour.
\newblock Policy gradient methods for reinforcement learning with function
  approximation.
\newblock In \emph{Advances in Neural Information Processing Systems 12},
  NeurIPS '99. MIT Press, 1999{\natexlab{b}}.

\bibitem[Riemer et~al.(2018)Riemer, Liu, and Tesauro]{Riemer2018}
M.~Riemer, M.~Liu, and G.~Tesauro.
\newblock Learning abstract options.
\newblock In \emph{Advances in Neural Information Processing Systems 31},
  NeurIPS '18. Curran Associates, Inc., 2018.

\bibitem[Fox et~al.(2017)Fox, Krishnan, Stoica, and Goldberg]{Fox2017}
R.~Fox, S.~Krishnan, I.~Stoica, and K.~Goldberg.
\newblock Multi-level discovery of deep options.
\newblock \emph{arXiv preprint:1703.08294}, 2017.

\bibitem[Levy et~al.(2019)Levy, Konidaris, Platt, and Saenko]{Levy2019}
A.~Levy, G.~Konidaris, R.~Platt, and K.~Saenko.
\newblock Learning multi-level hierarchies with hindsight.
\newblock In \emph{Proceedings of the 7th International Conference on Learning
  Representations}, ICLR '19, 2019.

\bibitem[Dietterich(2000)]{Dietterich2000}
T.~G. Dietterich.
\newblock Hierarchical reinforcement learning with the {MAXQ} value function
  decomposition.
\newblock \emph{Journal of Artificial Intelligence Resesearch}, 13\penalty0
  (1):\penalty0 227--303, 2000.

\bibitem[McGovern et~al.(1997)McGovern, Sutton, and Fagg]{McGovern1997}
A.~McGovern, R.~S. Sutton, and A.~H. Fagg.
\newblock Roles of macro-actions in accelerating reinforcement learning.
\newblock In \emph{Grace Hopper Celebration of Women in Computing}, volume~1,
  pages 13--18, 1997.

\bibitem[Sutton and Precup(1998)]{Precup1998}
R.~S. Sutton and D.~Precup.
\newblock Intra-option learning about temporally abstract actions.
\newblock In \emph{Proceedings of the 15th International Conference on Machine
  Learning}, pages 556--564. Morgan Kaufman, 1998.

\bibitem[Watkins(1989)]{Watkins1989}
C.~J. C.~H. Watkins.
\newblock \emph{Learning from delayed rewards}.
\newblock PhD thesis, King's College, Cambridge United Kingdom, 1989.

\bibitem[Mahadevan and Maggioni(2007)]{Mahadevan2007}
S.~Mahadevan and M.~Maggioni.
\newblock Proto-value functions: A {Laplacian} framework for learning
  representation and control in {Markov} decision processes.
\newblock \emph{Journal of Machine Learning Research}, 8\penalty0 (10), 2007.

\bibitem[Konidaris and Barto(2009)]{Konidaris2009}
G.~Konidaris and A.~Barto.
\newblock Skill discovery in continuous reinforcement learning domains using
  skill chaining.
\newblock In \emph{Advances in Neural Information Processing Systems 22},
  NeurIPS '09, pages 1015--1023. Curran Associates, Inc., 2009.

\bibitem[Farahani and Mozayani(2020)]{Davoodabadi2020}
D.~M. Farahani and N.~Mozayani.
\newblock Evaluating skills in hierarchical reinforcement learning.
\newblock \emph{International Journal of Machine Learning and Cybernetics},
  11\penalty0 (10):\penalty0 2407--2420, 2020.

\end{thebibliography}

%%%%%%%%%%%%%%%%%%%%%%%%%%%%%%%%%%%%%%%%%%%%%%%%%%%%%%%%%%%%
%%%%%%%%           Supplementary Material           %%%%%%%%
%%%%%%%%%%%%%%%%%%%%%%%%%%%%%%%%%%%%%%%%%%%%%%%%%%%%%%%%%%%%
\newpage
\appendix
\normalsize

\setcounter{figure}{0} 

\title{Creating Multi-Level Skill Hierarchies in Reinforcement Learning\\\vspace{0.2cm} {\sc{Supplementary Material}}}

\author{%
  Joshua B.~Evans\\
  Department of Computer Science\\
  University of Bath\\
  Bath, United Kingdom \\
  \texttt{jbe25@bath.ac.uk} \\
  \And
  {\"O}zg{\"u}r {\c{S}}im{\c{s}}ek \\
  Department of Computer Science\\
  University of Bath\\
  Bath, United Kingdom \\
  \texttt{o.simsek@bath.ac.uk} \\
}

\appendix

\settitle

\section{Environments}
\label{sect:sample_domains}

\textbf{Gridworlds} include Rooms, Grid, Office, and Maze~\citep{Ramesh2019}. They had four primitive actions: north, south, east, and west. These actions move the agent in the intended direction (unless the intended direction faces a wall, in which case the agent remains in the same state). The reward is \(-0.001\) for each action and an additional \(+1.0\) for reaching the goal state. There is a single start state and a single goal state, selected for each run from a list of possibilities. 

\textbf{Multi-Floor Office} is an extension of Office to multiple floors. All foors are connected by an elevator, which occupies the same grid square on each floor, where the agent has two additional primitive actions: up and down. These actions move the agent to the adjecent floor in the intended direction (unless there is no other floor in that direction). 

\textbf{Taxi} is a \(5 \times 5\) grid with four special squares (R, G, B, and Y) that serve as possible pick-up and drop-off locations for a passenger. An episode starts with the taxi at a random square, the passenger at a random special square, and the destination another random special square. Six actions are available in each state: north, south, east, west, pick-up, and put-down. The navigation actions are identical to those in the gridworlds, as described above. Pick-up and put-down have the intended effect when appropriate; otherwise they do not change the state. The reward is $-0.001$ for each action and an additional $+1.0$ when the passenger is put down at the destination. 

\textbf{Towers of Hanoi} contains four discs of different sizes, placed on three poles. An episode starts with all discs stacked on the leftmost pole. Primitive actions move the top disc from one pole to any other pole, with the constraint that a disc cannot be placed on top of a disc smaller than itself. The reward is $-0.001$ for each action and an additional reward of $+1.0$ at the goal state, where when all three discs are stacked on the rightmost pole.

\section{Methodology}
\label{sect:methodology}

\textbf{Generating options.} To generate Louvain options, the Louvain algorithm (\(\rho = 0.05\)) was applied to the state transition graph, resulting in a set of partitions. Any partition where the mean number of nodes per cluster was smaller than \(4\) was discarded.
For all remaining partitions, options were defined for efficiently taking the agent from a cluster $c_i$ to each of its neighbouring clusters $c_j$ if a directed edge existed from a node in $c_i$ to a node in $c_j$.
The Louvain options were arranged into a multi-level hierarchy, where options for navigating between clusters in partition \(i\) could call skills for navigating between clusters in partition \(i-1\). Only options at the lowest level of the hierarchy could call primitive actions. Options generated using alternative methods called primitive actions directly. 

Eigenoptions~\citep{Machado2017} were generated by computing the normalised Laplacian of the state transition graph and using its eigenvectors to define pseudo-reward functions for each Eigenoption to maximise. In Office, the first \(32\) eigenvectors (and their negations) were used. In all other environments, the first \(16\) eigenvectors (and their negations) were used. Options for navigating to local maxima of betweenness~\cite{Simsek2009} were generated by selecting all local maxima as subgoals and defining options for navigating to each subgoal from the nearest \(30\) states.

\textbf{Learning option policies.} For all methods except Eigenoptions, option policies were learned using macro-Q learning \cite{McGovern1997}, with learning rate \(\alpha= 0.6\), initial action values \(Q_{0}=0\), and discount factor \(\gamma=1\). All agents used an \(\epsilon\)-greedy exploration strategy with \(\epsilon = 0.2\). For options based on clustering, the agent started in a random state in the source cluster. It received a reward of \(-0.01\) for each action taken and an additional \(+1.0\) for reaching a state in the goal cluster. For options based on node betweenness, the agent started in a random state in the initiation set. It received a reward of \(-0.01\) at each decision stage, an additional \(+1.0\) for reaching the subgoal state, and an additional \(-1.0\) for leaving the initiation set. For Eigenoptions, value iteration was used to produce policies that maximised the pseudo-reward function of each Eigenoption.

\textbf{Producing learning curves.} All agents were trained using macro-Q~\cite{McGovern1997} and intra-option learning~\cite{Precup1998} updates, which were performed every time an option at any level of the hierarchy terminated. A learning rate of \(\alpha = 0.4\), discount factor of \(\gamma = 1\), and initial action values of \(Q_{0} = 0\) were used in all experiments. All agents used an \(\epsilon\)-greedy exploration strategy with \(\epsilon = 0.1\). All learning curves show evaluation performance. After training each agent for one epoch, the learned policy was evaluated (with exploration and learning disabled) in a separate instance of the environment.

\textbf{Applying the Louvain algorithm in Pinball.} The state transition graph was constructed by following the method used by \citet{Mahadevan2007}. Initially, \(4000\) states were randomly sampled and a node representing each state was added to the graph. Edges were then added between each node and its \(10\) nearest neighbours. An between two locations \(u\) and \(v\) with Euclidean distance \(d\) between them was assigned a weight of \(e^{-4d^2}\). Finally, the Louvain algorithm (\(\rho = 0.05\)) was applied to the resulting graph.

\section{Comparison to Skills that Navigate to Local Maxima of Betweenness}
\label{sect:node_betweenness}

Here we explore the relationship between Louvain skills and the skill chracterisation proposed by \citet{Simsek2009}. The latter is a subgoal-based approach that captures various definitions of the bottleneck concept. It defines skills that navigate to local maxima of betweenness. Both approaches address the conceptual problem of what makes a useful skill, explicitly defining a target set of skills for the agent to learn. 

There is substantial overlap between Louvain skills and skills that navigate to local maxima of betweenness. They both include skills that traverse bottleneck states, including those that navigate between rooms in Rooms, picking up the passenger in Taxi, and navigating different parts of the grid in Taxi. In Towers of Hanoi, all Louvain skills traverse states that are also identified as local maxima of betweenness. The highest local maxima of betweenness correspond to the states that separate Louvain clusters at level 3; the second highest local maxima of betweenness correspond to the states that separate Louvain clusters at level 2. 

On the other hand, there are many Louvain skills that do not correspond to navgating to local maxima of betweenness. Examples include the Louvain skills that navigate within a single room in Rooms. 

Most importantly, Louvain skills and skills that navigate to local maxima of betweenness differ in how they can be arranged hierarchically. Even in Towers of Hanoi, where there is a clear hierarchical structure between the larger and smaller local maxima of betweenness, it is not clear how to exploit the betweenness metric to form a multi-level hierarchy. In contrast, the modularity metric approximated by the Louvain algorithm provides a clear and principled way of building a multi-level hierarchy.

\section{Cluster Hierarchies}
\label{sect:full_sample_hierarchies}
Figure~\ref{fig:skills_supp} shows the cluster hierarchies produced by the Louvain algorithm when applied to the state transition graphs of Grid and Maze. It also shows the first level of the cluster hierarchy in Office, which was omitted in the main paper due to space limitations.

\newsavebox{\suppimagebox}
\begin{figure}[t]
    \centering
	\begin{subfigure}{\linewidth}    
        \savebox{\suppimagebox}{\begin{subfigure}{0.24\linewidth}\includegraphics[width=0.8\linewidth]{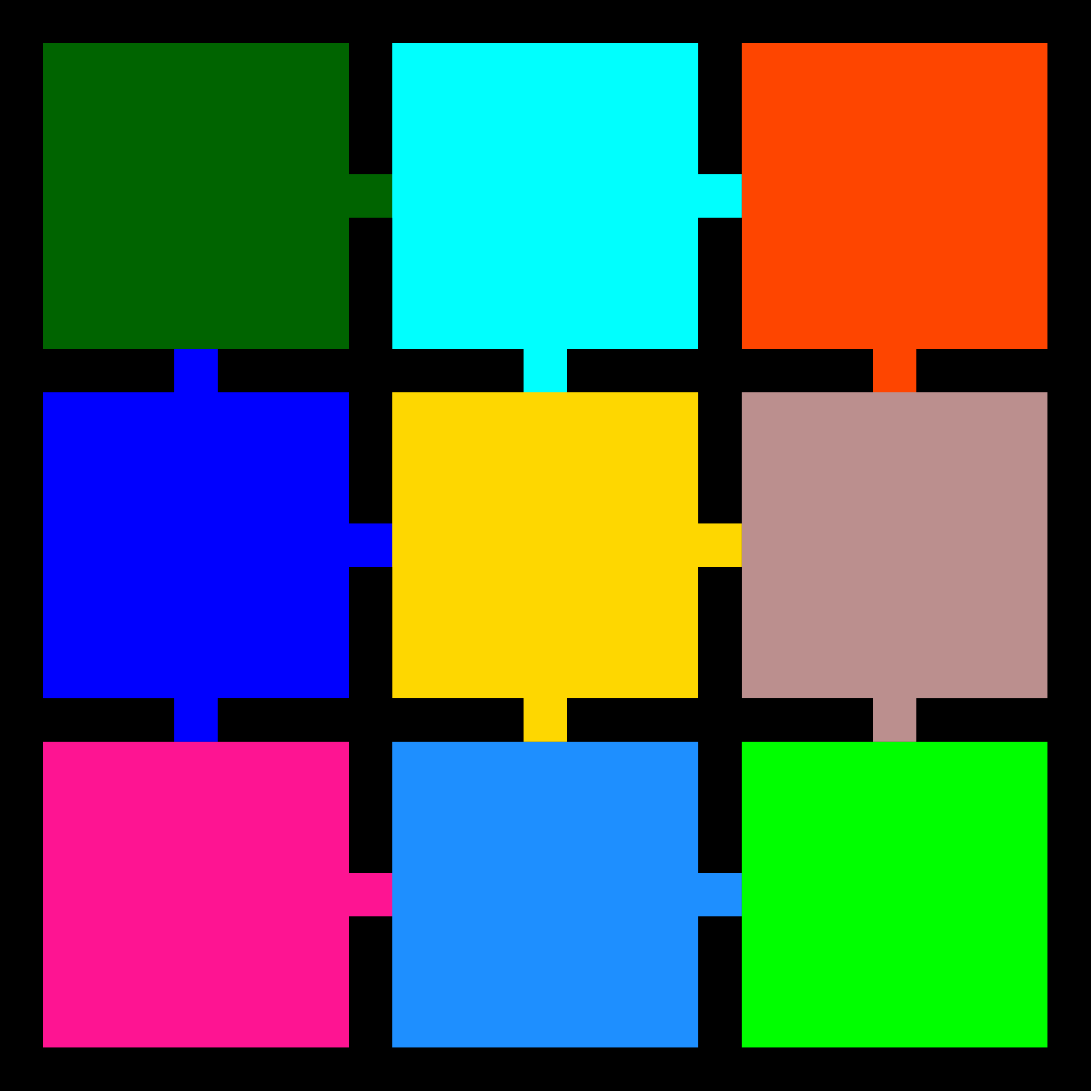}\end{subfigure}}% 
    	\begin{subfigure}{0.0175 \linewidth}
			\centering\raisebox{\dimexpr.5\ht\suppimagebox-.5\height+\baselineskip}{% Raise smaller image into place
            \includegraphics[width=\linewidth]{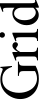}}%
		\end{subfigure}%
        \hspace{0.01\linewidth}%
        \begin{subfigure}{0.24 \linewidth}
        	\setlength{\abovecaptionskip}{2.0pt}
            \centering
            \includegraphics[width=0.8\linewidth]{"figures/supplementary_figures/grid/grid_level_4".pdf}
            \caption*{Level 4}
        \end{subfigure}%
        \hfill%
        \begin{subfigure}{0.24 \linewidth} 
            \centering
        	\setlength{\abovecaptionskip}{2.0pt}
            \includegraphics[width=0.8\linewidth]{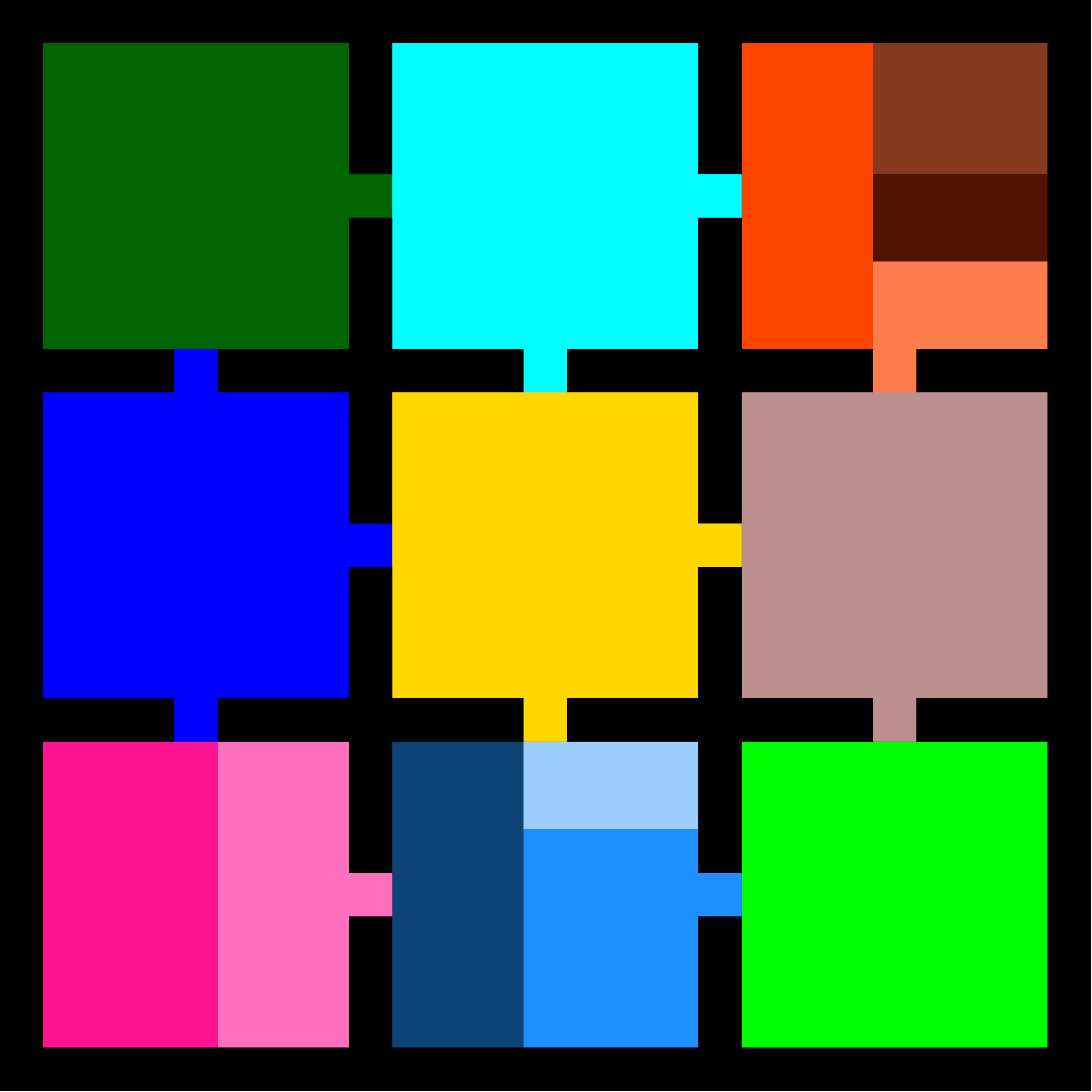}
            \caption*{Level 3}
        \end{subfigure}%
        \hfill     
        \begin{subfigure}{0.24 \linewidth}     
            \centering
           	\setlength{\abovecaptionskip}{2.0pt}
            \includegraphics[width=0.8\linewidth]{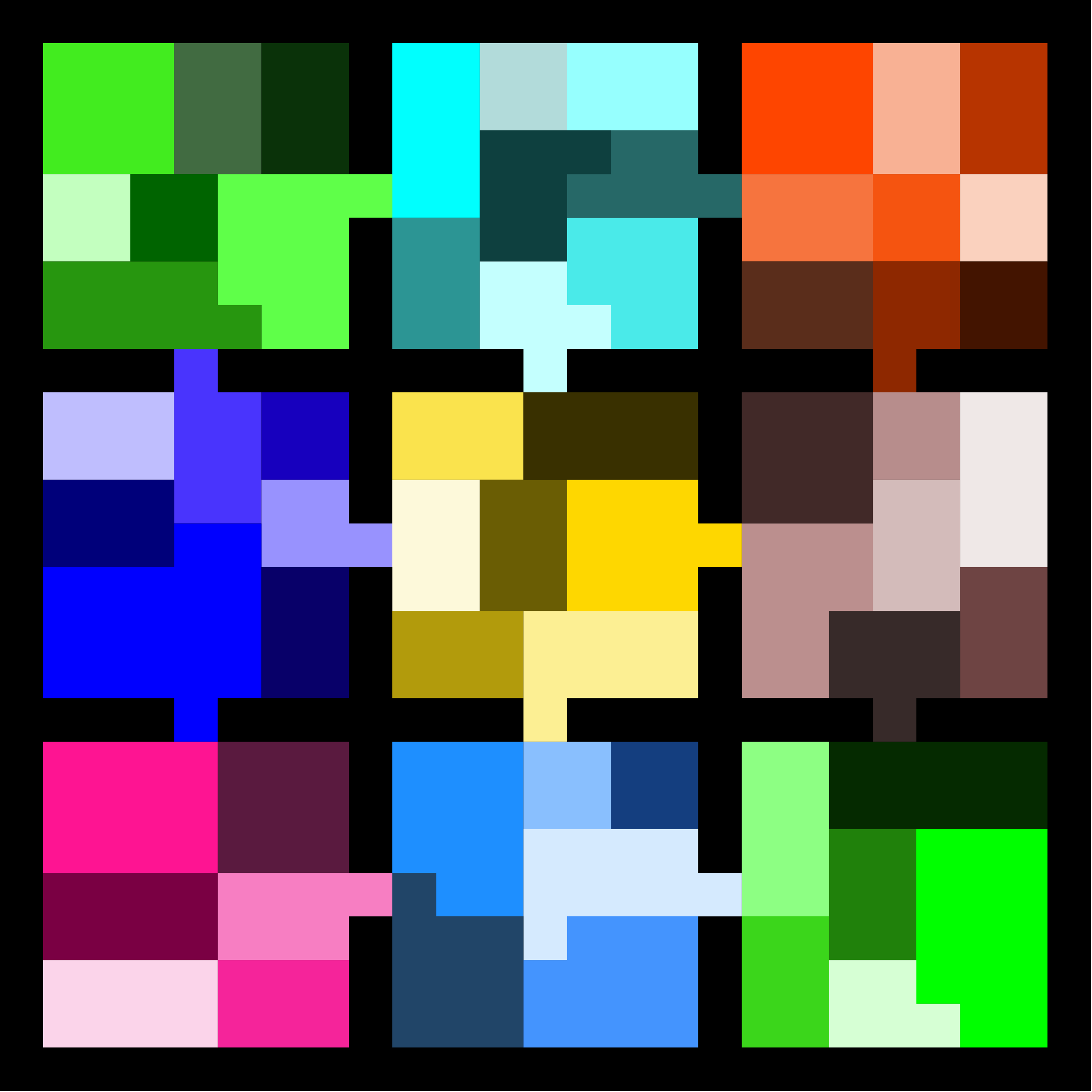}
            \caption*{Level 2}
        \end{subfigure}%
        \hfill 
        \begin{subfigure}{0.24 \linewidth}     
            \centering
            \setlength{\abovecaptionskip}{2.0pt}
            \includegraphics[width=0.8\linewidth]{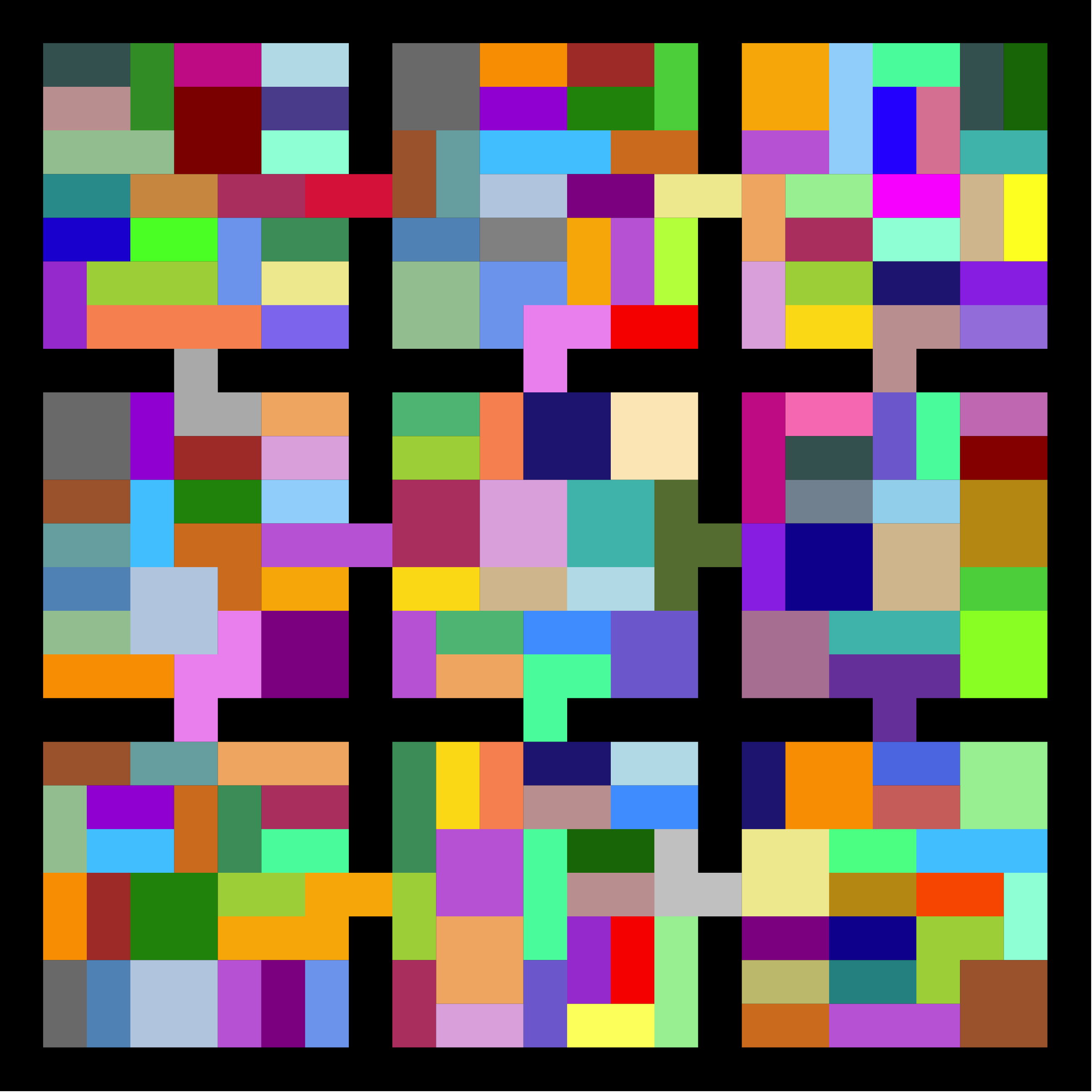}
			\caption*{Level 1}
        \end{subfigure}%
    \end{subfigure}%

	\vspace{0.25cm}
	\begin{subfigure}{\linewidth}
        \savebox{\suppimagebox}{\begin{subfigure}{0.19\linewidth}\includegraphics[width=0.9\linewidth]{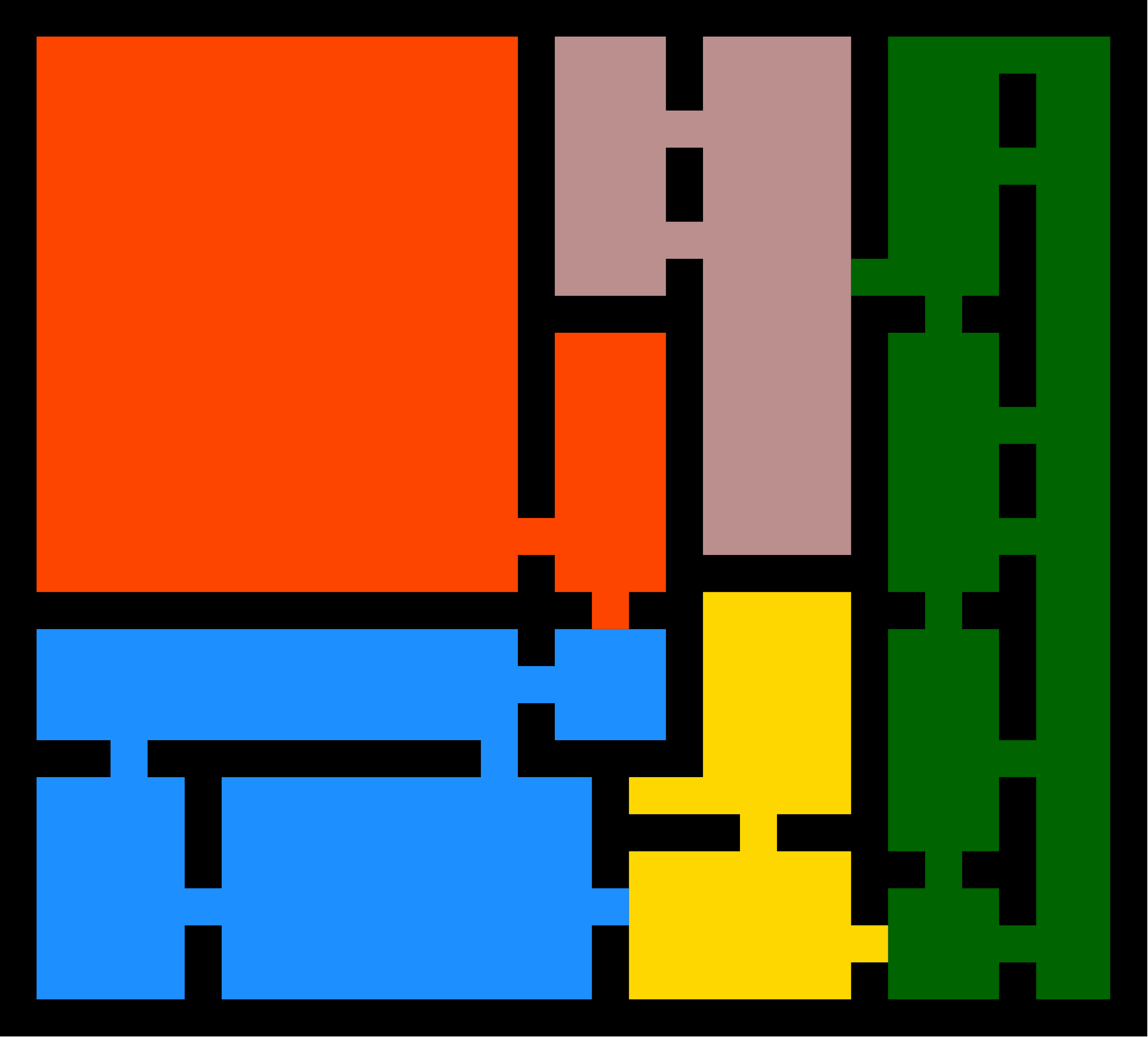}\end{subfigure}}% 
    	\begin{subfigure}{0.0175 \linewidth}
			\centering\raisebox{\dimexpr.5\ht\suppimagebox-.5\height+\baselineskip}{% Raise smaller image into place
            \includegraphics[width=\linewidth]{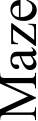}}%
		\end{subfigure}%
        \hspace{0.01\linewidth}%
        \begin{subfigure}{0.19 \linewidth}
            \centering
            \setlength{\abovecaptionskip}{2.0pt}
            \includegraphics[width=0.9\linewidth]{"figures/supplementary_figures/maze/maze_level_5".pdf}
			\caption*{Level 5}
        \end{subfigure}%
        \hfill
        \begin{subfigure}{0.19 \linewidth}
            \centering
           	\setlength{\abovecaptionskip}{2.0pt}
            \includegraphics[width=0.9\linewidth]{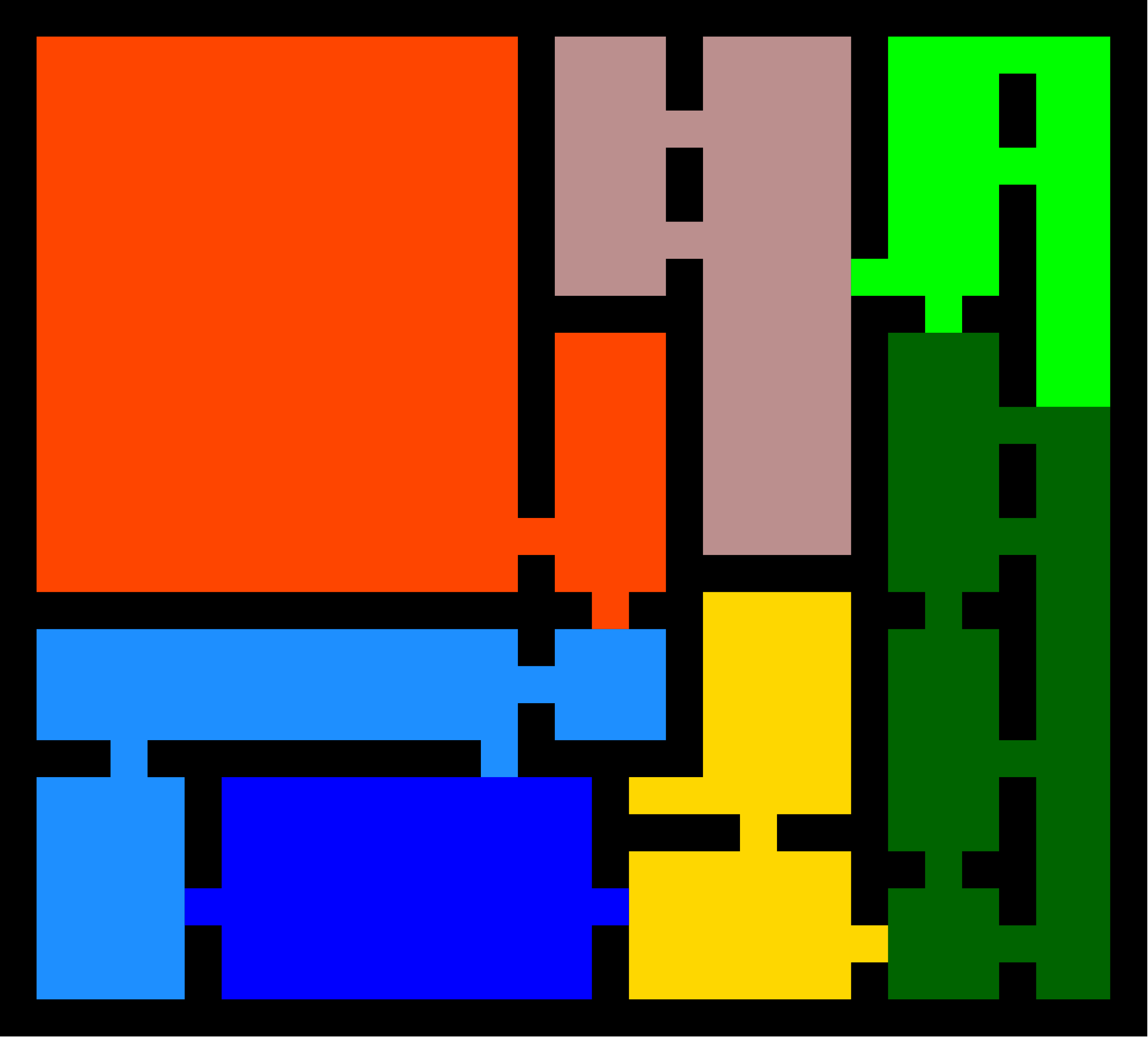}
			\caption*{Level 4}
        \end{subfigure}%
        \hfill
        \begin{subfigure}{0.19 \linewidth}
            \centering
           	\setlength{\abovecaptionskip}{2.0pt}
            \includegraphics[width=0.9\linewidth]{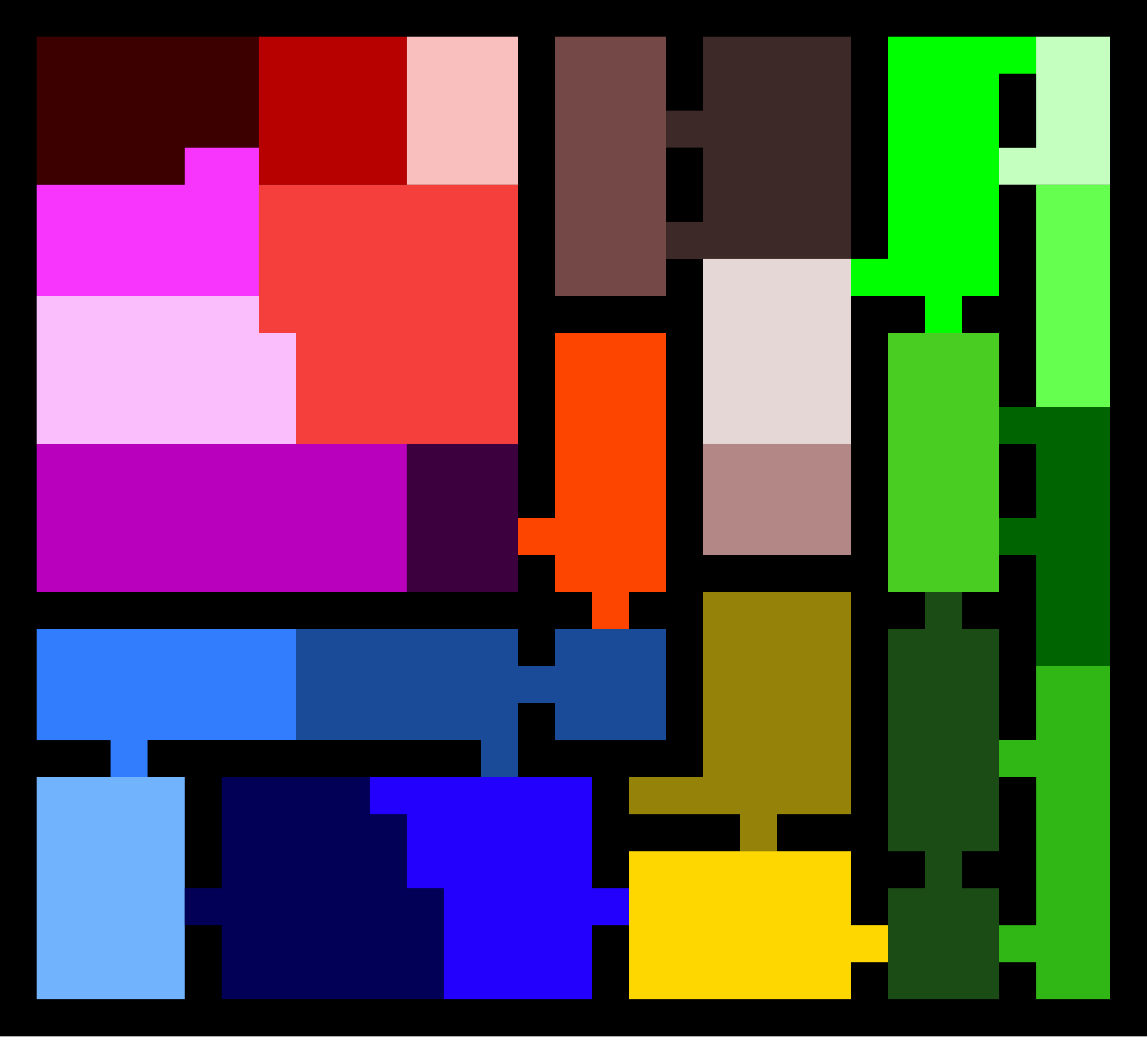}
            \caption*{Level 3}
        \end{subfigure}%
        \hfill
        \begin{subfigure}{0.19 \linewidth}
            \centering
           	\setlength{\abovecaptionskip}{2.0pt}
            \includegraphics[width=0.9\linewidth]{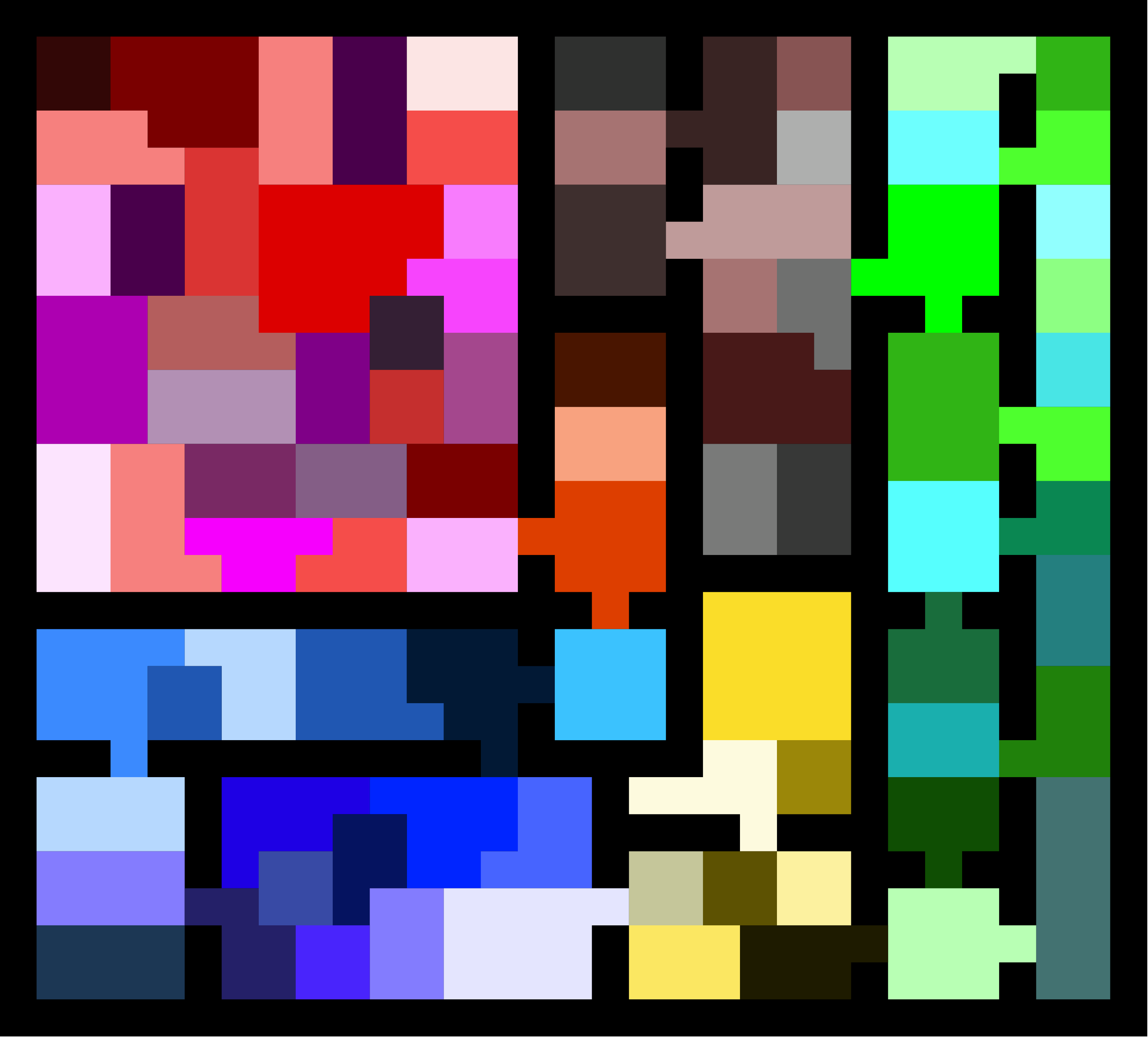}
            \caption*{Level 2}
        \end{subfigure}%
        \hfill
        \begin{subfigure}{0.19 \linewidth}
            \centering
           	\setlength{\abovecaptionskip}{2.0pt}
            \includegraphics[width=0.9\linewidth]{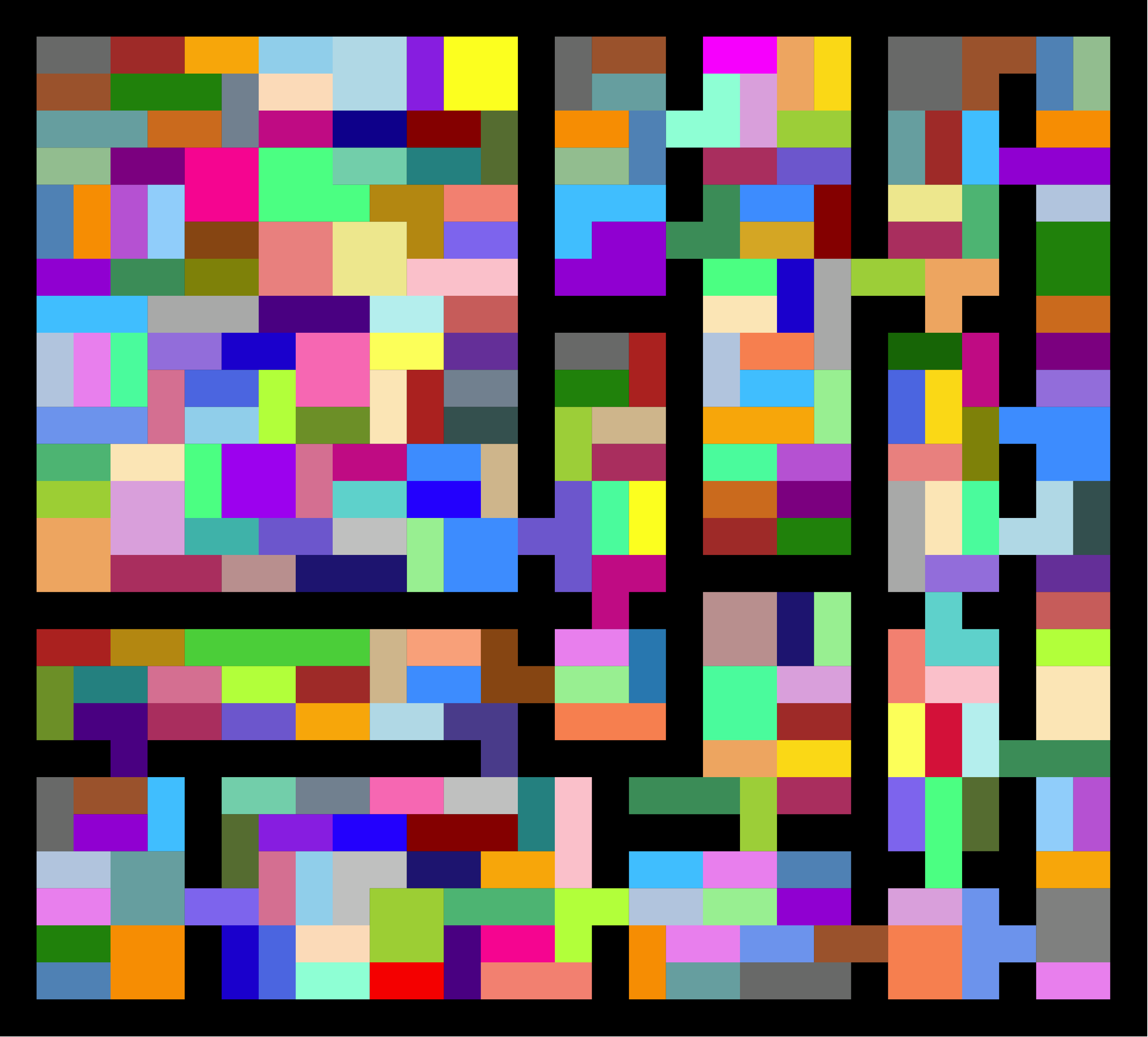}
            \caption*{Level 1}
        \end{subfigure}%
    \end{subfigure}% 

	\vspace{0.25cm}
    \begin{subfigure}{\linewidth}
        \savebox{\suppimagebox}{\begin{subfigure}{0.24\linewidth}\includegraphics[width=0.95\linewidth]{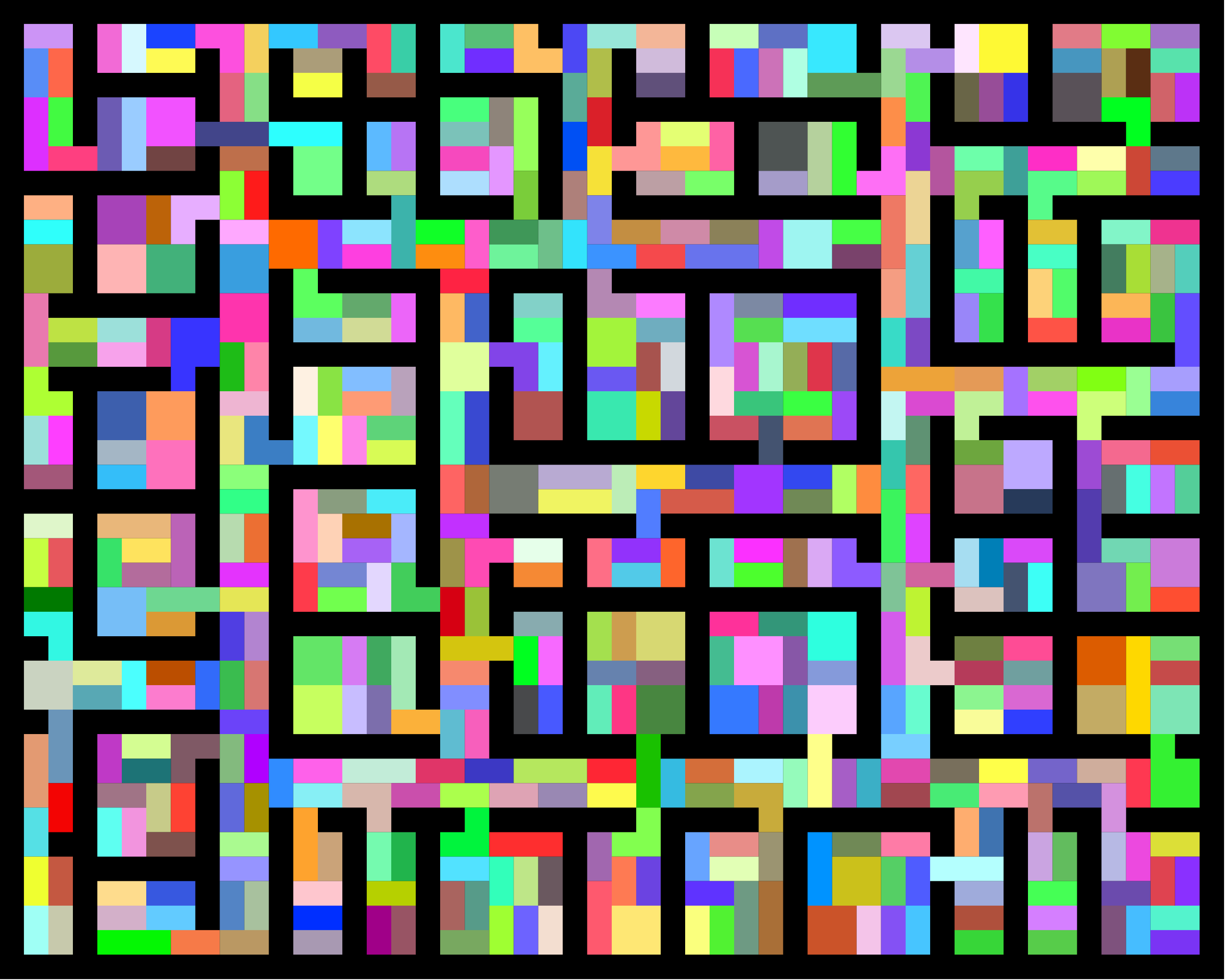}\end{subfigure}}% 
    	\begin{subfigure}{0.0175 \linewidth}
			\centering\raisebox{\dimexpr.5\ht\suppimagebox-.5\height+\baselineskip}{% Raise smaller image into place
            \includegraphics[width=\linewidth]{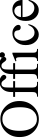}}%
		\end{subfigure}%
        \hspace{0.01\linewidth}%
        \begin{subfigure}{0.24\linewidth}
            \centering
           	\setlength{\abovecaptionskip}{2.0pt}
            \includegraphics[width=0.95\linewidth]{"figures/supplementary_figures/office/office1k_level_0".pdf}
            \caption*{Level 1}
        \end{subfigure}%
        \hfill
    \end{subfigure}%  
    \caption{Top two rows: Cluster hierarchies produced by the Louvain algorithm in Grid and Maze. Bottom row: The lowest level of the cluster hierarchy in Office.}
\label{fig:skills_supp}
\end{figure}

\section{Sensitivity to the Resolution Parameter}
\label{sect:resolution_sensitivity}

\begin{figure}[b]
	\begin{subfigure}{0.5\linewidth}
		\centering
		\includegraphics[width=0.75\linewidth]{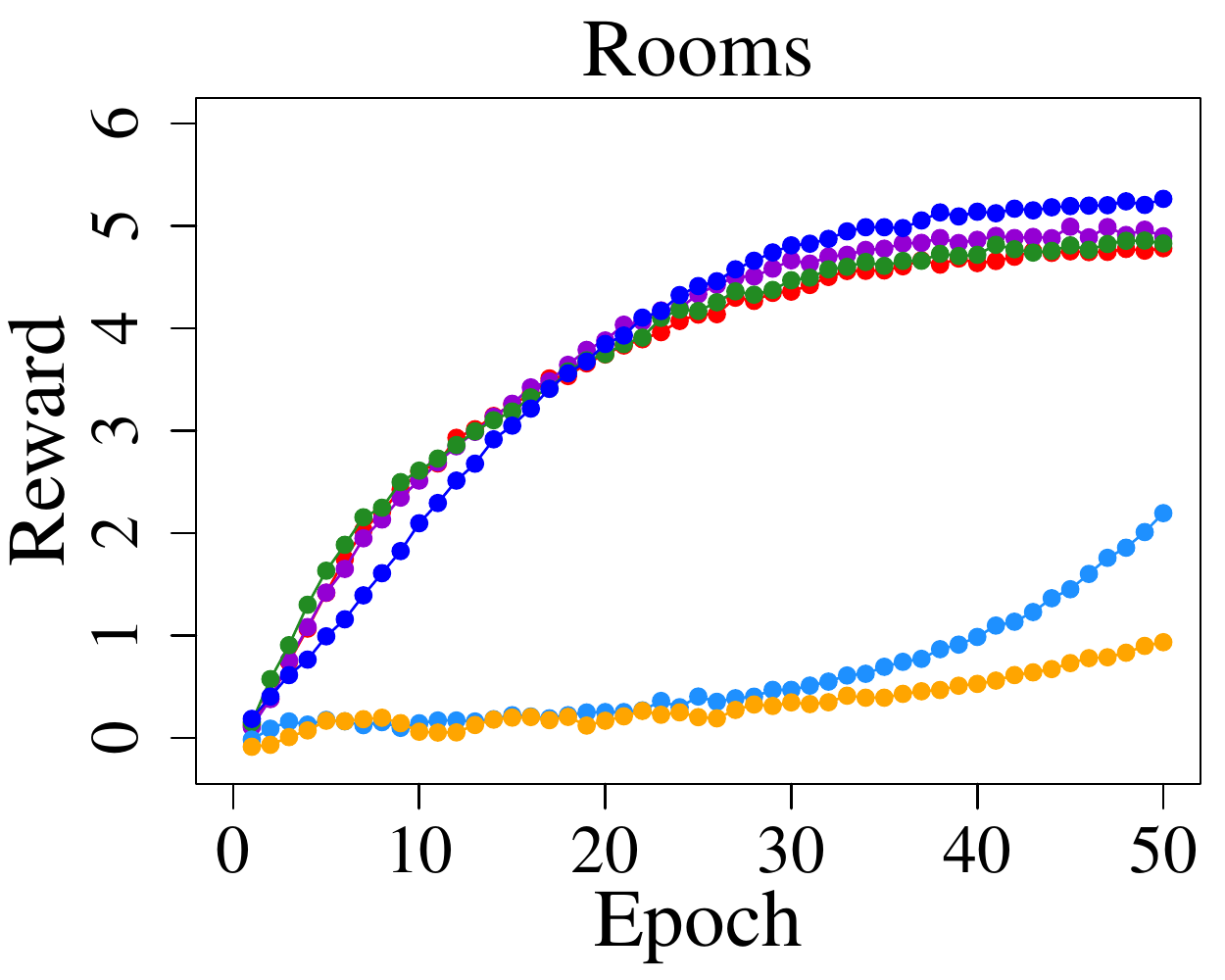}
	\end{subfigure}%
	\hfill
	\begin{subfigure}{0.5\linewidth}
		\centering
		\includegraphics[width=0.75\linewidth]{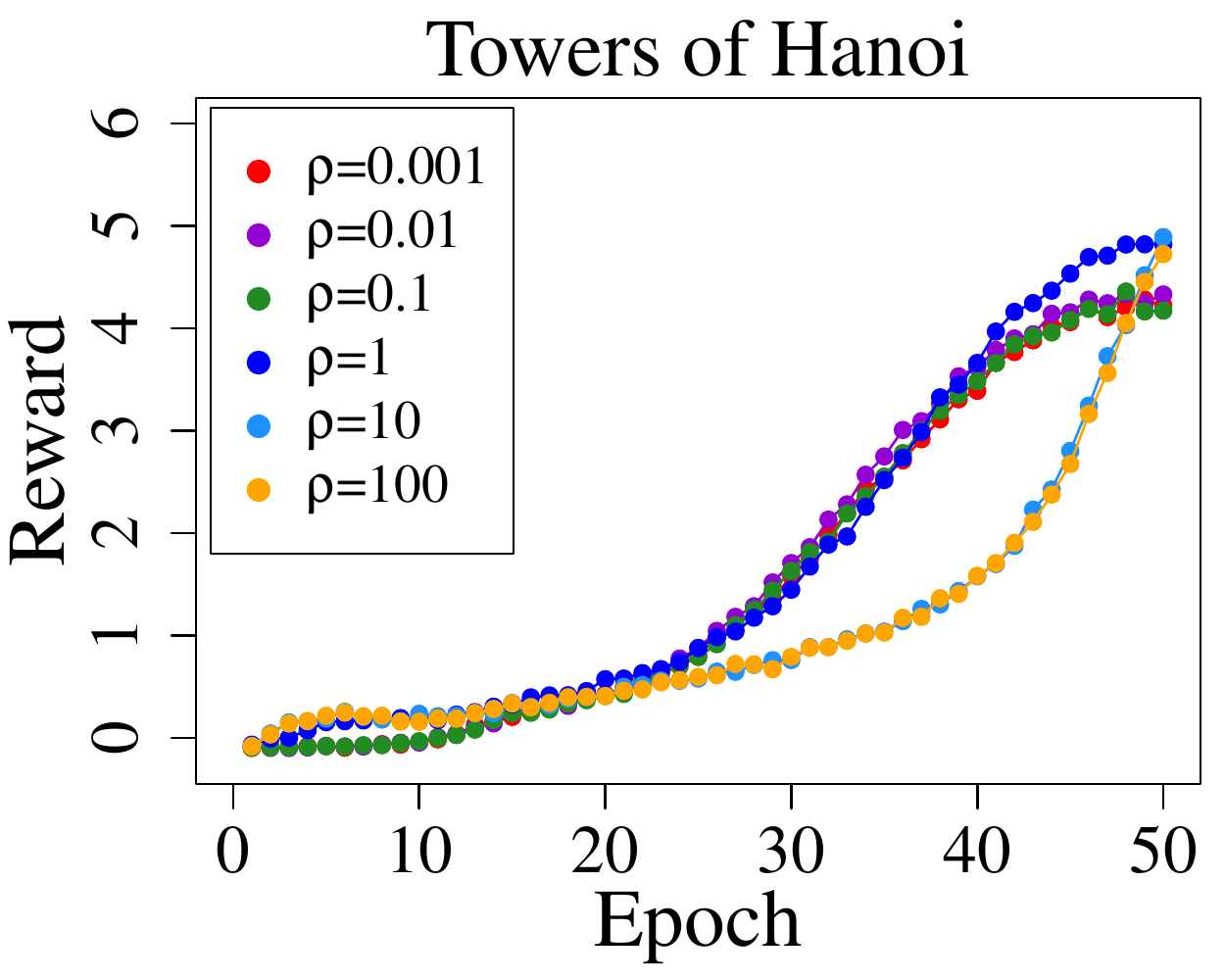}
	\end{subfigure}%
	\hfill
	\caption{Agent performance with Louvain skills generated using different settings of the resolution parameter \(\rho\).}
	\label{fig:ResolutionPerformance}
\end{figure}

Figure~\ref{fig:ResolutionPerformance} shows the performance of agents in Rooms and Towers of Hanoi using Louvain skills generated using different values of \(\rho\). Louvain skills created using lower values of \(\rho\) consistently outperformed those created by using higher values, and very high values (\(\rho \geq 10\)) generally led to performance similar to that obtained by using primitive actions only. Lower \(\rho\) values lead to deeper hierarchies that contain skills for navigating the environment over varying timescales. In contrast, higher \(\rho\) values result in shallower skill hierarchies that contain few---or, in the extreme, no---levels of skills above primitive actions. While there are better and worse values of \(\rho\), it may be argued that there is no ``bad'' choice; lowering \(\rho\) will result in deeper hierarchies, but existing levels of the hierarchy will remain intact.

\section{Compute Resource Usage}
The experiments were run using a shared internal CPU cluster with the specifications shown below.
Approximately 40 CPU cores were utilised for approximately 336 hours when producing the final set of results presented in the paper. Prior to this, approximately 20 CPU cores were utilised for approximately 168 hours during preliminary testing. GPU acceleration was not used because the experiments involved tabular reinforcement learning methods.

%\centering
\begin{tabular}{@{}ll@{}}
\toprule
\textbf{Processor}           & 2\(\times\) AMD EPYC 7443     \\
\textbf{Cores per Processor} & 24 Cores                      \\
\textbf{Clock Speed}         & 2.85GHz--4GHz                 \\
\textbf{RAM}                 & 512 GB                        \\
\textbf{RAM Speed}           & 3200MHz DDR4                  \\ \bottomrule
\end{tabular}

\section{Source Code}
\label{sect:source_code}
An implementation of the proposed approach, implementations of the test environments, and the code used to generate all results presented in the paper can be found in the following GitHub repository: \url{https://github.com/bath-reinforcement-learning-lab/Louvain-Skills-NeurIPS-2023}

\newpage
\section{Louvain Algorithm}
\label{sect:louvain_pseudocode}
Algorithm \ref{alg:louvain} shows pseudocode for the Louvain algorithm \citep{Blondel2008}. It takes a graph \(G_0 = (V_{0}, E_{0})\) as input and outputs a set of partitions of that graph into clusters. Each iteration of the algorithm (lines 4--25) produces one partition of the graph.

\SetKwInOut{KwParams}{parameters}
\SetKwBlock{RepeatForever}{repeat}{end}

\begin{algorithm}
\caption{Louvain Algorithm}
\label{alg:louvain}

% Parameters and inputs.
\textbf{parameters:} resolution parameter \(\rho \in \mathbb{R}^{+}\)\;
\textbf{input:} \(G_{0} = (V_{0}, E_{0})\) \tcp*[f]{e.g., the state transition graph of an MDP}
\vspace{0.15cm}

\(i \leftarrow 0\)\;
\RepeatForever{	
	\(C_{i} \leftarrow \lbrace \lbrace u \rbrace \mid u \in V_{i} \rbrace\) \tcp*[r]{define singleton partition}
	
	\(Q_{\text{old}} \leftarrow\) modularity from dividing \(G_{i}\) into partition \(C_{i}\)\;
	\vspace{0.15cm}
	
	\Repeat{\(C_{\text{before}} = C_{\text{after}}\)}{
		\(C_{\text{before}} \leftarrow C_{i}\)\;		
		\ForEach{\(u \in V_{i}\)}{
			find clusters neighbouring \(u\), \(N_{u} \leftarrow \lbrace c \mid c \in C_{i}, v \in V_{i}, v \in c, (u,v) \in E_{i} \rbrace\)\;
			compute the modularity gain from moving \(u\) into each cluster \(c \in N_{u}\)\;
			update \(C_i\) by inserting \(u\) into cluster \(c \in N_{u}\) that maximises modularity gain
		}
		\(C_{\text{after}} \leftarrow C_{i}\)\;
	}(\tcp*[f]{no nodes changed clusters during an iteration})
	\vspace{0.15cm}	
		
	\(Q_{\text{new}} \leftarrow\) modularity from dividing \(G_{i}\) into revised partition \(C_{i}\)\;
	
	\vspace{0.15cm}	
	
	\eIf{\(Q_{\text{new}} > Q_{\text{old}}\)}{
		\(V_{i+1} \leftarrow \lbrace c \mid c \in C_{i} \rbrace\)\;
		\(E_{i+1} \leftarrow \lbrace (c_{j},c_{k}) \mid c_{j} \in C_{i}, c_{k} \in C_{i}, (u,v) \in E_{i}, u \in c_{j}, v \in c_{k} \rbrace\)\;
		\(G_{i+1} \leftarrow (V_{i+1}, E_{i+1})\) \tcp*[r]{derive aggregate graph from current partition}
		\(i \leftarrow i+1\)\;
	}
	{
		break\;	
	}	
}

\vspace{0.15cm}

\textbf{output:} partitions \(C_{0}, \ldots, C_{i-1}\)\;

\end{algorithm}

\newpage
\section{Incremental Discovery of Louvain Skills}
\label{sect:incremental_psuedocode}

Algorithm~\ref{alg:high_level_incremental_pseudocode} presents an approach for incrementally developing a hierarchy of Louvain skills over time, starting with only primitive actions. To update the agent's skill hierarchy, Algorithm~\ref{alg:high_level_incremental_pseudocode} calls upon either Algorithm~\ref{alg:louvain} or Algorithm~\ref{alg:louvain_incremental}, depending on whether the agent's existing skill hierarchy is to be replaced or updated. Algorithm~\ref{alg:louvain_incremental} presents an approach for incrementally updating Louvain partitions. It integrates new nodes into an existing cluster hierarchy.

\begin{algorithm}
\caption{Incremental Discovery of Louvain Skills}
\label{alg:high_level_incremental_pseudocode}

% Variant as input.
\textbf{input:} \(\text{variant} \in \lbrace 1, 2 \rbrace\) \tcp*[r]{which variant of the incremental algorithm to use}
\textbf{input: \(N = \lbrace n_{1}, n_{2}, \ldots \rbrace\)} \tcp*[r]{decision stages to revise skill hierarchy after}

\vspace{0.15cm}	

\(V \leftarrow \emptyset\) \tcp*[r]{initialise empty set of nodes}
\(E \leftarrow \emptyset\) \tcp*[r]{initialise empty set of edges}
\(G \leftarrow (V, E)\) \tcp*[r]{initialise empty state transition graph (STG)}
\(C \leftarrow \emptyset\) \tcp*[r]{initialise empty set of partitions of the STG}
\(V_{\text{new}} \leftarrow \emptyset\) \tcp*[r]{initialise empty set for recording novel states}
\(E_{\text{new}} \leftarrow \emptyset\) \tcp*[r]{initialise empty set for recording novel transitions}

\vspace{0.15cm}

initialise \(Q(s, a)\) for all \(s \in \mathcal{S}, a \in \mathcal{A}(s)\) arbitrarily, with \(Q(\text{terminal state}, \cdot) = 0\)\;

\(t \leftarrow 0\)\;

\RepeatForever{
	initialise environment to state \(S\)
	
	\vspace{0.15cm}		
	
	\If{\(S \notin V\)}{
			\(V_{\text{new}} \leftarrow V_{\text{new}} \cup \lbrace S \rbrace\)\;
		}
	
	\vspace{0.15cm}

	\While{\(S\) is not terminal}{	
		choose \(A\) from \(S\) using policy derived from \(Q\) (e.g., \(\epsilon\)-greedy)\;
		take action \(A\), observe next-state \(S'\) and reward \(R\)\;
		perform macro-Q and intra-option updates using \((S, A, S', R)\)\;
		\(S \leftarrow S'\)\;

		\vspace{0.15cm}	

		\If{\(S' \notin V\)}{
			\(V_{\text{new}} \leftarrow V_{\text{new}} \cup \lbrace S' \rbrace\) \tcp*[r]{add novel state to set of new nodes}
		}

		\vspace{0.15cm}	
		
		\If{\((S, S') \notin E\)}{
			\(E_{\text{new}} \leftarrow E_{\text{new}} \cup \lbrace (S, S') \rbrace\)\tcp*[r]{add novel transition to set of new edges}
		}

		\vspace{0.15cm}	

		\If{\(t \in N\)}{
			add each state \(u \in V_{\text{new}}\) to \(V\)\;
			add each transition \((u, v) \in E_{\text{new}}\) to \(E\)\;
			\(V_{\text{new}} \leftarrow \emptyset\)\;
			\(E_{\text{new}} \leftarrow \emptyset\)\;

			\vspace{0.15cm}	

			\uIf{\(\text{variant} = 1\)}{
				\(C \leftarrow\) partitions of the STG derived from \((V,E)\) using Algorithm~\ref{alg:louvain}\;
				\textit{replace} existing skill hierarchy with skills derived from \(C\)\;
			}
			\ElseIf{\(\text{variant} = 2\)}{
				\(C \leftarrow\) partitions of the STG derived from \((V, E, C)\) using Algorithm~\ref{alg:louvain_incremental}\;
				\textit{revise} existing skill hierarchy based on skills derived from \(C\)\;
			}
			initialise entries in \(Q\) for all new skills arbitrarily, with  \(Q(\text{terminal state}, \cdot) = 0\)\;
			remove entries from \(Q\) for all skills that no longer exist in the revised hierarchy\;
			
		}	
			
		\(t \leftarrow t + 1\)\;				
	}
}

\end{algorithm}

\begin{algorithm}
\caption{Update Louvain Partitions}
\label{alg:louvain_incremental}

\textbf{parameters:} resolution parameter \(\rho \in \mathbb{R}^{+}\)\;
\textbf{input:} \(G_{0} = (V_{0}, E_{0})\) \tcp*[r]{e.g., the state transition graph (STG) of an MDP}

\textbf{input:} \(C = \lbrace C_{0}, C_{1}, \ldots, C_{n} \rbrace\) \tcp*[r]{an existing set of partitions of the STG}

\vspace{0.15cm}

\(i \leftarrow 0\)\;
\RepeatForever{
	\(V_{\text{new}} \leftarrow\) nodes in \(V_{i}\) not assigned to any cluster in \(C_{i}\)\;
	\(C_{i} \leftarrow C_{i} \cup \lbrace \lbrace u \rbrace \mid u \in V_{\text{new}} \rbrace\) \tcp*[r]{define singleton partition over new nodes}
	\(Q_{\text{old}} \leftarrow\) modularity from dividing \(G_{i}\) into partition \(C_{\text{i}}\)
	
	\Repeat{\(C_{\text{before}} = C_{\text{after}}\)}{
		\(C_{\text{before}} \leftarrow C_{i}\)\;		
		\ForEach{\(u \in V_{\text{\text{new}}}\)}{
			find clusters neighbouring \(u\), \(N_{u} \leftarrow \lbrace c \mid c \in C_{i}, v \in V_{i}, v \in c, (u,v) \in E_{i} \rbrace\)\;
			compute the modularity gain from moving \(u\) into each cluster \(c \in N_{u}\)\;
			update \(C_i\) by inserting \(u\) into cluster \(c \in N_{u}\) that maximises modularity gain
		}
		\(C_{\text{after}} \leftarrow C_{i}\)\;
	}(\tcp*[f]{no nodes changed clusters during an iteration})
	\vspace{0.15cm}	
	
	\(Q_{\text{new}} \leftarrow\) modularity from dividing \(G_{i}\) into revised partition \(C_{i}\)\;
	
	\vspace{0.15cm}	
	
	\eIf{\(Q_{\text{new}} > Q_{\text{old}}\) or \(i < |C|\)}{
		\(V_{i+1} \leftarrow \lbrace c \mid c \in C_{i} \rbrace\)\;
		\(E_{i+1} \leftarrow \lbrace (c_{j},c_{k}) \mid c_{j} \in C_{i}, c_{k} \in C_{i}, (u,v) \in E_{i}, u \in c_{j}, v \in c_{k} \rbrace\)\;
		\(G_{i+1} \leftarrow (V_{i+1}, E_{i+1})\) \tcp*[r]{derive aggregate graph from current partition}
		\(i \leftarrow i+1\)\;
	}
	{
		break\;	
	}	
}

\textbf{output:} partitions \(C_{0}, \ldots, C_{i - 1}\)

\end{algorithm}

\end{document}